\documentclass[journal=jacsat,manuscript=article]{achemso}
\usepackage{amsmath,amssymb,epsfig,multirow,hyperref,xcolor}

\newcommand{\q}{\quad}

\newcommand{\na}{\nabla}

\newcommand{\txf}{\textbf}

\newcommand{\de}{\delta}

\newcommand{\veps}{\varepsilon}

\newcommand{\ka}{\kappa}

\newcommand{\ph}{\phi}

\newcommand{\Om}{\Omega}

\newcommand{\fr}{\frac}
\newcommand{\pa}{\partial}

\usepackage{chemformula} 
\usepackage[T1]{fontenc} 

\author{Elyssa Sliheet}
\affiliation[SMU]{Department of Mathematics, Southern Methodist University, Dallas TX 75275}
\altaffiliation{Current Affiliation: Vistra Corp.}
\author{Md Abu Talha}
\affiliation[SMU]{Department of Mathematics, Southern Methodist University, Dallas TX 75275}
\author{Weihua Geng}
\affiliation[SMU]{Department of Mathematics, Southern Methodist University, Dallas TX 75275}
\email{wgeng@smu.edu}

\title[An \textsf{achemso} demo]
  {A DNN Biophysics Model with Topological and Electrostatic Features}

\begin{document}

\begin{tocentry}
%
%
%
%
%
\includegraphics[width=3.25in]{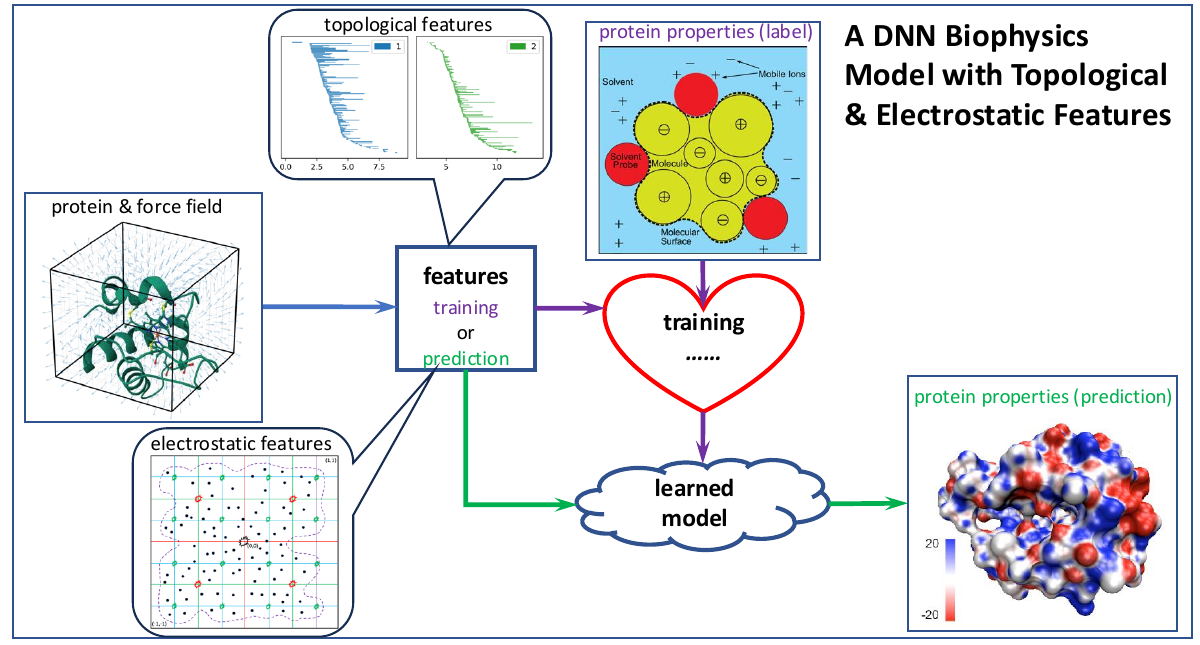}
\end{tocentry}

\begin{abstract}
In this project, we present a deep neural network (DNN)-based biophysics model that uses multi-scale and uniform topological and electrostatic features to predict protein properties, such as Coulomb energies or solvation energies. 
The topological features are generated using element-specific persistent homology (ESPH) on a selection of heavy atoms or carbon atoms.  The electrostatic features are generated using a novel Cartesian treecode, which adds underlying electrostatic interactions to further improve the model prediction. These features are uniform in number for proteins of varying sizes; 
therefore, the widely available protein structure databases can be used to train the network. 
These features are also multi-scale, allowing users to balance resolution and computational cost.  
The optimal model trained on more than 17,000 proteins for predicting Coulomb energy achieves MSE of approximately 0.024, MAPE of 0.073 and $R^2$ of 0.976. Meanwhile, the optimal model trained on more than 4,000 proteins 
for predicting solvation energy 
achieves MSE of approximately 0.064, MAPE of 0.081, and $R^2$ of 0.926, showing the efficiency and fidelity of these features in representing the protein structure and force field. 
The feature generation algorithms also have the potential to serve as general tools for assisting machine learning based  prediction of protein properties and functions.  
\end{abstract}


\section{Introduction}
One of the overarching themes of biology is that structure determines function, 
which becomes more significant when we explore the biological world at the molecular level. 
With the advancement and availability of structure determination techniques such as 
X-ray crystallography \cite{Bragg:1912, Crowfoot:1935}, 
NMR spectroscopy \cite{Aue:1976}, 
cryo-electron microscopy \cite{Cheng:2015}, etc., 
there are rapidly growing 3D structures of protein, nucleic acids, and complex assemblies, etc. 
on publicly accessible repository such as Protein Data Bank (PDB) \cite{Berman:2000}, GenBank \cite{Benson:2012}, etc. 
These structure determination techniques 
have strengths and limitations that make them suitable 
for different types research studies.  
For examples, X-ray crystallography, due to the requirement of crystallization, is best for static high-resolution structures.  
NMR is valuable for studying dynamics in solution, 
particularly for the studies of smaller proteins and complexes.
Cryo-electron microscopy which becomes more and more popular recently, 
excels in visualizing large, complex, and flexible biomolecular systems. 


In addition to experiments, 
computing approaches are also available for the prediction of 
protein structures and properties. 
The main stream of Machine Learning (ML) based protein prediction 
focuses on using amino acid sequence to predict protein structure or structure-related properties, 
resulting in the trending protein language models (PLM) \cite{Bepler:2021, Weissenow:2025}, 
which is a transformer-based model inspired by Natural Language Processing (NLP). PLM 
has a wide range of applications such as protein pKa prediction \cite{Xu:2025}, 
protein structure prediction \cite{Lin:2023}, 
protein design \cite{Madani:2023},  
protein evolution \cite{Zhang:2024}, etc.  
Recently, the appearance of AlphaFold developed by Google DeepMind 
revolutionarily reshaped the structures determination techniques \cite{Abramson:2024}. 



Both PLM and AlphaFold take protein sequences as inputs 
to predict structures or embeddings which keep important features. 
However, the development of ML methods in protein property predictions 
is hindered by the difficulty 
to represent protein structures and force fields using reliable features. 
To this end, the structure data, which is largely variant from protein to protein, 
needs to be converted to uniform size of features 
to be loaded into the ML models.  
Successful cases are the use of graph convolutional network (GCN) for predicting protein functions 
by features extracted from a protein language model and protein structure \cite{Gligorijevic:2021}, 
protein-DNA binding specificity prediction using a wide range of features from geometry \cite{Sagendorf:2024, Mitra:2025},  
protein-ligand binding affinity prediction using topological features, \cite{Wang:2017, Wee:2022}, molecular structure representation using persistent diagram \cite{Townsend:2020}, etc.  
However, the important electrostatic interactions are often ignored in the formation of the protein features because of the challenges from long-range and pairwise nature of the electrostatic interactions. 

In this paper, we provide a novel approach 
that utilizes the far-field expansion in the Cartesian treecode algorithm \cite{Li:2009} to generate electrostatic features. 
To this end, the charges at the atomic centers are represented by multipole moments at the cluster centers, which inherit the key idea of the treecode algorithm \cite{Li:2009}: replacing pairwise particle-particle interactions with particle-cluster interaction. By combining electrostatic features with the topological features generated using a modification of the persistent homology approach used by Cang and Wei \cite{Cang:2017}, we represent the protein structures and the associated force field 
in a uniform and multiscale fashion.
With these features, we train a deep neural network (DNN) to predict protein Coulombic energy and electrostatic solvation energy efficiently and accurately. 
Here, the Coulomb Energy and Solvation energy used as labels are generated by pair-wise Coulombic interactions and the surrogate model for
solving the Poisson-Boltzmann (PB) model 
using a second-order Matched Interface and Boundary (MIB) method based PB solver \cite{Chen:2011}.
Apart from the recent research using Physics Informed Neural Network (PINN) to solve the Poisson-Boltzmann model \cite{Achondo:2025, Chen:2025}, the present work is fundamentally different as a data-driven method with uniform and multiscale features.  

There exist machine learning methods for the prediction of solvation energy 
under the framework of the PB model. 
For examples, 
Huang et al. developed a neural-network formulation for solving the dielectric-boundary PB equation by minimizing a variational electrostatic free-energy functional, 
enabling direct prediction of PB electrostatic solvation energies 
while preserving the physical structure of the continuum model \cite{Huang:2026}; 
R{\"o}cken et al. proposed the ReSolv framework
trained using experimental hydration free energies and {\it ab initio} data, 
achieving near experimental accuracy while accelerating solvation energy prediction 
\cite{Rocken:2024}; 
Ferraz-Caetano et al. introduced an explainable supervised learning approach based on ensemble regression models that predicts solvation energies from physicochemical descriptors
\cite{FerrazCaetano:2024}; 
Wu and Luo presented SPACIER \cite{,Wu:2025}, a large-scale benchmark dataset of solvation energies grounded computed by using AMBER-PBSA \cite{Case2023AmberTools}. 
The major advantage of the proposed method is 
at the feature generation independent of the PB framework, 
thus in addition to this particular applications in predicting Coulombic energy 
and PB-based electrostatic solvation energy, 
the feature generation algorithm developed in this work potentially 
has broader applications to many other ML models, 
as far as protein structures and force fields are the inputs represented uniformly across proteins of varying sizes.

This manuscript is organized as follows. Following this introduction, we present the methods section, which includes both the modeling framework and numerical algorithms, followed by the simulation results. The manuscript ends with a section of conclusion.

\section{Methods}
The objective of this project is to design uniform and multiscale features that represent protein structure and electrostatics, and then to use ML models, such as DNN, to evaluate the effectiveness of these features. The models are trained using labels such as solvation energy computed from surrogate models, e.g. the numerically solved Poisson-Boltzmann model.
We hereby introduce the models and algorithms involved in this work. To study protein structure–property relationships, we focus on the atomic description and the electrostatic interactions. 
The protein properties considered in this paper 
are Coulomb energy $E_\text{coul}$ and electrostatic solvation energy $E_\text{solv}$. 
For a protein, the Coulomb energy $E_\text{coul}$ is the energy required to move atomic charges from infinity to their present positions(i.e. atomic centers specified in the Protein Data Bank (PDB) file. The electrostatic solvation energy  
$E_\text{solv}$ is the electrostatic free energy required to solvate the protein in a solvent, such as water. Formal quantitative definitions of these two properties are provided through mathematical models. 
For the protein-solvent interaction, 
we use the implicit solvent model 
in which water is treated as a continuum in order to reduce computational cost.
From a comparative perspective, we introduce the two most popular implicit solvation models, 
namely the Poisson-Boltzmann model and the Generalized Born's model, which are used to generate labels or features. We then introduce persistent homology and describe how it is used to construct topological features. Finally, we present the Cartesian treecode \cite{Li:2009} algorithm, and its modification into a point-multipole representation, which forms the basis for generating electrostatic features.  

In this project, we adopt a combination of electrostatic features and topological features for the following reasons. Electrostatic features provide a uniform and multiscale representation of the atomic charges,   
accounting for both charge quantities and their locations. Topological features, on the other hand, encode information derived from the atomic centers of selected atoms, such as alpha-carbon atoms or heavy atoms. These features reveal the intrinsic topological invariants, which are not directly represented by electrostatic features. Together, these two types of features efficiently represent charge distributions, their locations, and underlying topological information. 





%


\subsection{The Poisson-Boltzmann model}

%

\begin{figure}[htb!]
\includegraphics[width=2.1in]{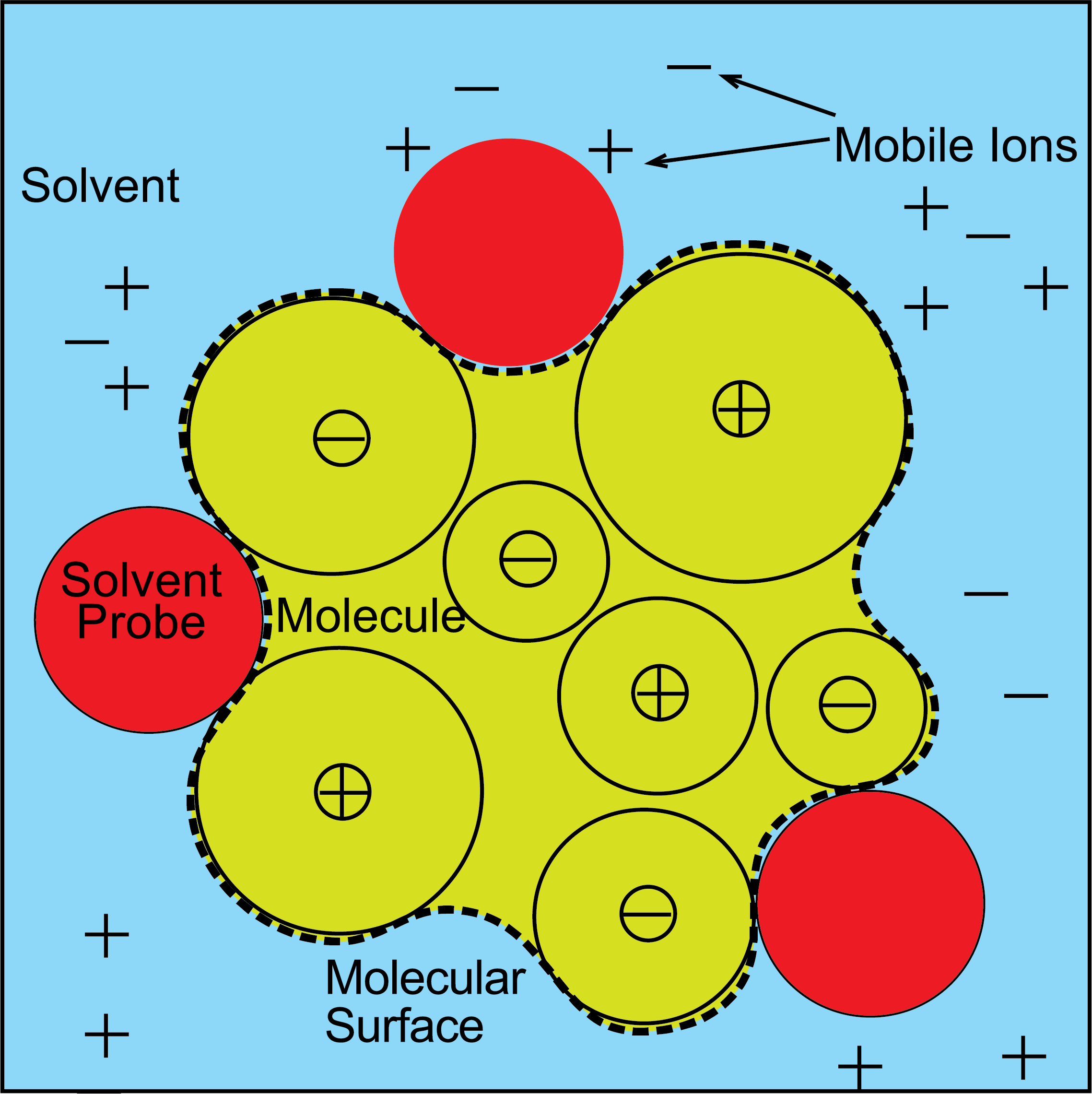}
\includegraphics[width=2.1in]{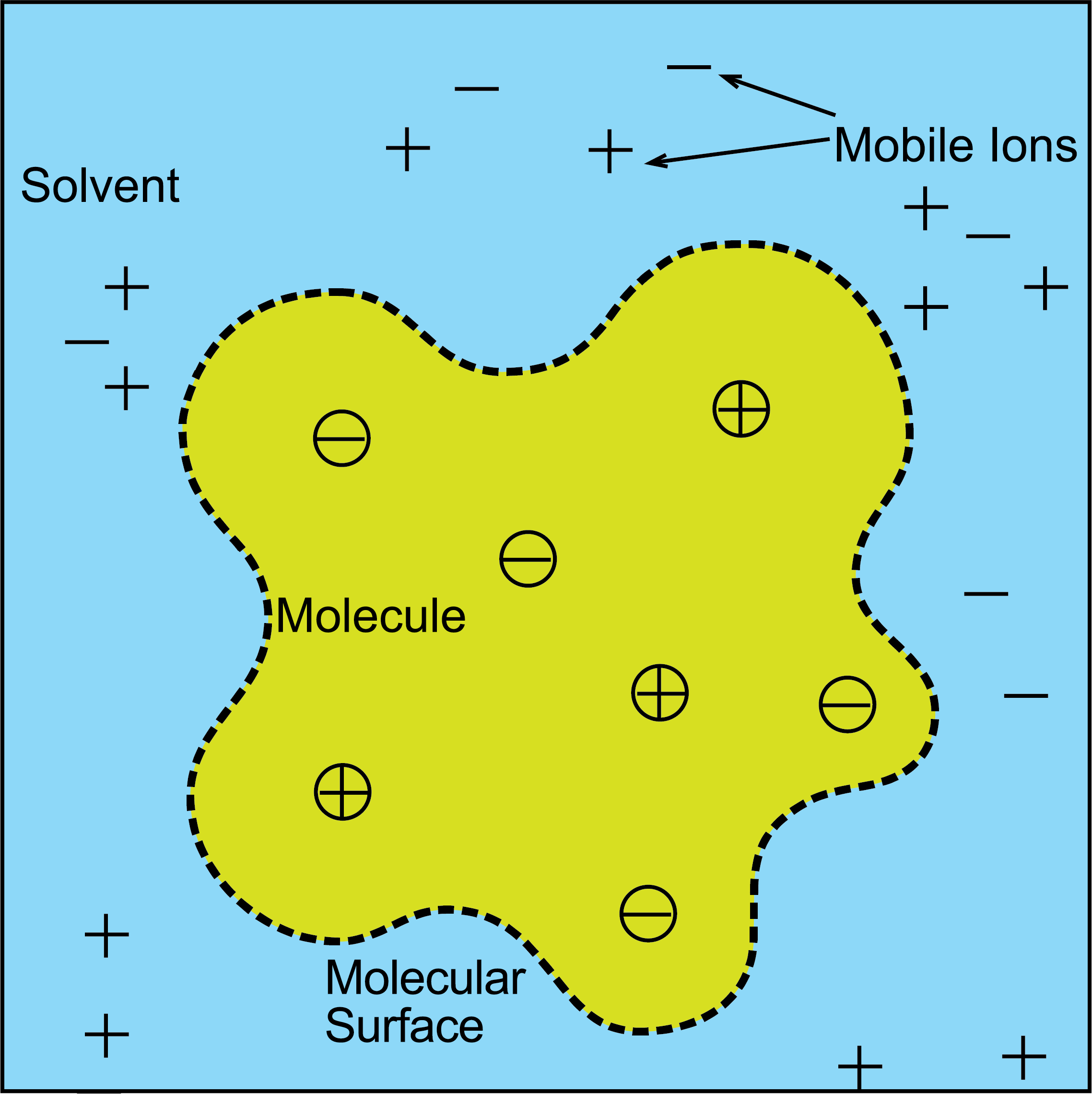}
\includegraphics[width=2.1in]{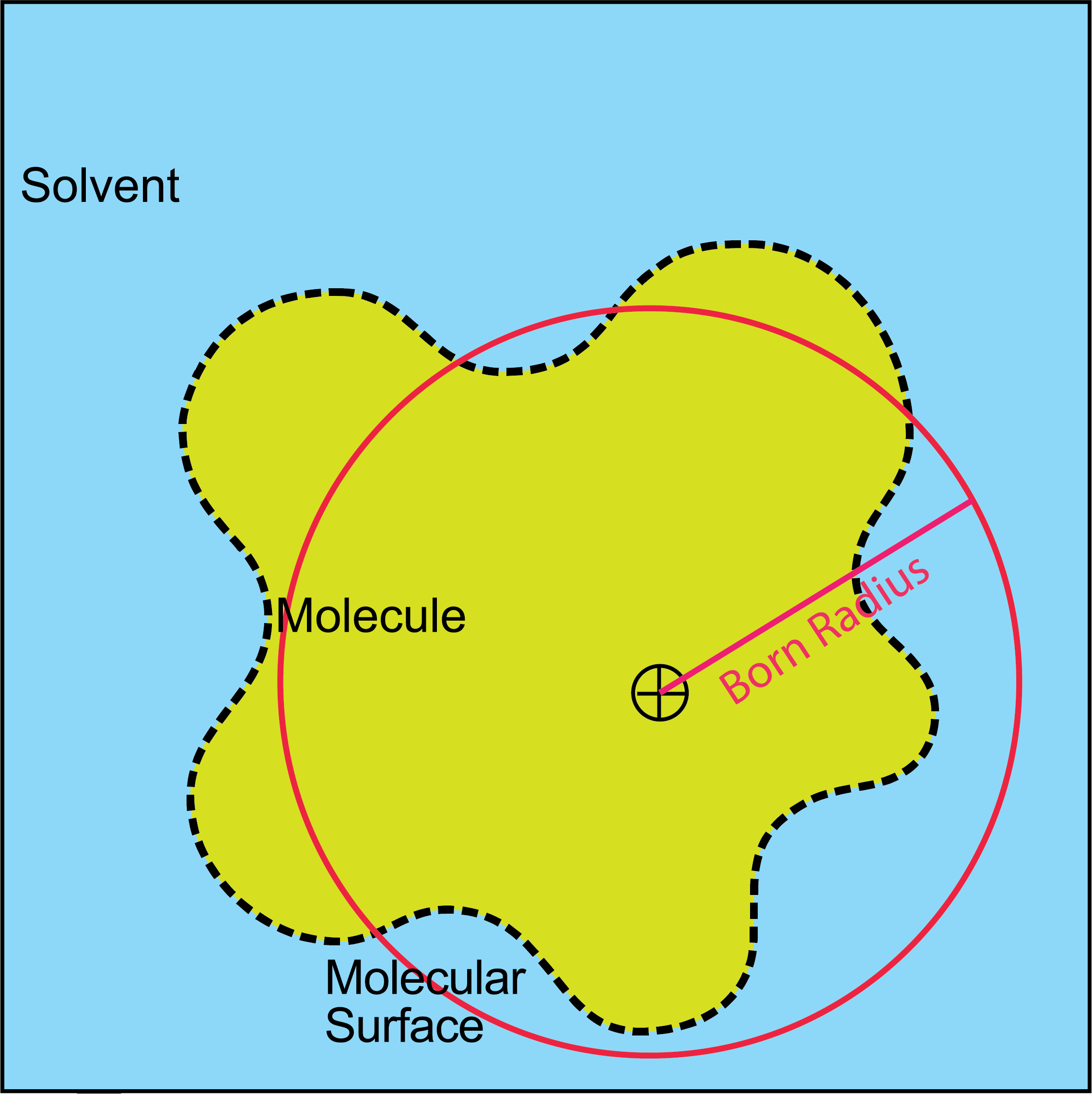}\\
\setlength{\unitlength}{1cm}
\begin{picture}(16.2,0)
\put(   2.5,  2.0){$q_k({\bf x}_k)$}
\put(   1.4,  3.1){$\Omega_1, \varepsilon_1$}
\put(   0.1, 5.2){$\Omega_2, \varepsilon_2$}
\put(   1.1,  1.0){$\Gamma$}
\put(   7.9,  2.0){$q_k({\bf x}_k)$}
\put(   6.9,  3.1){$\Omega_1, \varepsilon_1$}
\put(   5.55, 5.2){$\Omega_2, \varepsilon_2$}
\put(   6.5,  1.0){$\Gamma$}
\put(   12.1,  1.0){$\Gamma$}
\put(   13.5,  2.0){$q_k({\bf x}_k)$}
{\small \put(   14.5,  3.2){$R_k$}}
\end{picture}
\vskip -15pt
(a)~~~~~~~~~~~~~~~~~~~~~~~~~~~~~~~~~~~~ (b) ~~~~~~~~~~~~~~~~~~~~~~~~~~~~~~~~~ (c) 
\vskip -10pt
\caption{\small Implicit solvent models. (a) The Molecular Surface $\Gamma$: the trace of the solvent probe (shown in red) when it is rolled on contacting the spherical atoms of the protein (shown in green with a centered charge); (b) The Poisson--Boltzmann Model: two domains $\Omega_1$ (dielectric constant $\epsilon_1$ with partial charges as weighted summation of delta functions) and $\Omega_2$ (dielectric constant $\epsilon_2$ with mobile ions modeled by the Boltzmann distribution) are separated by the molecular surface $\Gamma$; (3) The Generalized Born Model: the protein is represented as a collection of $N_c$ spherical atoms centered at ${\bf x}_k$ with charge $q_k$ and Born Radius $R_k$ (only the $k$th atom is shown).} 
\label{fig_models}
\end{figure}

Figure~\ref{fig_models} depicts the popular implicit solvent models. 
In Fig.~~\ref{fig_models}(a) a protein is represented by a collection of $N_c$ spherical atoms with centered partial charges.  
The molecular surface $\Gamma$ (also known as the solvent excluded surface \cite{Richards:1977, Connolly:1985}) 
is defined by the trace of a water molecule represented by a red sphere  
rolling on contacting with the protein atoms.  The Poisson-Boltzmann model is shown in Fig.~1(b), where   
the molecular surface $\Gamma$ divides the entire computational domain $\Omega$ into 
the protein domain $\Om_1$ with
dielectric constant $\veps_1$
and
atomic charges $q_k$ located at ${\bf x}_k, k=1\,{:}\,N_c$,
and the solvent domain $\Om_2$ with 
dielectric constant $\veps_2$ and dissolved mobile ions.
Assuming a Boltzmann distribution for the ion concentration,
and
considering the case of two ion species with equal and opposite charges 
(e.g.~Na$^+$, Cl$^-$),
in the limit of weak electrostatic potential
one obtains  the linearized Poisson-Boltzmann (PB) model~\cite{Baker:2005},
\begin{subequations}
\begin{align}
\label{eq1}
\veps_1\na^2 \ph_1(\txf{x}) &= 
-{4\pi}\sum_{k=1}^{N_c} q_k\de (\txf{x}-\txf{x}_k), \quad {\bf x} \in \Om_1, \\
\label{eq2}
\veps_2\na^2 \ph_2(\txf{x}) &=
\ka^2\ph_2(\txf{x}), \quad {\bf x} \in \Om_2, \\
\label{jump}
\ph_1(\txf{x}) &= \ph_2(\txf{x}),\q \veps_1
\fr{\pa\ph_1}{\pa n}({\bf x}) = \veps_2\fr{\pa\ph_2}{\pa n}({\bf x}), 
\quad {\bf x} \in \Gamma,
\end{align}
\end{subequations}
where $\ka$ is the inverse Debye length measuring the salt concentration,
and
the potential satisfies a zero far-field boundary condition $\lim\limits_{|{\bf x}| \rightarrow \infty} \phi({\bf x}) = 0$.

The PB model governs the electrostatic potential $\phi$ in the entire space. Theoretically after $\phi$
is obtained, its gradient will produce electrostatic field while its integral will generate potential energy. However, there are many challenging issues on properly obtaining the field and energy (e.g. definition of field on molecular surface $\Gamma$ \cite{Gilson:1993, Geng:2011}). Our attention for this project is on the energy side as described below. 
The electrostatic free energy is given as 
{\small
\begin{equation}
E_\text{free} = \frac{1}{2} \int_\Omega \rho({\bf x}) \phi({\bf x}) d{\bf x} = \frac{1}{2}\sum\limits_{k=1}^{N_c} q_k\phi({\bf x}_k) = \frac{1}{2}\sum\limits_{k=1}^{N_c} q_k(\phi_\text{reac}({\bf x}_k)+\phi_\text{coul}({\bf x}_k)) = E_\text{solv}+E_\text{coul} \label{eq_energy}
\end{equation}}
where $\rho({\bf x}) = \sum\limits_{k=1}^{N_c} q_k\de (\txf{x}-\txf{x}_k)$ is the charge density as a sum of partial charges weighted delta functions and  the $E_\text{solv} =  \frac{1}{2}\sum\limits_{k=1}^{N_c} q_k \phi_\text{reac}({\bf x}_k)$ term is the electrostatic solvation energy. 
The $\phi_\text{reac}({\bf x}_k)$ is the reaction potential at ${\bf x}_k$ as the remaining component 
when Coulomb potential $\phi_\text{coul} ({\bf x}_k) = \sum\limits_{j=1, j \ne k}^{N_c} \frac{q_j}{\epsilon_1 |{\bf x}_k - {\bf x}_j|}$ is taken away from the electrostatic potential $\phi ({\bf x}_k)$.

Solving the PB model numerically 
by grid-based methods is challenging because
(1)~the protein is represented by singular point charges,
(2)~the molecular surface is geometrically complex,
(3)~the dielectric constant is discontinuous across the surface,
(4)~the domain is unbounded.
%
To overcome these numerical difficulties, 
many finite difference interface methods have been developed 
\cite{Chern:2003, 
LeVeque:1994, 
Qiao:2006}.
For the purpose of dealing with arbitrarily shaped  dielectric interfaces based
on a simple Cartesian grid,
a matched interface and boundary (MIB) PB solver
\cite{Zhou:2006a,Yu:2007,Geng:2007,Chen:2011,Geng:2017} 
has been developed through rigorous treatments of geometrical and charge singularities.
Alternatively, boundary element methods (BEM) for the PB model were developed
later~\cite{
Zauhar:1988,
Juffer:1991,
Liang:1997,
Boschitsch:2002,
Bordner:2003,
Lu:2007, 
Geng:2013b, 
Chen:2023} 
with several inherent advantages,
(1)~only the molecular surface is discretized rather than the 
entire solute/solvent volume,
(2)~the atomic charges are treated analytically,
(3)~the interface conditions are accurately enforced,
(4)~the far-field boundary condition is imposed analytically, 
(5)~the matrix-vector product in iterative methods can be accelerated by fast algorithms \cite{Barnes:1986,Greengard:1987, Li:2009, Tausch:2003}. In this paper, we computed the solvation energies using the MIBPB solver \cite{Chen:2011} since accuracy is of critical importance for the generation of labels ($E_\text{solv}$) that will be used to train the neural network.  


\subsection{Generalized Born's Model}
If ion effects are not considered, 
the solvation energy and the reaction potential can also be efficiently computed 
by using the generalized Born's (GB) model, 
an approximation to the Poisson's model. 
The GB model has significantly reduced computational cost 
as opposed to the PB or Poisson models, 
whose solutions require solving a 3D partial differential equation.  
Although we did not use the GB model to produce any numerical results for this project, 
we describe the GB model for its relationship with the PB model and its potential usage in electrostatic interactions.  
In fact, the GB model can calculate the electrostatic solvation energy which can be used 
either as the label or as a feature \cite{Chen:2021, Chen:2024} in our ML framework. 
It can also produce the reaction potentials at the atomic center, 
which can be used to generate the second level of electrostatic features 
as shown in Fig.~\ref{fig_e-features} to be explained within that content. 
Under the GB model, the total electrostatic solvation energy is
\begin{align}
E_{\text{solv}} =\frac{1}{2} \left(\frac{1}{\epsilon_2}-\frac{1}{\epsilon_1}\right)\left( \sum_{i=1}^N \frac{q_i^2}{a_i} + \sum_{j \neq i}^N \frac{q_i q_j}{r_{ij}}\right)
\approx \frac{1}{2} (\frac{1}{\epsilon_2}-\frac{1}{\epsilon_1}) \sum_{i=1}^N \sum_{j=1}^N\frac{q_i q_j}{f_{ij}^{\text{GB}}}
\label{eq_SolEngAll}
\end{align}
where $r_{ij}$ is the distance between the atomic centers of atoms $i$ and $j$, 
$f_{ij}^{\text{GB}}$ is the effective Born radii ($i=j$) or effective interaction distance ($i \ne j$). This PB-GB relationship can be seen in Fig.~\ref{fig_models}(c), in which the electrostatic solvation energy of the sphere equals that of the molecule, both with a charge at the atomic center. 
Some details about the GB model can be found in the supplement material.

\subsection{Topological Features}
This section involves many fundamental concepts in Algebraic Topology. 
We use {\it italic} fonts to emphasize important terms 
and use {\bf bold} fonts to indicate their sequential definition if there are any.    
The fundamental task of topological data analysis is to extract {\it topological invariants} as the intrinsic features of the underlying space. 
For the prediction of protein properties, 
we expect the topological invariants such as 
independent components, 
rings, 
cavities, etc. 
carry useful information 
which cannot be discovered from geometric observation and measurement. 
This is realized by the aid of {\it homology},  
a mathematical framework that assigns algebraic objects (groups) 
to a topological space to measure its shape, especially its holes in various dimensions. 
Below we briefly introduce the {\it simplicial homology} and the {\it persistent homology}, 
and their associated terms and definitions. 
Following that, we explain how these homologies and their computations with protein structures 
are used to generate topological features.

\subsubsection{Simplicial Homology} 
Topological invariants in a discrete data set (e.g. the collection of atomic positions of a protein)
can be studied using {\it simplicial homology} 
which uses a specific rule (e.g. the choice of proximity parameter $\epsilon$ and an associated procedure 
to determine the edges connecting the vertices) to identify {\it simplicial complexes} (will be defined later) from {\it simplexes}. 
Here the simplex represents the simplest possible polytope 
in any given dimension like point, 
line segment, 
triangle, 
tetrahedron, etc. 
Formally, A \textbf{$k$-simplex} is the convex hull of $k+1$ affinely independent points $v_0, v_1, \dots, v_k$:
\begin{equation}
\sigma^k = \left\{ \sum_{i=0}^k \lambda_i v_i \;\middle|\; \lambda_i \geq 0, \sum \lambda_i = 1 \right\}
\end{equation}
Each subset of $m+1$ of these vertices forms an $m$-{\bf face} of the simplex, which is itself an 
$m$-simplex.

The {\bf boundary} of a $k$-simplex $[v_0, \dots, v_k]$ is:

\begin{equation}
\partial_k [v_0, v_1, \dots, v_k] = \sum_{i=0}^{k} (-1)^i [v_0, \dots, \hat{v}_i, \dots, v_k]
\end{equation}
where $\partial_k$ is the boundary operator from $k$-chains to $(k-1)$-chains and $\hat{v}_i$ means vertex $v_i$ is omitted. 
Here a $k$-{\bf chain} is a {\it formal} linear combination of 
k-dimensional simplexes in a {simplicial complex}.

With the concepts of simplex, face, boundary, chain, etc.,  
we define that a \textbf{simplicial complex} $K$ is a collection of simplices such that:
\begin{enumerate}
    \item Every face of a simplex in $K$ is also in $K$.
    \item The intersection of any two simplices is either empty or a common face.
\end{enumerate}
Thus, each simplicial complex corresponds to a topological space by filling in the simplexes and gluing them together along shared faces.

Let \( C_k(K) \) be the group of \(k\)-chains, 
and let
\begin{equation}
\cdots \xrightarrow{\partial_{k+1}} C_k(K) \xrightarrow{\partial_k} C_{k-1}(K) \xrightarrow{\partial_{k-1}} \cdots
\end{equation}
be the sequence of boundary maps 
{$\partial_k$}, 
the $k$-{\bf homology group} of \(K\) is defined as:
\begin{equation}
H_k(K) = \frac{\ker(\partial_k)}{\operatorname{im}(\partial_{k+1})}
\end{equation}
where
\begin{itemize}
  \item \(\ker(\partial_k)\) is the group of \(k\)-cycles: \(Z_k = \{ c \in C_k \mid \partial_k(c) = 0 \}\)
  \item \(\operatorname{im}(\partial_{k+1})\) is the group of \(k\)-boundaries: \(B_k = \{ \partial_{k+1}(c) \mid c \in C_{k+1} \}\)
\end{itemize}
Hence, the \(k\)-th homology group is the group of cycles modulo boundaries:
\begin{equation}
H_k = Z_k / B_k.
\end{equation}
Taking $H_1$ as an example, $Z_1=\ker(\partial_1)$ collects 1-chains (formal sum of edges) without boundary points thus general loops, while $B_1=\operatorname{im}(\partial_{2})$ collects boundaries of 2-chains (a collection of triangles) that bound a filled-in region thus trivial holes, $H_1 = Z_1 / B_1$ is then the collection of actual holes in the space.

\subsubsection{Persistent Homology} 
However, the simplicial homology of a complex associated to a point set at a particular proximity parameter $\epsilon$ 
is insufficient to characterize the signal or noise.  
This calls for the ideas of {\it filtration} and {\it persistent homology} \cite{edelsbrunner:2000, zomorodian:2005}, 
which can identify and connect complexes at different level of complexity 
and record the appearance and disappearance of the homology group. 
To this end, we introduce the concept of 
\textbf{filtration}, 
which is a sequence of nested simplicial complexes:
\begin{equation}
\emptyset = K_0 \subseteq K_1 \subseteq \cdots \subseteq K_n = K.
\end{equation}
Each $K_i$ is a simplicial complex, and the sequence grows over time or scale.
Instead of computing just 
the $k$-{\bf homology group} $H_k(K)$, 
which is an algebraic object capturing the 
$k$-dimensional holes in a topological space represented by a simplicial complex $K$,
we compute $H_k(K_i)$ for each 
$i$, the index for time or scale parameters.  

As $i$ or its corresponding scale parameter grows,  
the duration, as measured by the change of $i$ or its corresponding scale parameter, of the homology class $H_k(K_i)$ between its appearance and disappearance 
is called {\it persistence}.   
The importance of persistence is that given a parameterized family of spaces, 
topological features which persist over a significant parameter ranges are considered as {\it signal} 
while those with short-lived features are treated as {\it noise} \cite{Ghrist:2007}. 
In practice, the persistent homology is computed using the pipeline that involves steps of 
turning data into a sequence of spaces (a filtration), 
tracking the birth and death of topological features using algebra, 
implementing using linear algebra tools, and 
visualization. The pipeline is realized by many established software, e.g. GUDHI \cite{Boissonnat:2014}, as our choice. 

We next introduce the barcode, persistent diagram, and persistent image as approaches to visualize the persistent homology. These approaches are different in format and visualization  but are equivalent mathematically. Following the definition, we provide the algorithms to extract topological features from barcode and persistent image. 

\subsubsection{Barcode, Persistence Diagram and Persistence Image }
Persistent homology captures the evolution of topological features throughout a filtration process, which could be encoded in the forms of barcode, persistent diagram, and persistent image. 

A {\it barcode} for dimension $k$ is the multiset $\{[b_j, d_j)\}_j$, where each interval corresponds to a $k$-dimensional homology class with index $j$, 
which appears at $b_j$ and disappears at $d_j$. 
In the picture of a barcode, 
each class is visualized as a bar starting at $b_j$ and ending at $d_j$. 
Thus the {\bf barcode} is a graphical representation of persistent homology  
as a collection of horizontal line segments in a plane 
whose horizontal axis corresponds to the filtrition parameter 
and whose vertical axis represents an (arbitrary) ordering of homology generators such as $H_0$, $H_1$, $H_2$, etc. 
Different colors can be used to distinguish different dimensions.  

A {\it persistent diagram} represents each homology class as a point $(b_j,d_j)$ in the birth–death plane, where $b_j$ and $d_j$ denote the filtration parameters at which the class appears and disappears. The persistent diagram thus is a scatter plot of points, where distance from the diagonal measures their significance. 

From their definitions, we can tell barcode and persistence diagram carry equal amount of information. 
We therefore only use barcode to generate topological features. 
The persistent image, which is built on top of persistent diagram to generate fixed-size 2-d image, 
is readily used as features or inputs for the convolution neural network. 
Details about persistent image and its formal mathematical definition can be found in supplementary materials.   

\subsubsection{Feature generation from Barcode}
To make the topological features uniform and physically informed, 
we choose to use Element-Specific Persistent Homology (ESPH) 
to extract topological features at different levels of complexity. 
ESPH has been successfully applied to the prediction of protein–ligand binding. \cite{Cang:2017}. The core idea is to select atomic elements that form the point cloud by incorporating physical, chemical, and biological information relevant to the target problem.
In this project, since solvation energy and Coulomb energy 
are closely related to the amino acid chains, we use two collections of point clouds. One consists of all carbon atoms, which form the backbone of amino acids. The other consists of all heavy atoms \{C, N, O, S\}, whose locations and interconnections are also critical to protein structures. Once the collections of point clouds are determined, the corresponding barcodes \cite{Ghrist:2007}, which encode the persistent homology of the point clouds, are constructed. We use the GUDHI software package \cite{Boissonnat:2014} to generate these barcodes. The resulting barcodes are then processed by the algorithm below to produce vectors of topological features.

\begin{figure}[htbp!]
\begin{center}
\includegraphics[width=3.2in]{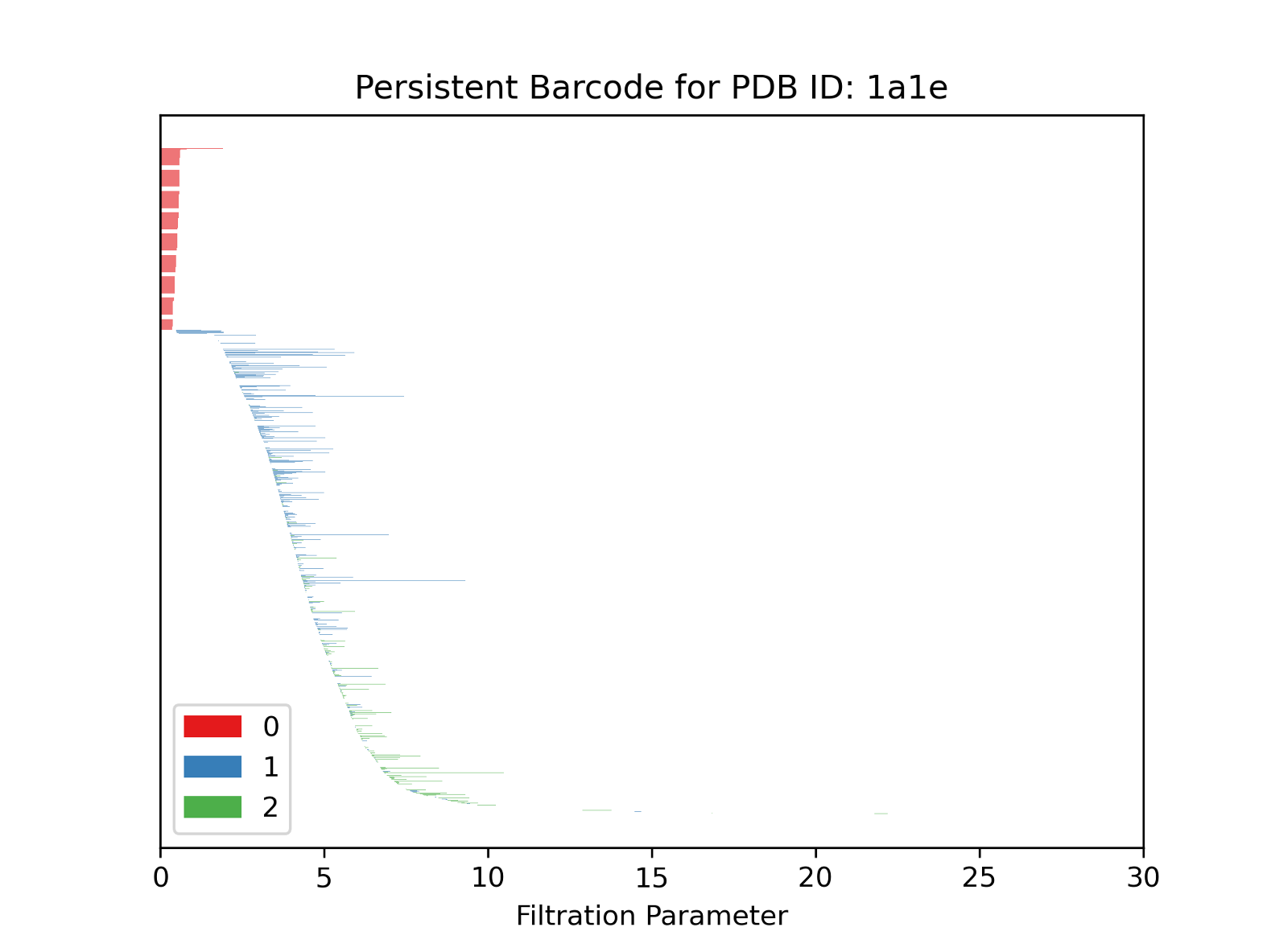}~~~~
\includegraphics[width=3.2in]{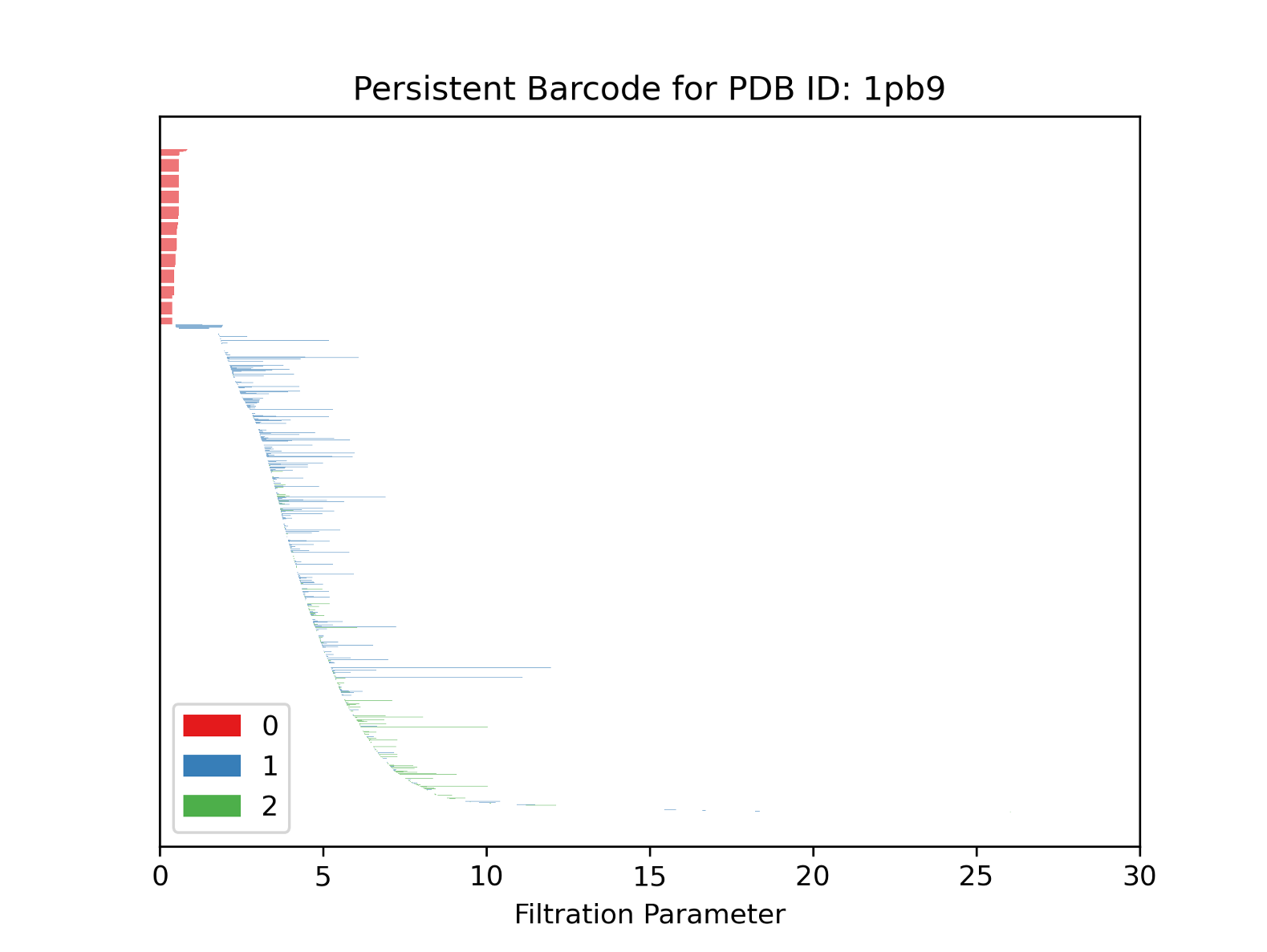}
\caption{\small The barcode of $H_0, H_1, H_2$ generated by the collection of cloud points using heavy atoms in proteins (a) 1a1e and (b) 1pb9.}
\label{fig_barcode}
\end{center}
\end{figure}

Figure~\ref{fig_barcode} shows the barcodes of $H_0, H_1, H_2$ for protein 1a1e, the C-SRC Tyrosine Kinase, and protein 1pb9, the N-methyl-D aspartate receptor, using heavy atoms. 
The persistent homology analysis conducted in this study characterizes key biophysical phenomena, including intermolecular interactions, solvation effects, and hydrophobicity. Topological features are extracted from $H_1$ and $H_2$ barcodes, where the birth, death, and persistence values are computed based on the spatial distribution of carbon atoms and all heavy atoms within the protein structure, 
resulting in a total of 12 feature channels. 

The significance  of 1- and 2-dimensional homology groups lies in the fact that they represent 2-d rings and 3-d voids, respectively, which persist over a wide range of scales and are both topologically and geometrically meaningful. In contrast, the 0-dimensional homology group represents connected components, whose number decreases from the total number of points to 1 and therefore carries limited persistent signal. Higher-dimensional homology groups, although mathematically defined, are typically short-lived in persistence, behave as combinatorial noise, and lack clear geometric meaning.
We use both the collection of heavy atoms and the collection of carbon atoms to capture critical structural and chemical information of proteins.
Features derived from all heavy atoms represent geometric interactions between specific types of atoms in the protein, whereas those computed from carbon atoms primarily reflect hydrophobic interactions and indirectly capture solvation effects. 
The analysis is performed over a distance scale of 0 to 50\AA, which is discretized into uniform bins of 0.25\AA~ to ensure fine-grained resolution of topological patterns. The topological information encoded in the barcodes is then converted into features using the following approach \cite{Cang:2017}.  
 
We define the collection of barcodes as $\mathbb{B}(\alpha, \mathcal{C}, \mathcal{D})$ with the following specifications. \\
$\alpha$ : atom labels (a specific way to obtain a collection of atoms {i.e., all heavy atoms in protein, heavy atoms close to the ligand, atoms near the mutated residue, etc.})\\
$\mathcal{C}$ : type of simplicial complex (i.e., Ribs or Cech)\\
$\mathcal{D}$ : dimension (i.e., $H_1$, $H_2$, etc.)

Using the collection, the structured vectors ${\bf V}^b$, ${\bf V}^d$, and ${\bf V}^p$ can be constructed to respectively describe the birth, death, and persistent patterns of the barcodes in various spatial dimensions. Practically, the filtration interval $[0, L]$ is divided into $n$ equal length subintervals and the patterns are characterized on each subinterval. 
The description vectors using their $i$th component for $1\le i < n$,  are defined as:
\begin{align}
    {\bf V}_i^b &= || \{(b_j, d_j) \in \mathbb{B}(\alpha, \mathcal{C}, \mathcal{D}) | (i-1)L/n \le b_j \le iL/n \} ||, \nonumber\\
    {\bf V}_i^d &= || \{(b_j, d_j) \in \mathbb{B}(\alpha, \mathcal{C}, \mathcal{D}) | (i-1)L/n \le d_j \le iL/n \} ||,  \\\nonumber
    {\bf V}_i^p &= || \{(b_j, d_j) \in \mathbb{B}(\alpha, \mathcal{C}, \mathcal{D}) | (i-1)L/n \ge b_j, iL/n \le d_j \} ||.
\end{align}
 
These vectors can be viewed as (1D) images whose pixel value is $||\cdot||$, the cardinality of the set. 
Each pixel is specialized by $i$, $\alpha$, $\mathcal{C}$, $\mathcal{D}$ as indices of corresponding sets. 
In this project, $\alpha \in \{\text{all Carbon atoms in the protein, all heavy atoms in the protein} \}$, $\mathcal{C} = \text{Cech}$, $\mathcal{D} \in \{\text{$H_1$, $H_2$}\}$, resulting in 12 vectors with the choice of $b, d, p$. 
This barcode-based approach is comparable to the method of persistence imaging \cite{Adams:2017}, 
which can be directly integrated into standard machine learning pipelines (e.g., SVMs, neural networks, etc.) 
at the cost of more preprocessing such as smoothing and discretization as explained previously. 
 
\subsubsection{Feature generation from Persistent Image}

The topological features can also be constructed using persistence image alternatively. 
To this end, we first extract persistence diagrams from protein structures 
and convert them into stable, fixed-size image representations. 
The persistence diagrams are obtained separately from two types of point clouds:
the collection of all carbon atoms and the collection of all heavy atoms. 
These two approaches are also used in the barcode-based feature construction as detailed previously. 
For each approach, we consider topological features in homology groups $H_1$ and $H_2$. 
Each persistence diagram is filtered using physically meaningful thresholds 
such that birth and persistence values are restricted to maximum bounds of $12.5$~\AA. 
The resulting persistence domain is discretized using a resolution of $0.25$~\AA, 
which produces a uniform image grid $D=[0, b_{\text{thr}}] \times [0, p_{\text{thr}}]$ with $b_{\text{thr}}=p_{\text{thr}}=50$  for all proteins. 
To emphasize the contribution of long-lived topological features, 
each point $(b_i, p_i) \in D$ is assigned a smooth exponential weight
\begin{equation}
\omega(b_i, p_i) = e^{p_i} - 1 .
\end{equation}

\begin{figure}[htbp!]
\centering
\vspace{0.5em}
~\includegraphics[width=1.5in]{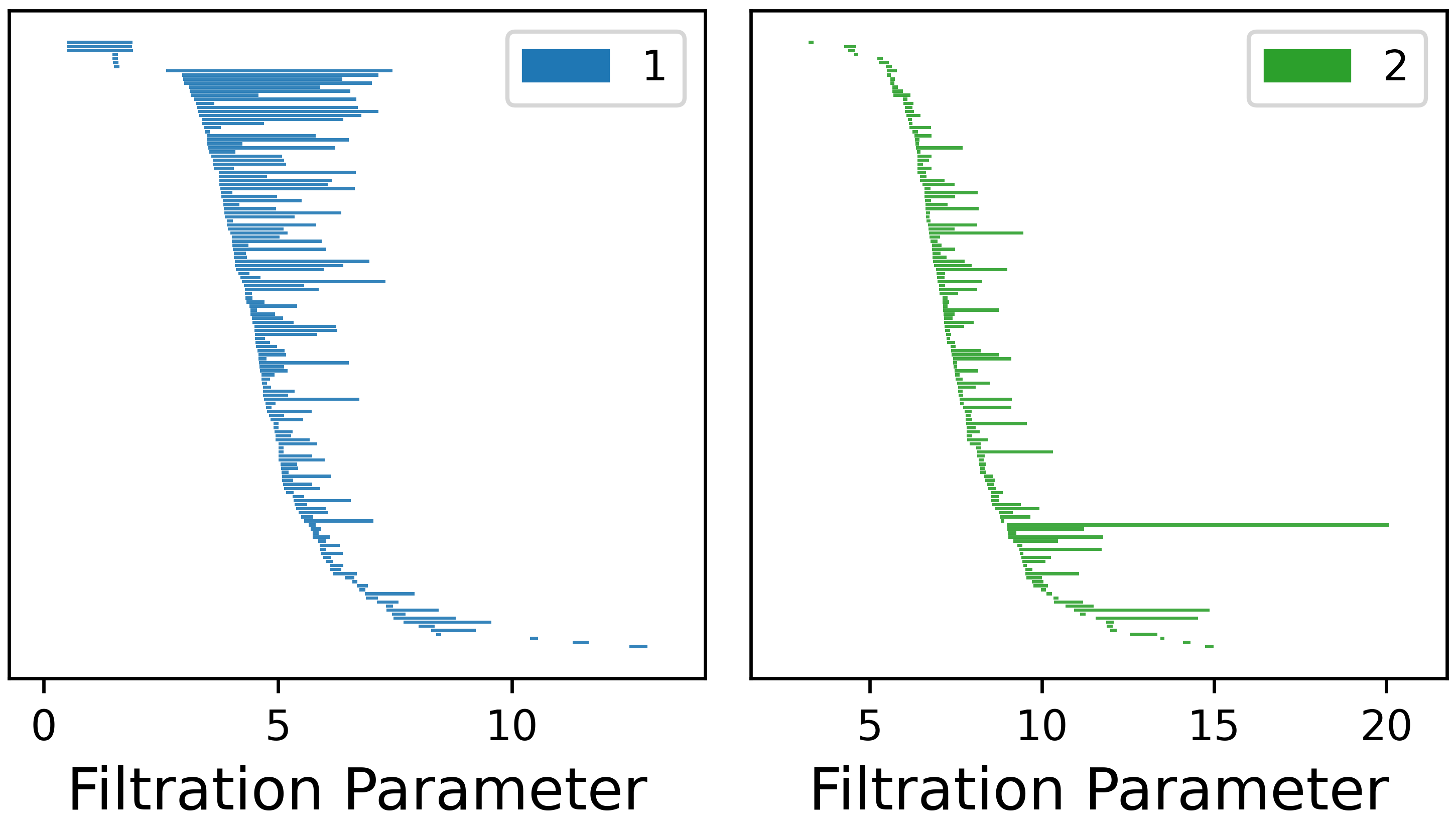}
\includegraphics[width=1.5in]{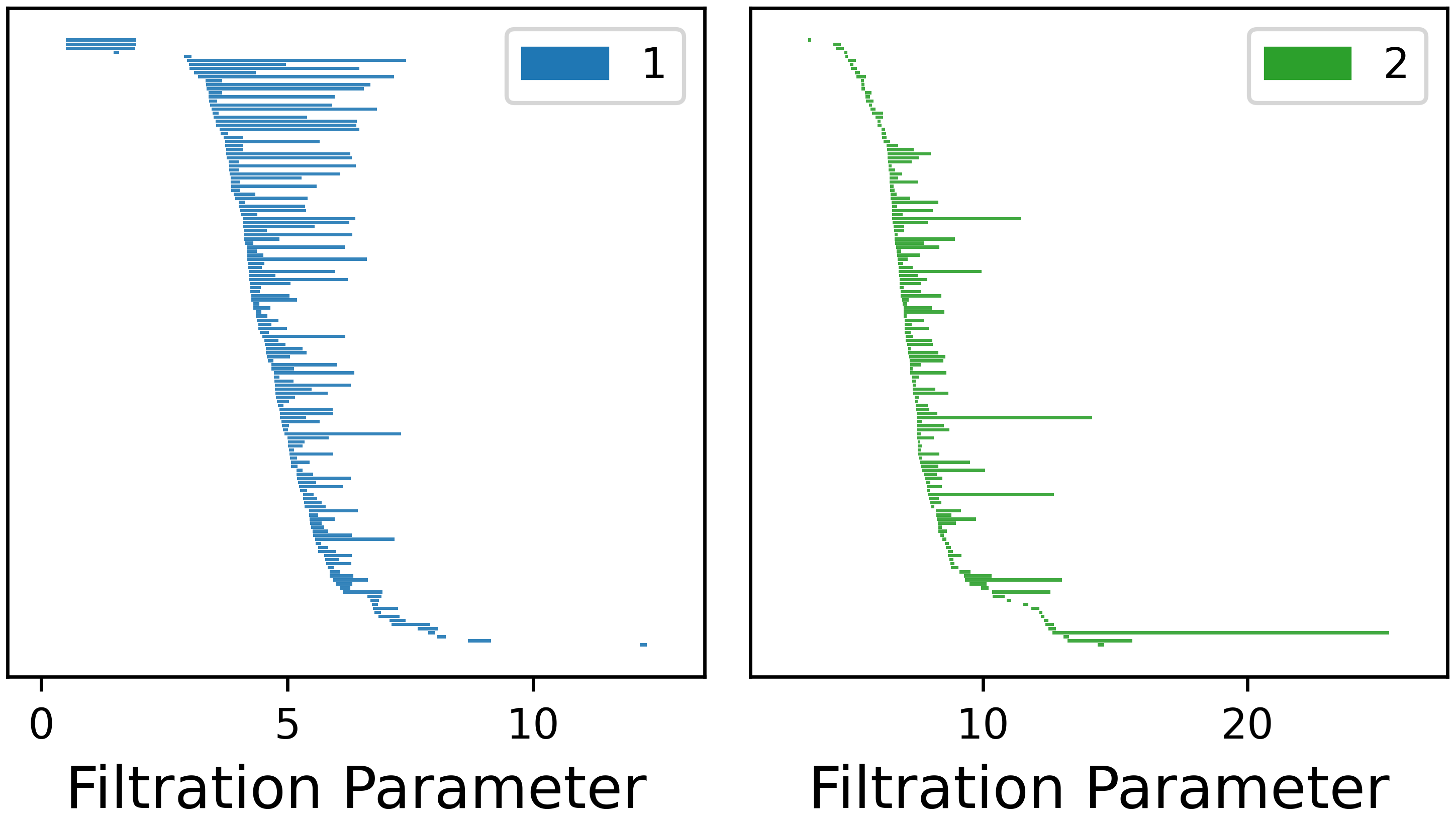}~~~~~
\includegraphics[width=1.5in]{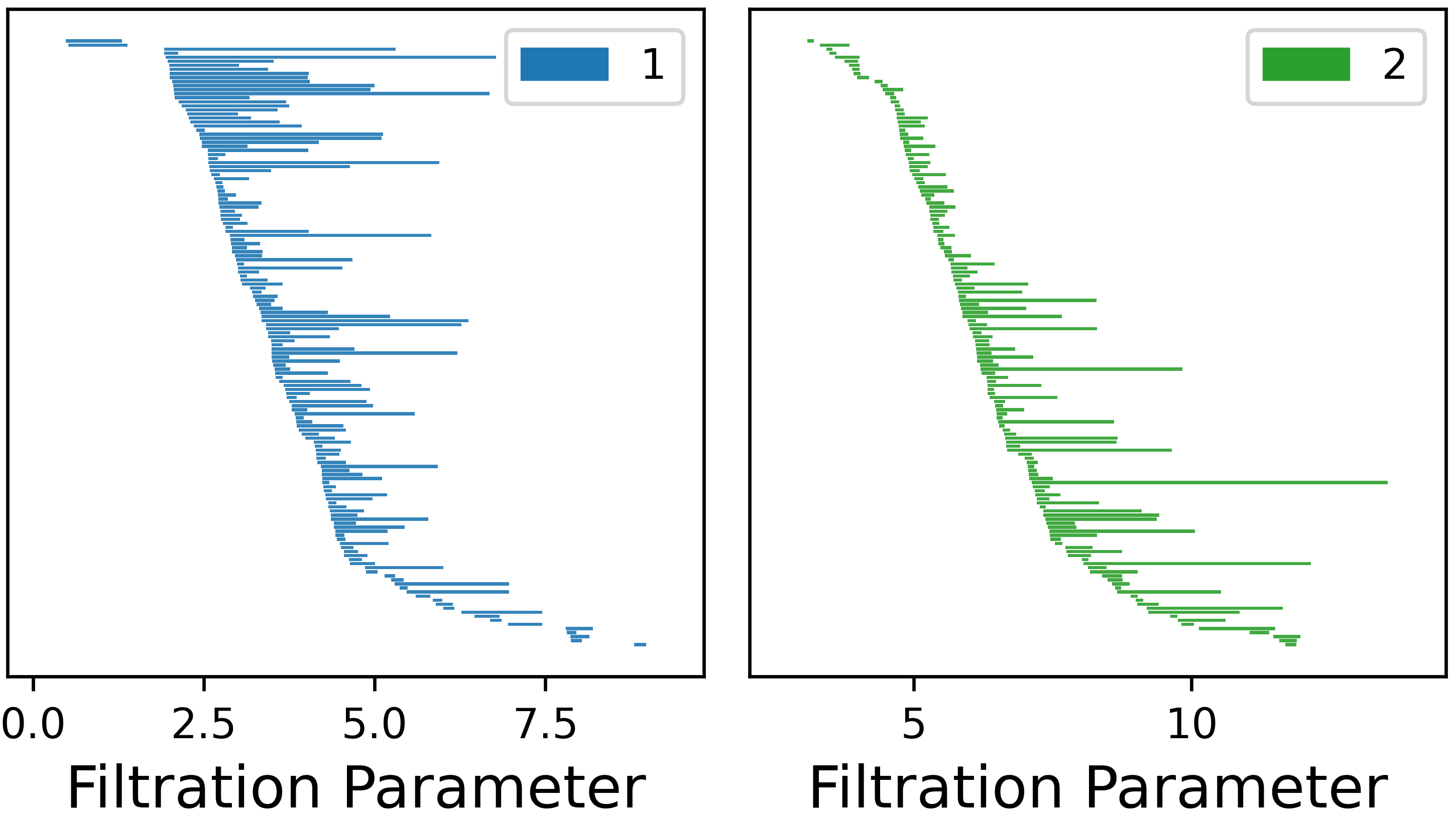}
\includegraphics[width=1.5in]{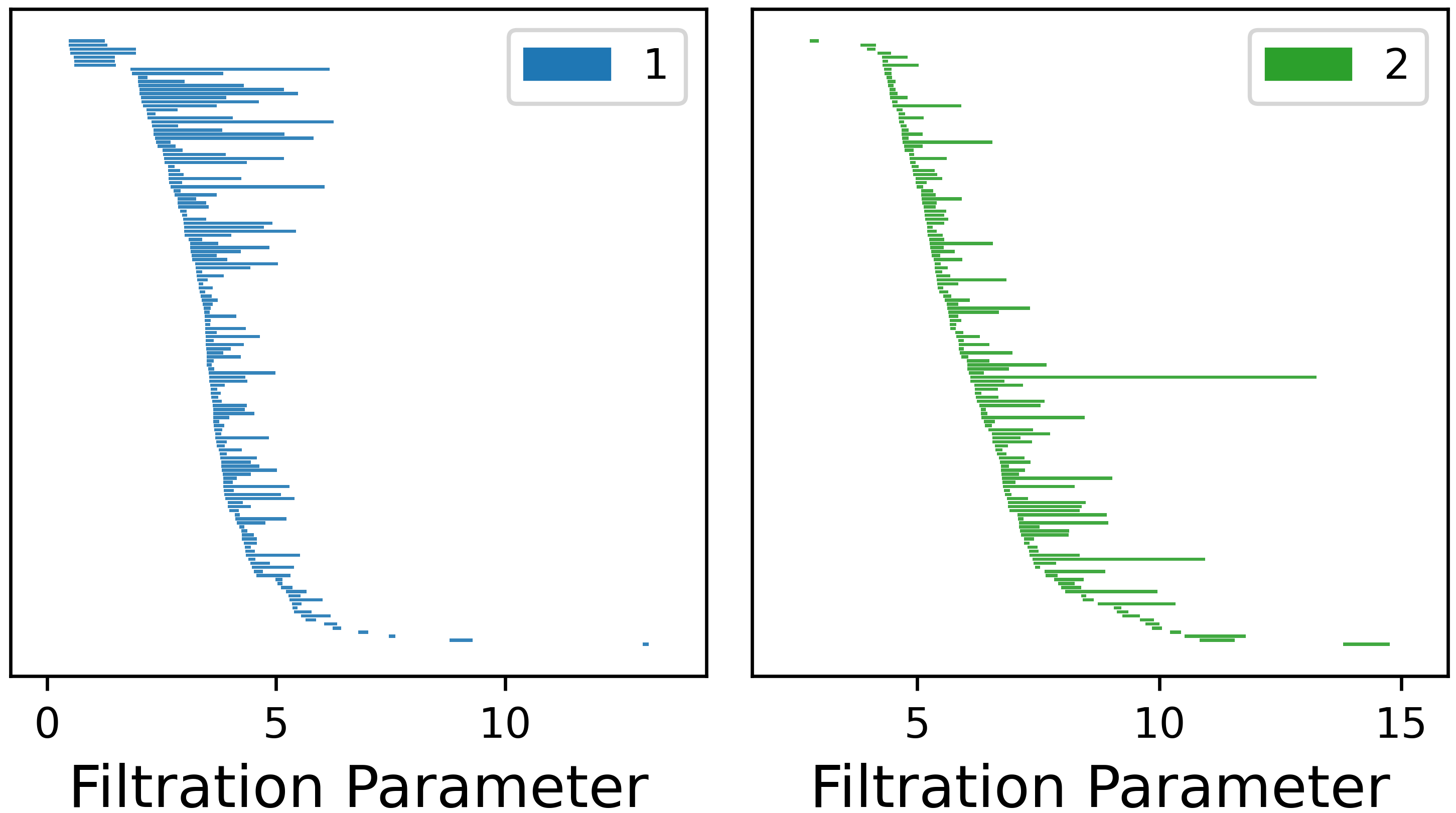}\\

\vspace{0.5em}
\includegraphics[width=0.49\textwidth]{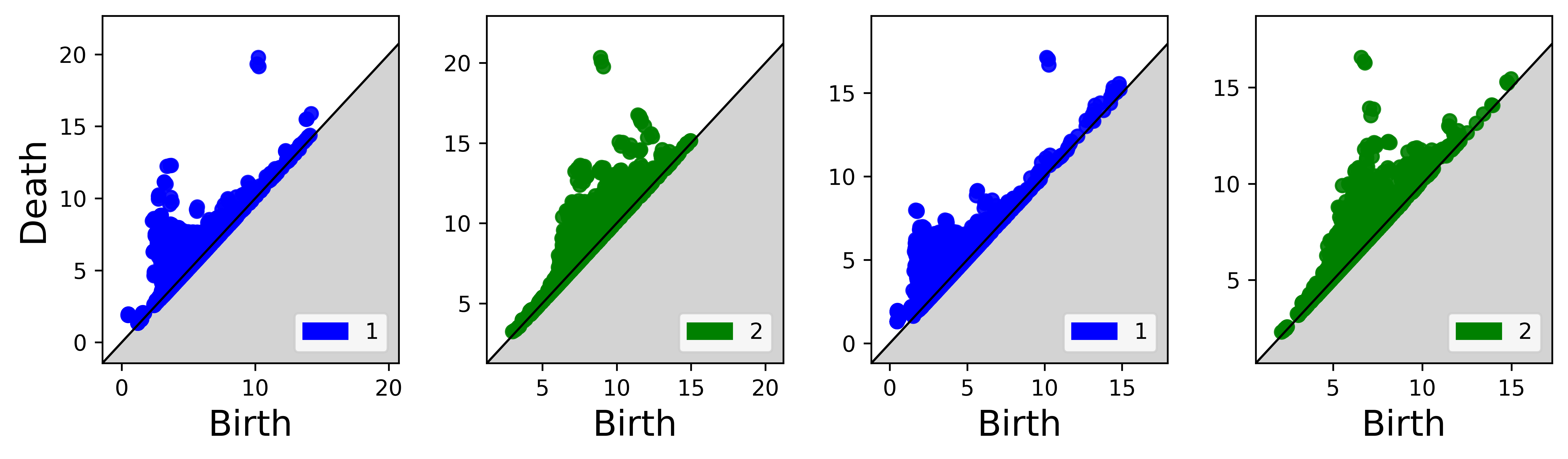}
\hfill
\includegraphics[width=0.49\textwidth]{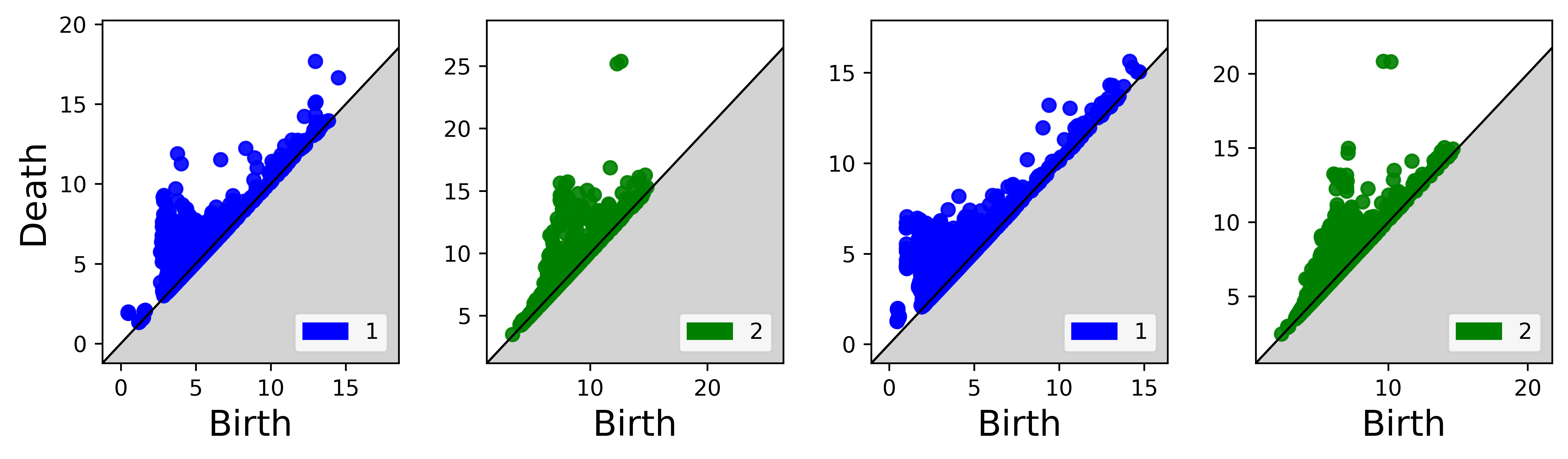}
\vskip 5pt
\includegraphics[width=0.49\textwidth]{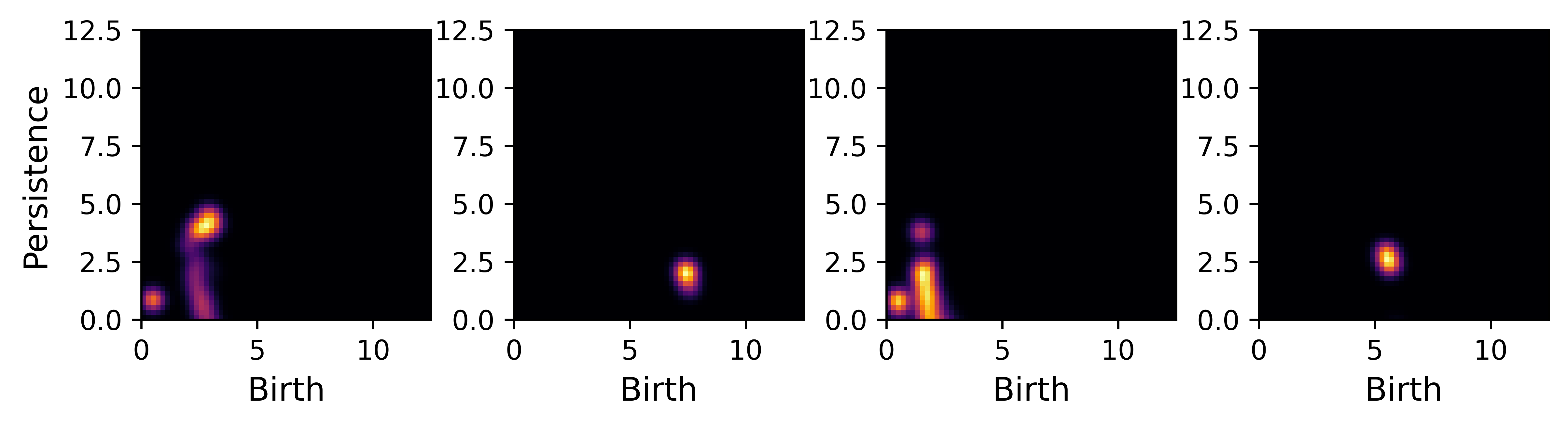}
\hfill
\includegraphics[width=0.49\textwidth]{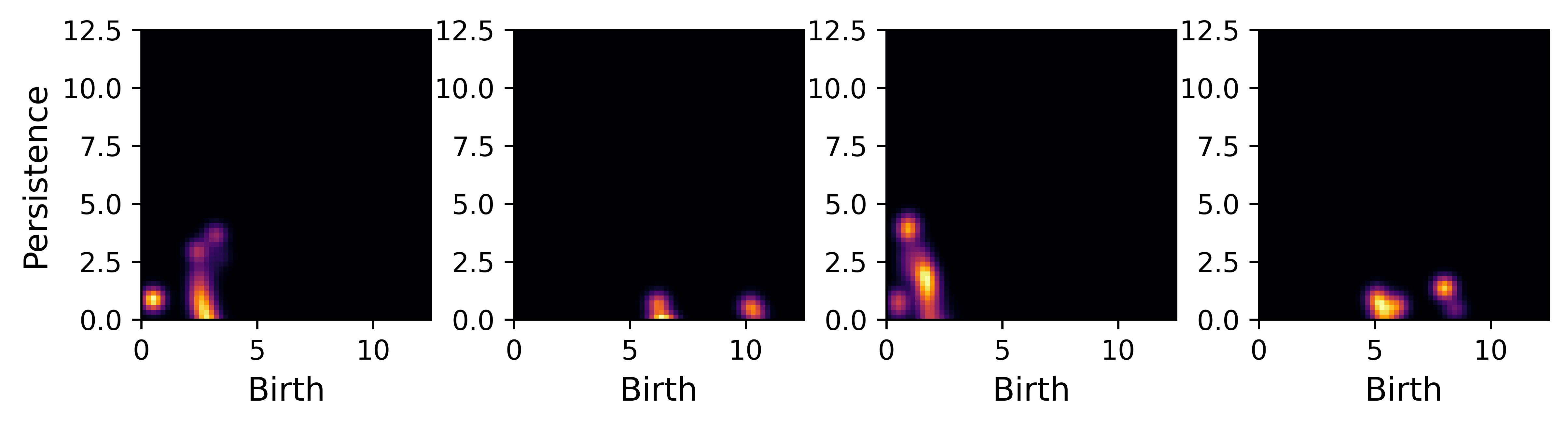}
\makebox[0.48\textwidth][c]{(a) 1a0t}
\makebox[0.48\textwidth][c]{(b) 1a4q}
\caption{\small
Barcode (top) , persistence diagrams (middle) and persistence image (bottom) for proteins 1a0t (a) and 1a4q (b). For each protein, from left to right are: $H_1$ and $H_2$ for carbon atoms, $H_1$ and $H_2$ for all heavy atoms.}
\label{fig_PD_PI}
\end{figure}

This exponential weighting enhances features with larger persistence 
while assigning near-zero weight to short-lived topological noise, 
thereby improving the contrast between significant and insignificant structures in the resulting persistence images.
In this project, persistence images are generated 
using the Gaussian kernel with bandwidth of $0.4$~\AA, 
which is implemented by using the GUDHI library \cite{Boissonnat:2014}.  
For each protein, four persistence images are produced: 
carbon atoms only for $H_1$ and $H_2$, 
all heavy atoms for $H_1$ and $H_2$. 
These four images are stacked to form a four-channel topological feature tensor,
$\mathbf{X} \in \mathbb{R}^{b_{\text{thr}} \times p_{\text{thr}} \times 4}$.
Finally, the resulting persistence image tensor is stored 
in the NumPy format and used as the input topological representation for downstream deep learning models. 
This approach provides a stable, compact, and fully vectorized encoding of protein topology
thus is well suited for ML-based prediction.


In Fig.~\ref{fig_PD_PI}, the barcode of $H_1$ and $H_2$ using carbon atoms and using all heavy atoms for protein 1a0t and 1a4q are provided for comparison purpose with the persistent diagram and persist image. 
The persistence diagrams of the two proteins 1a0t and 1a4q 
show that most of the topological features are short-lived and concentrated near the diagonal, 
while only a small number of them persist. 
The corresponding persistence images highlight these persistent structures 
as bright and localized regions, with background noise effectively suppressed. 
The observed differences between persistence images from the cloud of points of carbon atoms 
and those from all heavy atoms show that topological invariants 
are influenced by the underlying chemical composition.

\subsection{Electrostatic Features}
\subsubsection{Introduction}
\begin{figure}[htbp!]
\begin{center}
\includegraphics[width=3.0in]{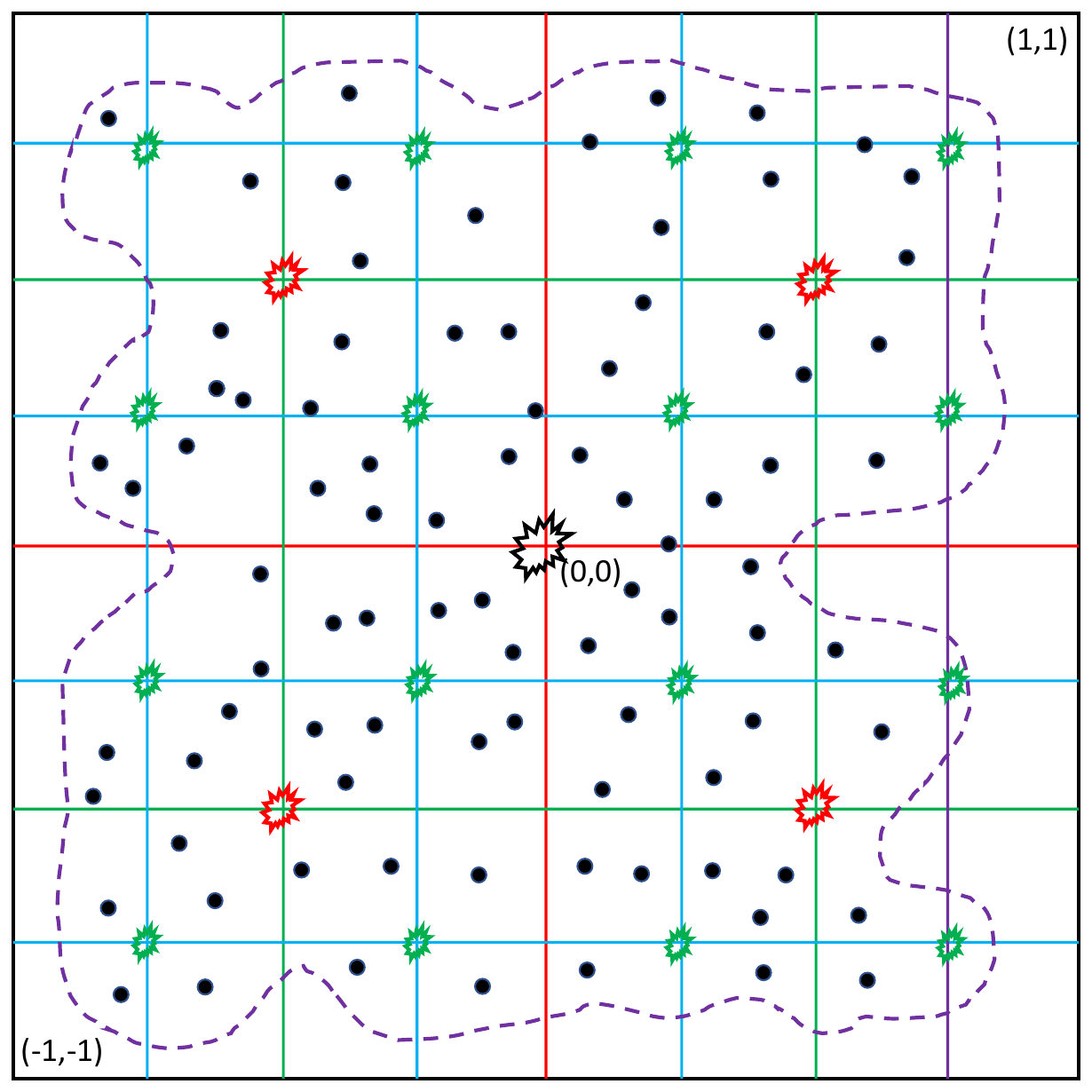}
\caption{\small An 2-d illustration for 3-d uniform and multi-scale electrostatic features for a protein (purple dashed line) with charges (black dots); Charges $q$ and reaction potential $\phi_\text{reac}$ are redistributed as point-multipoles (explosion symbols) using Cartesian treecode \cite{Li:2009} or FMM \cite{Tausch:2003} at the centers of the cluster at different levels (level 0: black; level 1: red; level 2: green).
}
\label{fig_e-features}
\end{center}
\end{figure}

In most molecular simulations, including Monte Carlo simulation,  Brownian Dynamics,  Molecular Dynamics, etc., the electrostatic interactions are characterized  by the partial charges assigned at the atomic centers. 
These partial charges are assigned according to the force fields, 
which are a set of mathematical equations and parameters 
used to calculate the potential energy of a molecule 
based on its atomic structure and interactions. 
The force fields are determined by experiment or quantum chemistry.

The algorithm for obtaining the mulit-scale, physics-informed, uniform electrostatic features 
can be explained using Fig.~\ref{fig_e-features}. 
The electrostatic profile of the protein is determined 
by the partial charges $q_n({\bf r}_n)$ or reaction potential $\phi_\text{reac}({\bf r}_n)$ 
at the atomic centers ${\bf r}_n$'s as illustrated by the black dots for $n=1,2,\cdots,N_c$ with $N_c$ as the number of atoms of the protein.  
Partial charges are the source of electrostatic interactions 
while reaction potentials are the outcomes of electrostatic interactions for the solvated potein, 
which includes the important solute-solvent interactions.
Consider the variation of $N_c$, the number of atoms, from protein to protein, 
it is difficult to directly use $q_n({\bf r}_n)$ and/or $\phi_\text{reac}({\bf r}_n)$ as features for machine learning. 
Our strategy to alleviate this pain is to use point-multipoles 
as illustrated by the explosion symbols at the cluster centers 
to carry redistributed charges or reaction potentials 
in terms of moments of the multipole expansion. 
The number of cluster centers are uniform for proteins with various numbers of atoms. 
It is also multiscale as being determined by the combination of levels of the tree $L$ and number of terms of the multipole expansion $p$. The users specify $L$ and $p$ as needed.

In Fig.~\ref{fig_e-features}. 
the $N_d$ point multipoles (explosion symbols) are organized in a hierarchical way 
such that different level of details are given 
by different number of symbols at different levels.  
For the $m$th multipole, $m=1,\cdots,N_d$, 
using a 3-d index ${\bf k} = (k_1, k_2, k_3)$, $|{\bf k}| = k_1+k_2+k_3$, in a cluster $c$ with cluster center ${\bf x}_c$, 
the moments are defined as
\begin{equation}
	\mathcal{M}_c^{\bf k} = \sum\limits_{{\bf x}_j \in c} q_j({\bf x}_j - {\bf x}_c)^{\bf k}.
\end{equation}
Note 
when $|{\bf k}| = 0$, ${\bf k} = (0,0,0)$; 
when $|{\bf k}| = 1$, ${\bf k} \in \{(1,0,0), (0,1,0), (1,0,0)\}$;
when $|{\bf k}| = 2$, ${\bf k} \in \{(1,1,0),(1,0,1),(0,1,1),(2,0,0),(0,2,0),(0,0,2)\}$, etc. 

If the Cartesian treecode algorithm \cite{Li:2009} is used, 
the computational cost is $O(N_c\log N_c)$ for obtaining all the moments as detailed in \cite{Li:2009}, 
where $N_c$ is number of the particles/charges and $\log N_c$ is the level of trees. 
In fact, the computational cost can be as low as $O(N_c)$ using the strategies in \cite{Tausch:2003} such that the moments at the finest cluster are computed first and a M2M (moments to moments) transformation can be used to efficiently compute moments at any desired level from the bottom to the top. Our simulations so far showed that the $O(N_c\log N_c)$ cost is sufficiently fast for moments computation since $N_c$ is within the scale of tens of thousands even for large proteins.  
These moments are intrinsic properties of the cluster thus can serve as features for the protein, which carries simplified but important structural and charge information.  

The number of features up to level $L$ cluster (the largest box containing all particles is named as the level 0 cluster and each subdivision increases the level by 1) is 
\begin{equation}N_f(p,L) = N_p(1+8+8^2+\cdots + 8^L) = N_p\frac{8^{L+1}-1}{7}\end{equation}
where $N_p$ is the number of terms in multipole expansion with $N_p=1,4,10,20,35,56 \cdots$ as the sequence of 
tetrahedral numbers
when $p$th order multipole is used. 


\subsubsection{Implementation Details}
We provide the step-by-step procedure by which the electrostatic features are generated.  \\
\noindent
\underline{Step 1.} The protein structure (e.g. the PDB file) is obtained from publicly available repo e.g. the protein data bank \cite{Berman:2000}.
The protein is represented by its atomic locations and partial charges:  $q_{i}({\bf r}_i)$ for $i=1,\cdots,N_c$, e.g. the PQR file using the PDB2PQR software \cite{Dolinsky:2007}.\\
\underline{Step 2.} Two parameters $p$ and $L$ are given, where $p$ is the order in the Cartesian Taylor expansion and $L$ is the number of levels. With given $p$, we have the number of terms given as 
$N_p = \displaystyle {(p+1)(p+2)(p+3)}/{6}$.
The number of features up to level $L$ cluster is 
\begin{equation}N_f(p,L) = 
\frac{p(p+1)(p+2)(8^{L+1}-1)}{42}.\end{equation}
These $N_f$ numbers are ordered by level from 0 to $L$ 
and the coordinates of cluster centers in each level. 
This determines the dimension of our final feature vector $F \in \mathbb{R}^{N_f}$. \\
The following table gives how $N_f$ varies with the change of $p$ and $L$:
\begin{center}
\begin{tabular}{c|r|r|r|r|r}
\hline
$p$\textbackslash$L$ & 0 & 1 & 2 & 3 & 4 \\\hline
0 &1&9&73&585&4681 \\
1 &4&36&292&2340&18724 \\
2 &10&90&730&5850& 46810 \\
3 &20&180&1460&11700& 93620 \\
4 &35&315&2555&20475&163835 \\\hline
 \end{tabular}
\end{center}
\underline{Step 3.} The Cartesian treecode \cite{Li:2009} will be called and the moments for the multipoles will be output as the electrostatic features. These features can be used as part of the inputs for ML algorithms, which will be shown next.

\subsection{The Machine Learning Model}

\begin{figure}[htbp]
\begin{center}
\includegraphics[width=6.5in]{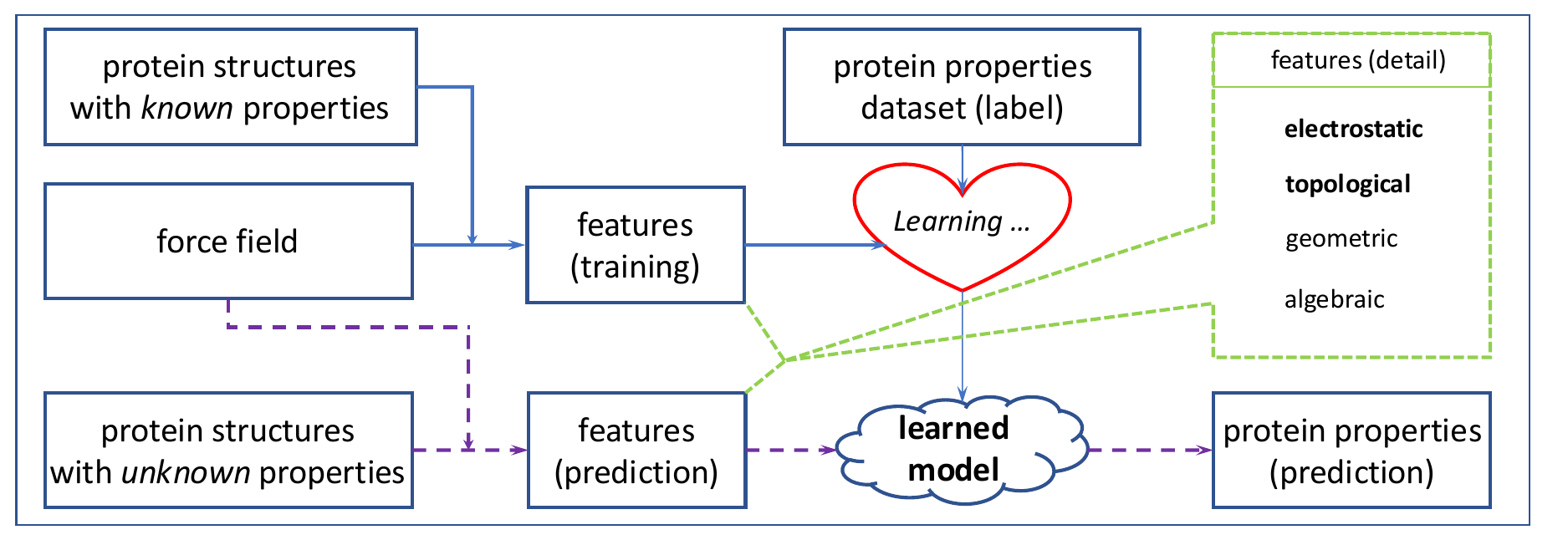}
\caption{\small The DNN based Machine Learning model using protein structures, force field, and known protein properties to mathematically generate electrostatic, topological, algebraic,  geometric, and features to train a learned model and then use the learned model to predict unknown properties for protein with available structures. This project focuses on electrostatic and topological features.}
\label{fig_MLmodel}
\end{center}
\end{figure}

The main goal of this project is to discover the hidden and useful information embedded 
in the protein structural data from the protein data bank in a simple and abstract way. 
The protein structure data and force field 
include bonded and non-bonded interactions with which 
the molecular simulation can be performed. 
We categorize these information into {\it electrostatic}, {\it topological}, {\it algebraic}, and {\it geometric} features. 
These four aspects are all important to the machine learning model. 
In this project, we put our focus on the electrostatic and topological features. 
The DNN based ML model is shown in Fig.~\ref{fig_MLmodel}, which uses the available protein structural data \cite{PDB}, force field such as AMBER \cite{Perlman:1995}, CHARMM\cite{Brooks:1983}, AMOEBA \cite{Zhang:2018}, etc. and known protein properties repositories  such as PDBbind Database \cite{Su:2019}, Protein pKa Database \cite{Pahari:2019}, etc to mathematically generate features to train a learned model and then use the learned model to predict unknown properties for protein with available structures.  In this project, we focus on the effectiveness of topological and electrostatic features using computed Coulomb energies and solvation  energies as labels. 

Mathematically, the surrogate models (Coulomb energies computed by treecode or solvation energies from the numerically solved PB model) 
\begin{equation}
    E=f(\mathbf{X},\mathbf{Q})
\end{equation}
are approximated by the machine learning models 
\begin{equation}
    \hat{E}=\hat{f}(F_e, F_t, \Theta)
\end{equation}
where $E$ and $\hat{E}$ are the energies, 
$\mathbf{X}$ and $\mathbf{Q}$ are the positions and charges of protein atoms, $\Theta$ are parameters in the ML model to be trained by data,  
$F_e = F_e(\mathbf{X},\mathbf{Q})$ and $F_t=F_t(\mathbf{X})$ are electrostatic features and topological features and their generators, which are non-trivial mappings from protein structure and electrostatics to features.

\subsection{Specification of the ML Model}
\begin{figure}[htbp]
\begin{center}
\includegraphics[width=\linewidth]{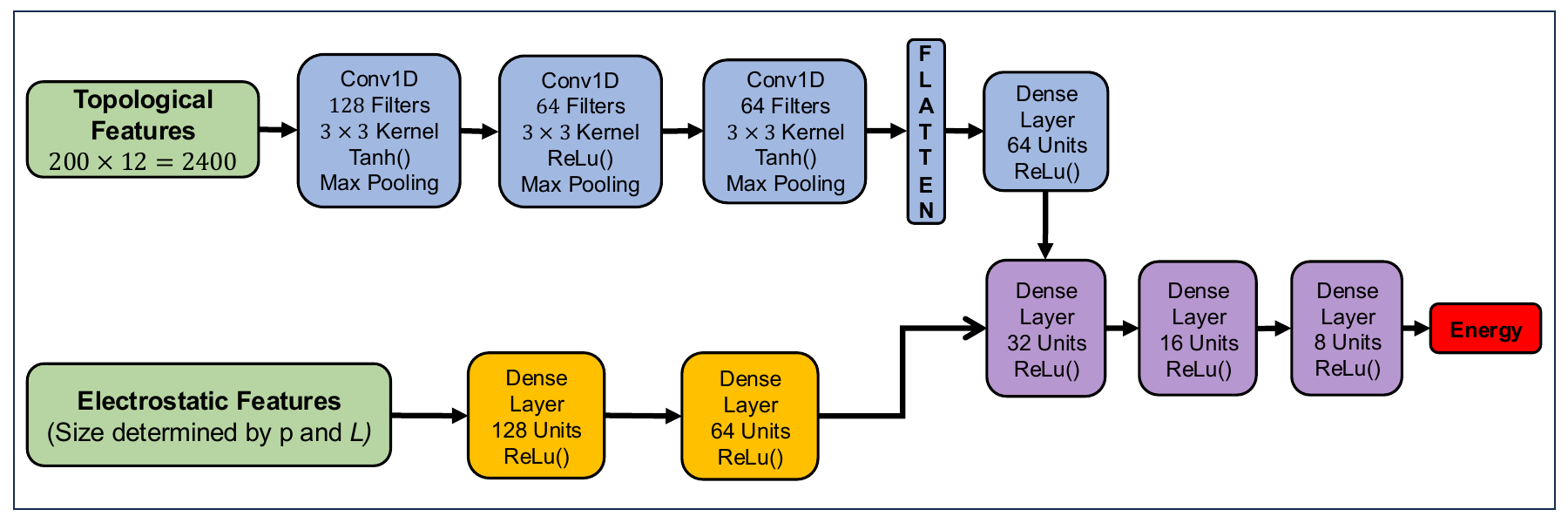}
\caption{\small The architecture of the  deep neural network with two branches taking in topological features and electrostatic features. The outputs of the two branches are concatenated into one deep neural network.}
\label{fig_network}
\end{center}
\end{figure}

The model developed in this study is to
to predict Coulomb and solvation energies using combined topological and electrostatic features. 
All target labels were standardized to zero mean and unit variance, 
and predictions were inverse-transformed to recover physical units (kcal/mol) for evaluation.
As illustrated in Fig.~\ref{fig_network}, the proposed architecture consists of two branches. 
The first branch, shown in blue, is a one-dimensional convolutional neural network (CNN) designed to process the topological features. 
These features are passed through three consecutive convolution–pooling blocks (with 128, 64, and 64 filters; kernel size = 3; and activation functions tanh, ReLU, and tanh, respectively), 
followed by a fully connected layer with 64 units. 
The electrostatic features are processed through two fully connected hidden layers with 128 and 64 units, respectively (shown in orange). 
After independent processing in these two branches, 
their outputs are concatenated and passed through three additional fully connected layers with 32, 16, and 8 units (shown in purple). 
The final dense layer then produces the regression output for the predicted energy (shown in red).
To improve generalization, dropout, $l_2$ regularization, and batch normalization were applied. 
The model was trained using the Adam optimizer (learning rate $1.0 \times 10^{-4}$) with mean squared error as the loss function. 
The dataset was randomly divided into 80\% training and 20\% testing subsets. 
Model selection was performed using five-fold cross-validation on the training set, 
and the best-validated model was subsequently evaluated on the held-out test set. 
The model architecture and hyperparameters were selected based on empirical considerations and adopted from previously published studies \cite{Cang:2017}. 
The complete hyperparameter configuration is provided in Table~$S1$ of the Supporting Information.

\subsection{Data}
The datasets used in this study are sourced from the PDBbind database \cite{Liu:2014}, a widely used benchmark for protein–ligand complexes. We categorize the data into two datasets as described below. For all structures, atomic van der Waals radii and partial charges are assigned using the AMBER ff14SB force field through the PDB2PQR software \cite{Dolinsky:2007}.

\noindent
\textbf{Dataset 1.}
{
This dataset is derived from the refined set of the PDBbind v2018 release \cite{Liu:2014}, which contains 4,463 protein structures. Outliers in the Coulombic energy $E_{\mathrm{coul}}$ and solvation free energy $E_{\mathrm{solv}}$ are removed using the interquartile range (IQR) method, in which samples falling outside the interval $[Q_1 - 1.5\times \text{IQR},\; Q_3 + 1.5\times \text{IQR}]$ are excluded. For $E_{\mathrm{coul}}$, the filtering  thresholds are $[-817.87,\; 170.83]$, where $Q_1 = -447.11$~kcal/mol, $Q_3 = -199.94$~kcal/mol, $\text{IQR} = 247.17$~kcal/mol. For $E_{\mathrm{solv}}$, the thresholds are $[-10,691.76 ,\; 2,210.83]$, where $Q_1 =-5,853.29$~kcal/mol, $Q_3 = -2,627.64$~kcal/mol, $\text{IQR} = 3,225.65$~kcal/mol. After this filtering step, 4,043 and 4,088 protein structures are retained for $E_{\mathrm{coul}}$ and $E_{\mathrm{solv}}$, respectively, and used for model training and evaluation. All $E_{\mathrm{solv}}$ values are computed using the second-order accurate Matched Interface and Boundary Poisson–Boltzmann (MIBPB) solver~\cite{Chen:2011} with a mesh spacing of $h = 0.2~\text{\AA}$.
}

\noindent
\textbf{Dataset 2.}
{
This dataset is constructed from the PDBbind v2020 release, comprising a general set of 14,127 complexes and a refined set of 5,316 complexes. Coulombic energies $E_{\mathrm{coul}}$ are computed for all proteins using a Cartesian treecode algorithm~\cite{Li:2009}. To ensure data quality, outliers are identified and removed using the IQR method, applying thresholds of $[-1,164.23,\; 338.04]$~kcal/mol, where $Q_1 = -600.88$~kcal/mol,\; $Q_3 = -225.31$~kcal/mol, $\text{IQR} = 375.57$~kcal/mol. Following this filtering step, a total of 17,007 protein structures are retained for subsequent model training and evaluation.
}

\begin{figure}[htbp!]
\begin{center}
\includegraphics[width=2.1in]{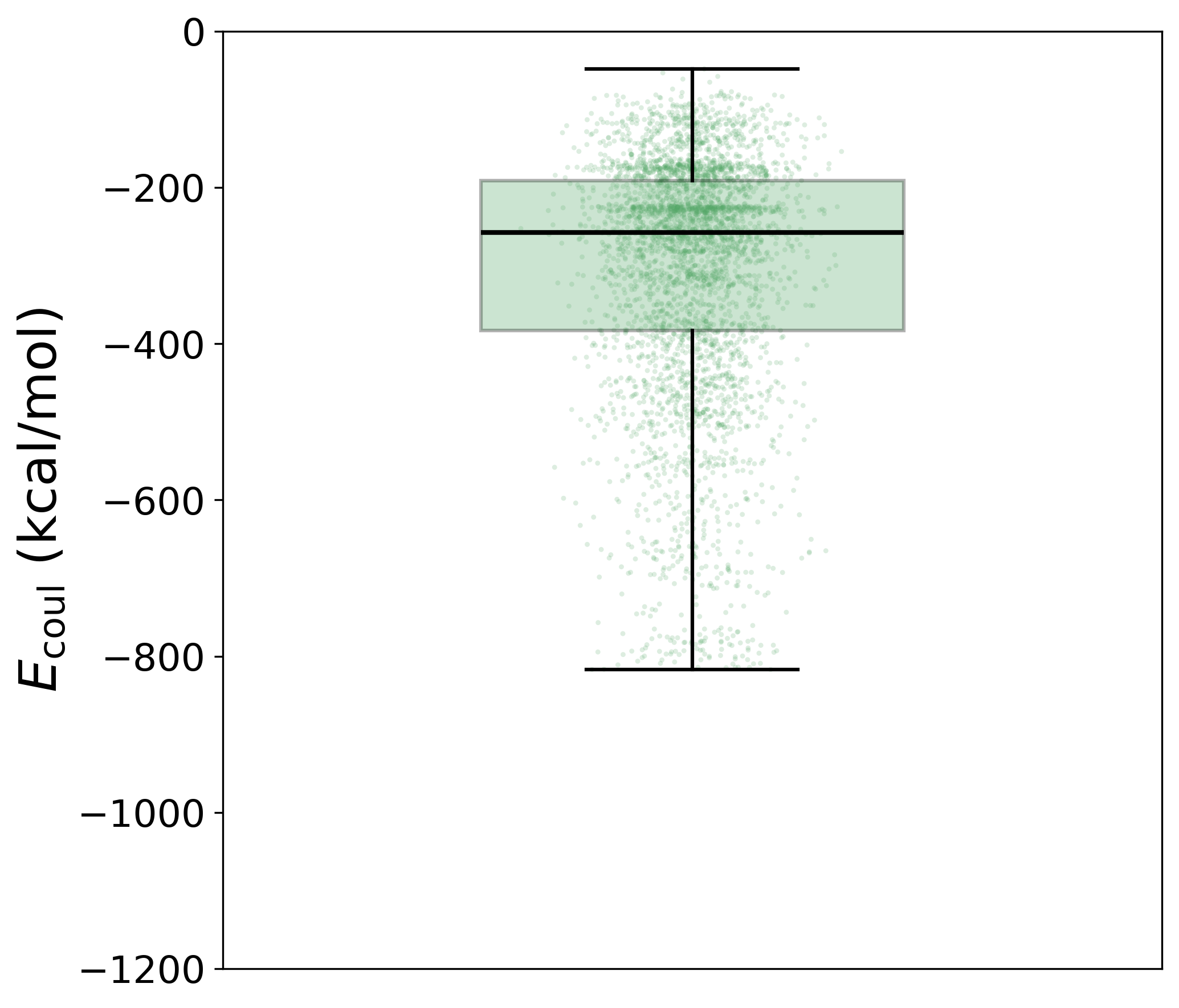}
\includegraphics[width=2.1in]{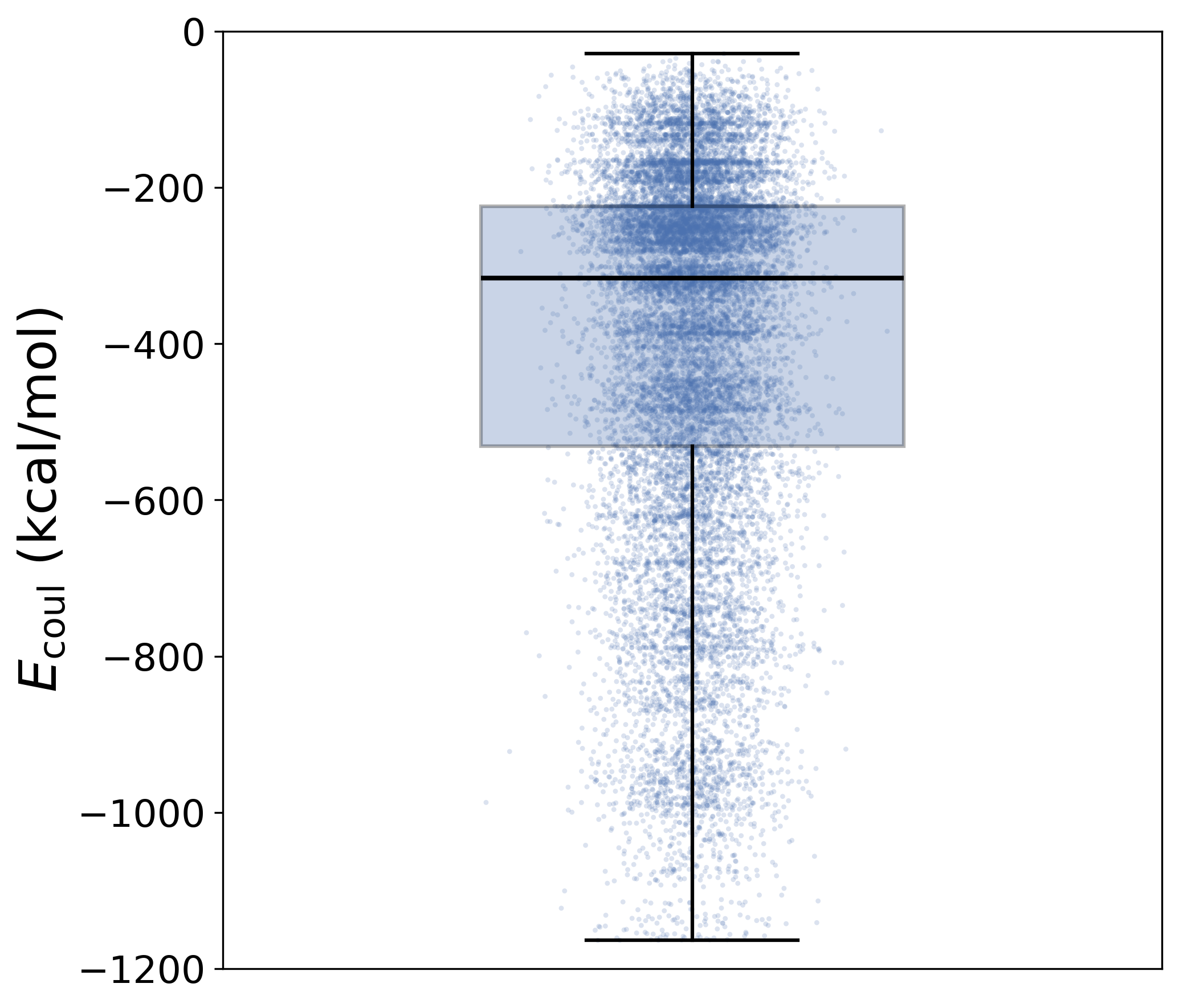}
\includegraphics[width=2.1in]{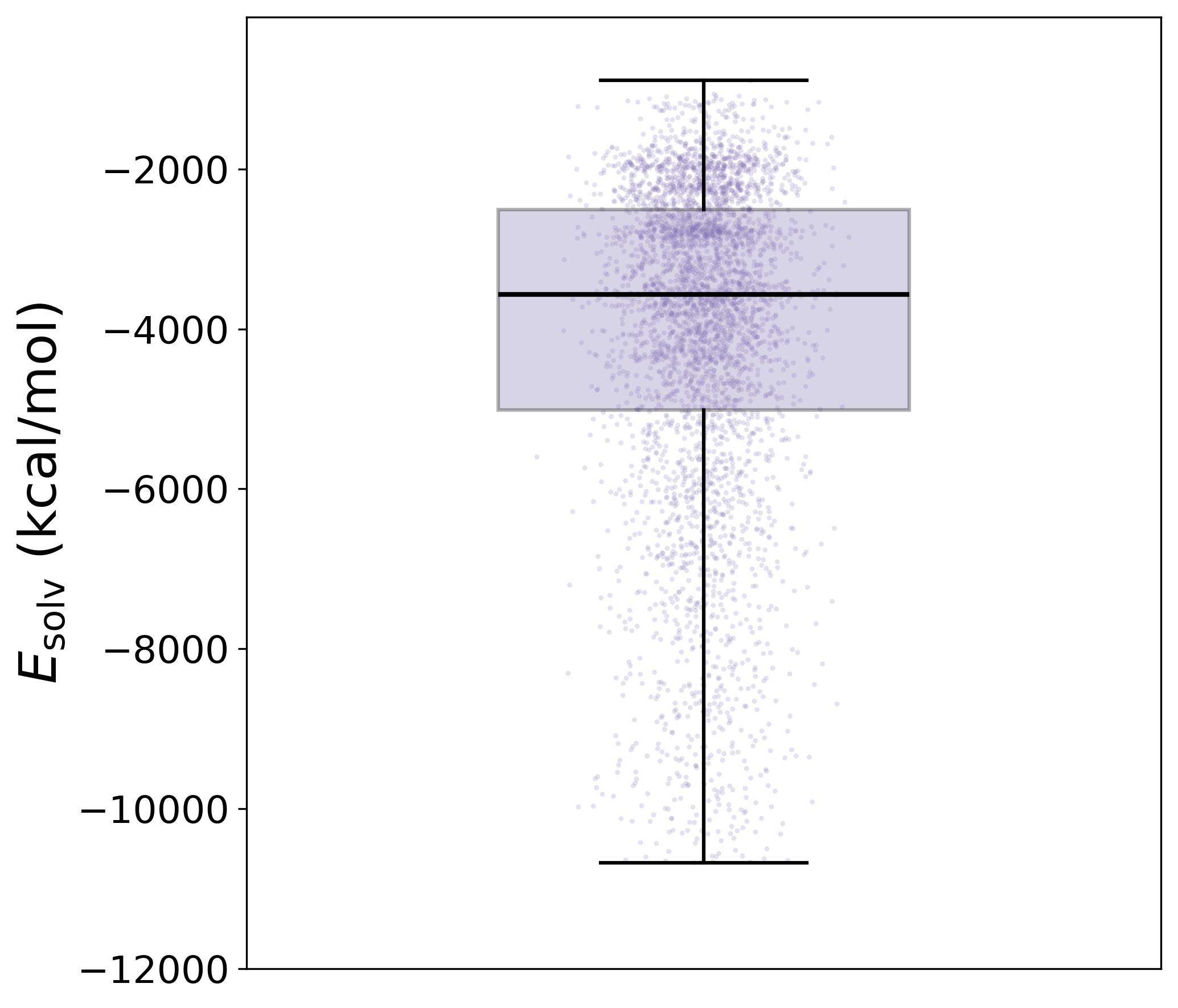}\\
~~~~~~~~~(a) \hskip 1.9in (b) \hskip 1.9in (c)
\caption{\small
Box-and-whisker plots showing the distributions of Coulombic energy $E_{\mathrm{coul}}$ for Dataset~1 in (a) and Dataset~2 in (b), and solvation energy $E_{\mathrm{solv}}$ for Dataset~1 in (c). Jittered points represent individual protein samples.
}
\label{fig_jitter_box}
\end{center}
\end{figure}

Figure~\ref{fig_jitter_box} illustrates the distributions of Coulombic and solvation energies across the two datasets. The Coulombic energies in both datasets are concentrated within a relatively narrow negative range, whereas the solvation energies exhibit a substantially broader distribution with greater dispersion. This contrast reflects the fundamentally different physical scales and variability associated with Coulombic interactions and solvation effects.

\subsection{Research Purposes and Metrics}
We provide simulation results to evaluate the performance of the proposed model. These results are designed to address the following three research purposes:
\\
\textbf{Purpose 1:} Training on a larger dataset leads to improved predictive performance.
\\
\textbf{Purpose 2:} When electrostatic features are incorporated, model performance improves as the treecode level $L$, Taylor expansion order $p$, and the corresponding number of features $N_f$ increase.
\\
\textbf{Purpose 3:} Combining electrostatic and topological features outperforms models that use only one type of feature.

\noindent
{\bf Metrics:} 
We use three common metrics to measure the performance of the model, as defined below. 
In all these metrics, we use
$y_i$ to denote actual value, 
$\hat{y}_i$ for predicted value,
$\bar{y}$ for mean of actual values,
$\bar{\hat{y}}$ for mean of predicted values. \\

\noindent
MAPE: The mean average percentage errors, which measures prediction accuracy as a percentage. 
This metric is not suitable if the actual values contain zero or values close to zero.
\begin{equation}\text{MAPE} = \frac{1}{n} \sum_{i=1}^n \left| \frac{y_i - \hat{y}_i}{y_i} \right| \times 100\%\end{equation}

\noindent
MSE: The mean-squared error, which measures the average squared difference between predicted and actual values.  
\begin{equation}\text{MSE} = \frac{1}{n} \sum_{i=1}^n (y_i - \hat{y}_i)^2\end{equation}

\noindent
$R^2$, the quantity indicating the proportion of variance in the dependent variable that is predictable from the independent variables. 
\begin{equation}R^2 = 1 - \frac{\text{SS}_{\text{res}}}{\text{SS}_{\text{tot}}}\end{equation}
where
$\text{SS}_{\text{tot}} = \sum_{i=1}^n (y_i - \bar{y})^2$ measures the total variance in the observed data 
and 
$\text{SS}_{\text{res}} = \sum_{i=1}^n (y_i - \hat{y}_i)^2$ measures the variance not explained by the model.

\section{Results}
We next provide simulation results using different datasets and models. Following that, additional comparisons with baseline models and different grouping strategies are provided to justify the learned model.

\subsection{Model Performance}
 In reporting and explaining the performance of each combination of model and dataset, we focus on achieving the three aforementioned purposes.

In describing the PB and GB models, we use the popular $e_c$/\AA~units for potential and $e_c^2$/\AA~units for energy. By multiplying the conversion factor 332.0716, the units of the potential and energy ($E_\text{coul}$ and $E_\text{solv}$) become kcal/mol/$e_c$ and kcal/mol respectively, which are the units reported for all results.

\noindent
\subsubsection{$E_\text{coul}$ predicted using electrostatic features only on Dataset 1 and Dataset 2} 

We first investigate the effectiveness of electrostatic features in predicting the Coulombic energy $E_\text{coul}$. Note that $E_\text{coul}$ is determined by the locations and charges of the particles, which also determine the moments of the multipoles as electrostatic features.

\begin{table}[h]
\caption{\small $E_\text{coul}$ (kcal/mol) prediction results using electrostatic features only on Dataset~1 and Dataset~2. The D2--D1 columns report the differences (Dataset~2 $-$ Dataset~1) for the corresponding metrics.}
\label{tb_ce_d1d2}
\begin{center}
{\small
\begin{tabular}{c|c|r||r|r|r||r|r|r||r|r|r}
\hline
$p$ & $L$ & $N_f(p,L)$
& \multicolumn{3}{c||}{Dataset~1}
& \multicolumn{3}{c||}{Dataset~2}
& \multicolumn{3}{c}{D2--D1} \\
\hline
& & 
& MSE & MAPE & $R^2$
& MSE & MAPE & $R^2$
& MSE & MAPE & $R^2$ \\
\hline
1 & 0 & 4     & 0.584 & 0.337 & 0.385 & 0.682 & 0.531 & 0.323 & 0.098 & 0.194 & -0.062 \\
2 & 0 & 10    & 0.353 & 0.246 & 0.628 & 0.340 & 0.331 & 0.663 & -0.013 & 0.085 & 0.035 \\
3 & 0 & 20    & 0.254 & 0.207 & 0.733 & 0.236 & 0.284 & 0.767 & -0.017 & 0.077 & 0.035 \\
4 & 0 & 35    & 0.325 & 0.186 & 0.668 & 0.249 & 0.321 & 0.753 & -0.076 & 0.135 & 0.085 \\
\hline
0 & 1 & 9     & 0.708 & 0.344 & 0.254 & 0.633 & 0.498 & 0.372 & -0.075 & 0.154 & 0.118 \\
1 & 1 & 36    & 0.256 & 0.205 & 0.730 & 0.241 & 0.273 & 0.761 & -0.015 & 0.068 & 0.030 \\
2 & 1 & 90    & 0.219 & 0.177 & 0.769 & 0.156 & 0.230 & 0.845 & -0.063 & 0.053 & 0.076 \\
3 & 1 & 180   & \textbf{0.169} & \textbf{0.155} & \textbf{0.822} & 0.163 & 0.237 & 0.838 & -0.006 & 0.083 & 0.016 \\
4 & 1 & 315   & 0.185 & 0.173 & 0.805 & 0.165 & 0.234 & 0.837 & -0.020 & 0.061 & 0.032 \\
\hline
0 & 2 & 73    & 0.403 & 0.278 & 0.575 & 0.322 & 0.324 & 0.681 & -0.081 & 0.047 & 0.105 \\
1 & 2 & 292   & 0.234 & 0.223 & 0.754 & 0.127 & 0.206 & 0.874 & -0.106 & -0.017 & 0.120 \\
2 & 2 & 730   & 0.204 & 0.187 & 0.786 & 0.133 & 0.208 & 0.868 & -0.070 & 0.021 & 0.082 \\
3 & 2 & 1460  & 0.182 & 0.163 & 0.808 & \textbf{0.123} & \textbf{0.195} & \textbf{0.880} & -0.059 & 0.032 & 0.071 \\
4 & 2 & 2555  & 0.247 & 0.225 & 0.740 & 0.145 & 0.229 & 0.857 & -0.102 & 0.004 & 0.118 \\
\hline
0 & 3 & 585   & 0.393 & 0.239 & 0.586 & 0.265 & 0.279 & 0.737 & -0.128 & 0.041 & 0.151 \\
1 & 3 & 2340  & 0.443 & 0.211 & 0.533 & 0.253 & 0.236 & 0.752 & -0.190 & 0.024 & 0.219 \\
2 & 3 & 5850  & 0.411 & 0.206 & 0.567 & 0.217 & 0.252 & 0.784 & -0.194 & 0.046 & 0.218 \\
3 & 3 & 11700 & 0.472 & 0.249 & 0.503 & 0.209 & 0.261 & 0.792 & -0.262 & 0.012 & 0.290 \\
4 & 3 & 20475 & 0.381 & 0.240 & 0.598 & 0.223 & 0.267 & 0.779 & -0.158 & 0.027 & 0.180 \\
\hline
\end{tabular}
}
\end{center}
\end{table}

\begin{figure}[h!]
\begin{center}
\includegraphics[width=2.12in, height=2.05in]{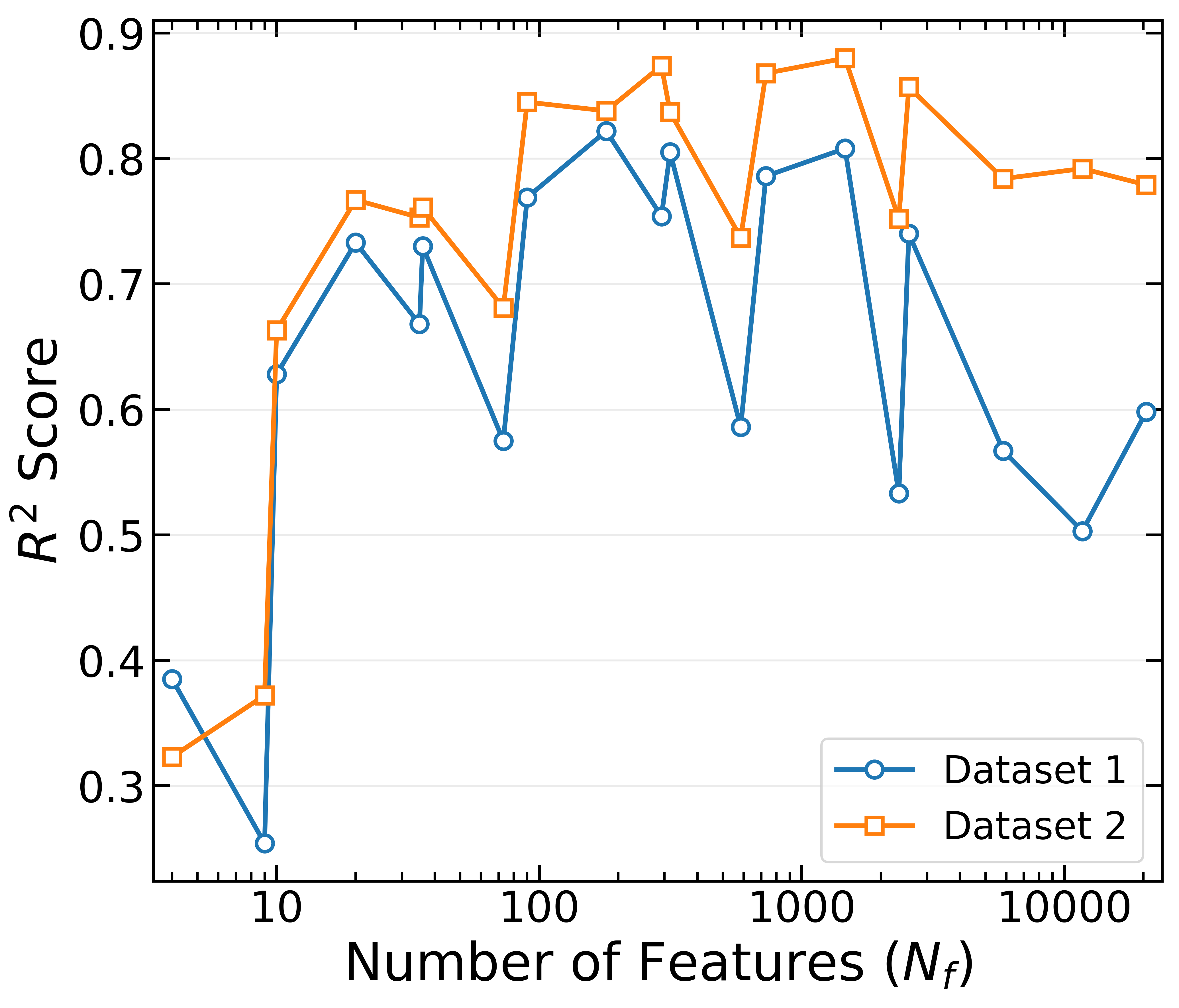}
\includegraphics[width=2.14in, height=2.05in]{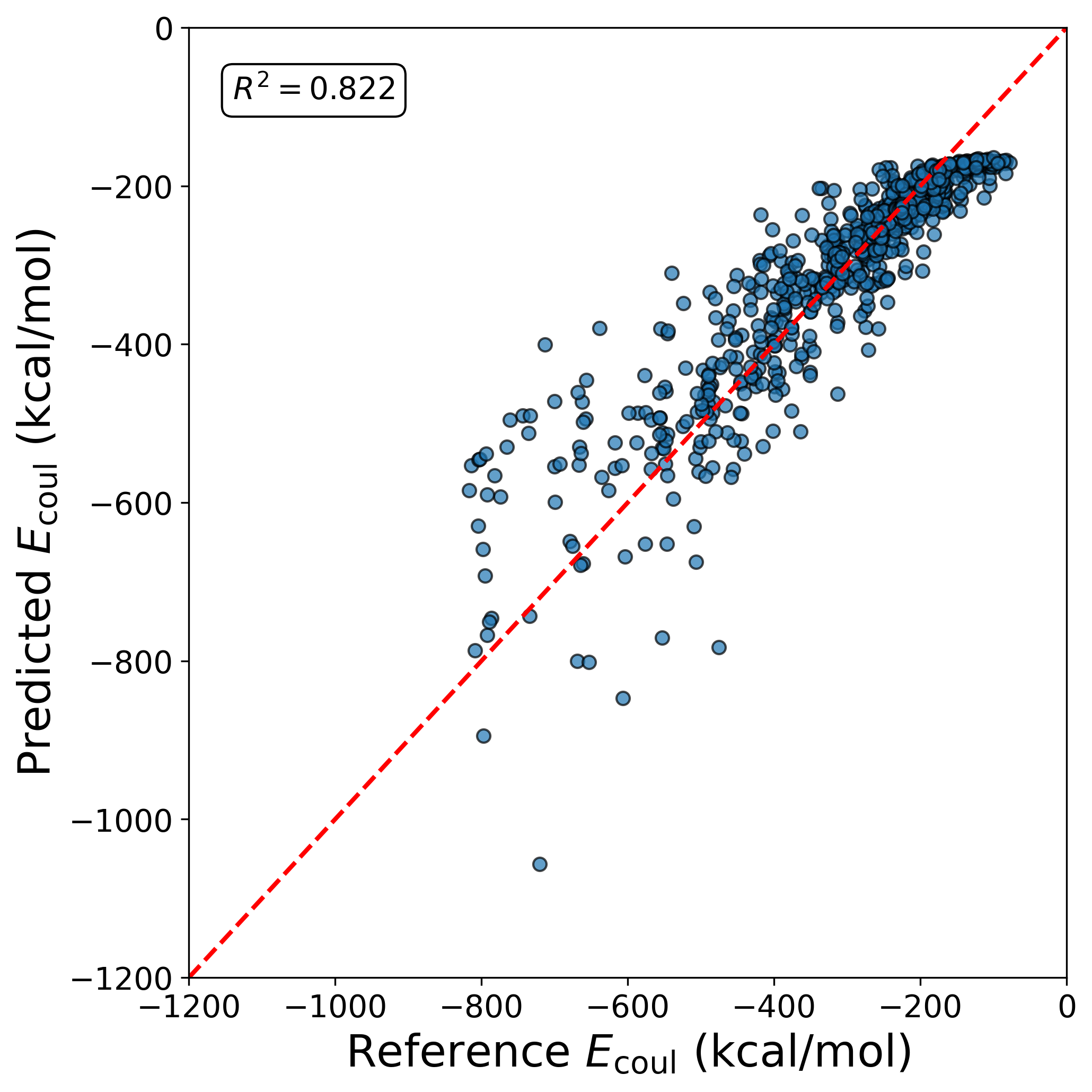}
\includegraphics[width=2.14in, height=2.05in]{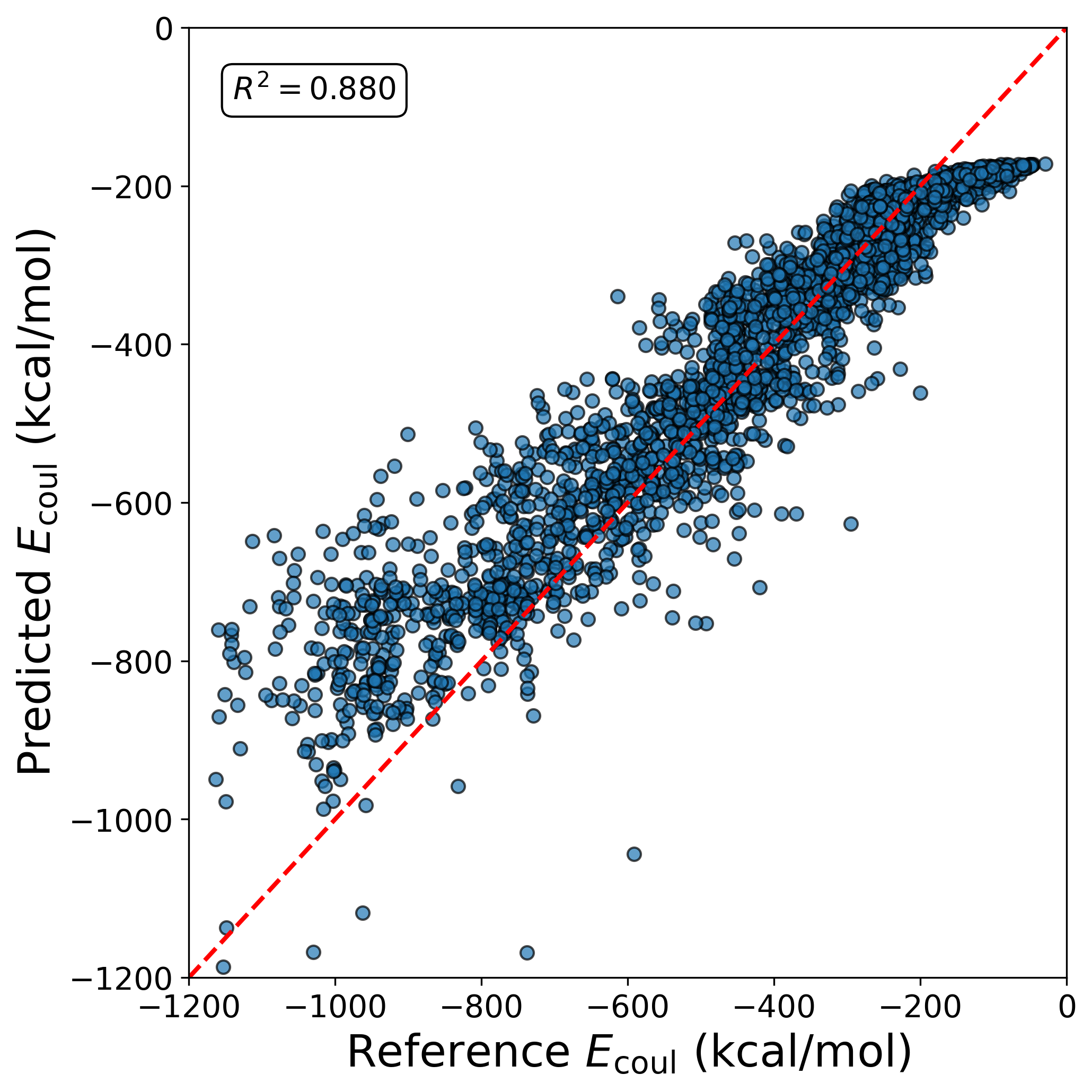}\\
(a) \hskip 2in (b) \hskip 2in (c)

\caption{\small  
(a) $R^2$ of model performance for models trained on the two datasets against number of features;
(b) scatter plot for Dataset 1 ($p=3$, $L=1$ with total 180 features);
(c) scatter plot for Dataset 2 ($p=3$, $L=2$ with total 1460 features)
}
\label{fig_5}
\end{center}
\end{figure}

Table~\ref{tb_ce_d1d2} reports the $E_\text{coul}$ predictions obtained using electrostatic features only on Dataset~1 and Dataset~2. The first important pattern observed from the table is that  the MSE and $R^2$ obtained from Dataset~2 are generally better than those from Dataset~1, as the differences in MSE (D2-D1) are almost all negative, while the difference in $R^2$ (D2-D1) is almost all positive. This pattern validates {\bf Purpose 2}, indicating the advantage of using a larger dataset, particularly in capturing more overall variance in the target. We also notice that the differences in MAPE(D2-D1) are almost all positive, apparently contradicting the trends observed for MSE and $R^2$. In fact, this behavior provides additional insight into the data distribution: Dataset~2, but not Dataset~1,
contains more samples with smaller target values as shown in Fig.~\ref{fig_jitter_box}, leading to larger percentage errors.

From Table~\ref{tb_ce_d1d2}, we also observe that in Dataset~1, the MSE decreases and $R^2$ increases as $p$ and $L$ increase, reaching an optimal value at $p=3$ and $L=1$ with 180 features. Similarly, in Dataset~2, the MSE decreases and $R^2$ increases with increasing $p$ and $L$, reaching an optimal value at $p=3$ and $L=2$ with 1460 features. This pattern validates {\bf Purpose~1}, demonstrating the advantage of using more features. However, we should be aware that the optimal choices of $p$ and $L$ are influenced by the size of the training dataset. When the number of features exceeds the number of samples, increasing the number of features may have a negative impact, such as overfitting.  

We plot the $R^2$ values in Fig.~\ref{fig_5}(a), where the horizontal axis represents the number of features $N_f$ given in the third column of Table~\ref{tb_ce_d1d2}. The validation of {\bf Purpose~1} is evident, as the blue squares are consistently lower than the red ones. The validation of {\bf Purpose~2} is also observed, as markers of the same color exhibit an increasing trend with the number of features until the feature set becomes excessively large. We also plot the best-performing cases for both datasets in Fig.~\ref{fig_5}(b) and (c), where the reference-predicted values are distributed closely around the $\hat{y} = y$ line.

\noindent
\subsubsection{$E_\text{coul}$ predicted using e-features and both features on Dataset 2}

\begin{table}[h]
\caption{$E_\text{coul}$ (kcal/mol) prediction results using electrostatic features and combined features on Dataset~2. The corresponding MSE, MAPE, and $R^2$ values obtained using Topological Features only are \textbf{0.030, 0.071,} and \textbf{0.970}, respectively.}
\label{tb_ce_both}
\begin{center}
{\small
\begin{tabular}{c|c|r||r|r|r||r|r|r||r|r|r}
\hline
$p$ & $L$ & $N_f(p,L)$ 
& \multicolumn{3}{c||}{e-features only} 
& \multicolumn{3}{c||}{both features} 
& \multicolumn{3}{c}{difference} \\
\hline
 & & 
& MSE & MAPE & $R^2$ 
& MSE & MAPE & $R^2$ 
& MSE & MAPE & $R^2$ \\
\hline
1 & 0 & 4     & 0.682 & 0.531 & 0.323 & 0.028 & 0.076 & 0.972 & -0.654 & -0.454 & 0.649 \\
2 & 0 & 10    & 0.340 & 0.331 & 0.663 & 0.035 & 0.072 & 0.966 & -0.305 & -0.258 & 0.303 \\
3 & 0 & 20    & 0.236 & 0.284 & 0.767 & 0.031 & 0.064 & 0.969 & -0.205 & -0.220 & 0.202 \\
4 & 0 & 35    & 0.249 & 0.321 & 0.753 & 0.027 & 0.082 & 0.973 & -0.222 & -0.239 & 0.221 \\
\hline
0 & 1 & 9     & 0.633 & 0.498 & 0.372 & 0.029 & 0.067 & 0.971 & -0.604 & -0.431 & 0.599 \\
1 & 1 & 36    & 0.241 & 0.273 & 0.761 & 0.030 & 0.068 & 0.971 & -0.211 & -0.205 & 0.210 \\
2 & 1 & 90    & 0.156 & 0.230 & 0.845 & 0.029 & 0.063 & 0.971 & -0.127 & -0.167 & 0.126 \\
3 & 1 & 180   & 0.163 & 0.237 & 0.838 & 0.029 & 0.067 & 0.971 & -0.134 & -0.170 & 0.133 \\
4 & 1 & 315   & 0.165 & 0.234 & 0.837 & \textbf{0.024} & \textbf{0.073} & \textbf{0.976} & -0.141 & -0.162 & 0.140 \\
\hline
0 & 2 & 73    & 0.322 & 0.324 & 0.681 & 0.031 & 0.069 & 0.969 & -0.291 & -0.255 & 0.289 \\
1 & 2 & 292   & 0.127 & 0.206 & 0.874 & 0.035 & 0.131 & 0.965 & -0.092 & -0.075 & 0.092 \\
2 & 2 & 730   & 0.133 & 0.208 & 0.868 & 0.033 & 0.109 & 0.968 & -0.101 & -0.099 & 0.100 \\
3 & 2 & 1460  & \textbf{0.123} & \textbf{0.195} & \textbf{0.880} & 0.033 & 0.092 & 0.967 & -0.089 & -0.103 & 0.087 \\
4 & 2 & 2555  & 0.145 & 0.229 & 0.857 & 0.026 & 0.067 & 0.974 & -0.119 & -0.162 & 0.117 \\
\hline
0 & 3 & 585   & 0.265 & 0.279 & 0.737 & 0.031 & 0.071 & 0.969 & -0.234 & -0.209 & 0.232 \\
1 & 3 & 2340  & 0.253 & 0.236 & 0.752 & 0.034 & 0.077 & 0.967 & -0.219 & -0.159 & 0.215 \\
2 & 3 & 5850  & 0.217 & 0.252 & 0.784 & 0.031 & 0.075 & 0.969 & -0.186 & -0.177 & 0.185 \\
3 & 3 & 11700 & 0.209 & 0.261 & 0.792 & 0.029 & 0.083 & 0.971 & -0.181 & -0.178 & 0.179 \\
4 & 3 & 20475 & 0.223 & 0.267 & 0.779 & 0.023 & 0.074 & 0.978 & -0.200 & -0.193 & 0.199 \\
\hline
\end{tabular}
}
\end{center}
\end{table}

We next investigate the effectiveness of the combined feature representation. This analysis consists of two parts. In the first part, $E_\text{coul}$ is used as the label on Dataset~2, while in the second part, $E_\text{solv}$ is used as the label on Dataset~1, taking into account the size of the datasets and the availability of labels.

\begin{figure}[htbp!]
\begin{center}
\includegraphics[width=2.1in]{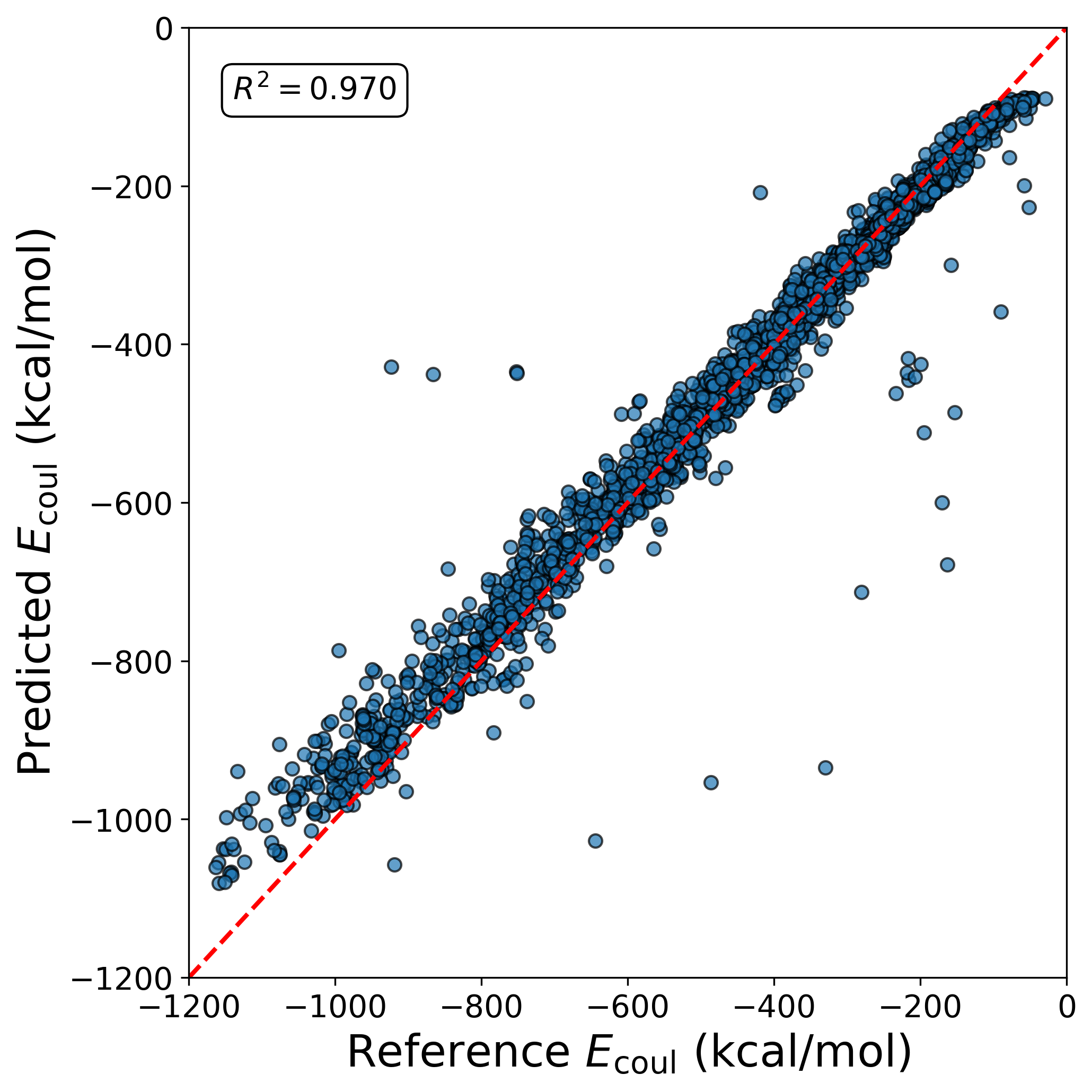}
\includegraphics[width=2.1in]{figures/D2_CE_scatter_electro_p3_L2.png}
\includegraphics[width=2.1in]{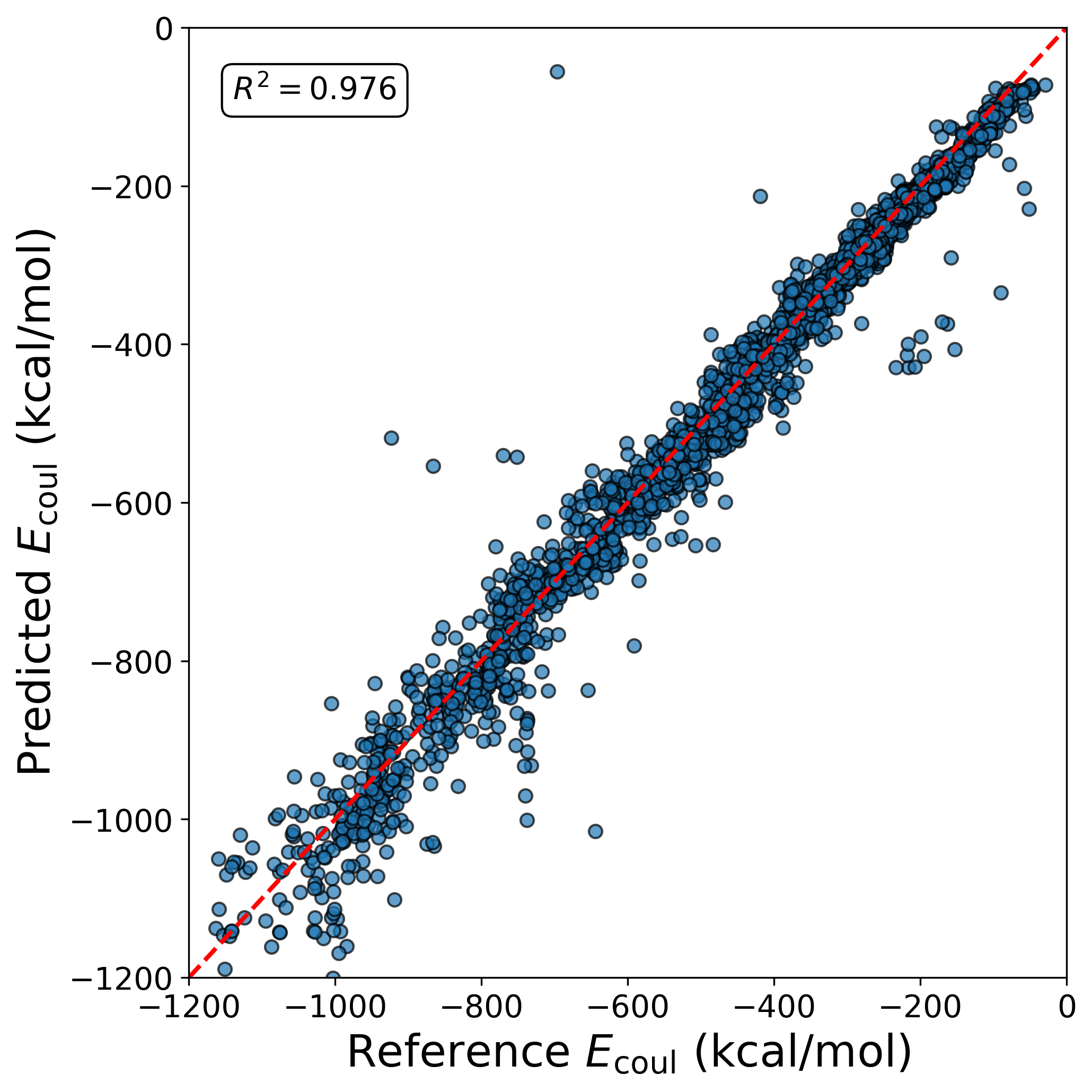}\\
(a) \hskip 2in (b) \hskip 2in (c)
\caption{\small  Scatter plots for the prediction of $E_\text{coul}$ on Dataset 2: (a) topological features only, (b) e-features only ($p=3$, $L=2$ with total 1460 features), (c) both features ($p=4$, $L=1$ with total 315 features). 
}
\label{fig_6}
\end{center}
\end{figure}


The MSE, MAPE, and $R^2$ values for the different models are reported in Table~\ref{tb_ce_both}. Using topological features alone yields MSE, MAPE, and $R^2$ values of 0.030, 0.071, and 0.970, respectively, indicating strong predictive accuracy. Moreover, the difference column clearly shows that incorporating topological features into the electrostatics-only model consistently reduces MSE and MAPE while increasing $R^2$. These results partially support {\bf Purpose~2}.

If we consider topological features first, it appears that adding electrostatic features does not significantly improve the results. This is because the performance obtained using topological features is already very strong, leaving limited room for the effectiveness of electrostatic features to manifest. Our next test case shows that when the topological features are relatively less effective, adding electrostatic features can significantly improve performance. Before moving to that case, we briefly examine Fig.~\ref{fig_6}, which shows scatter plots of predicted values versus true values using topological features, electrostatic features, and combined features. The improvement from (b) to (c) is very evident, while no obvious improvement is observed from (a) to (c). \\

\subsubsection{$E_\text{solv}$ predicted using e-features only and both features  on Dataset 1}

As mentioned earlier, computing electrostatic solvation energies accurately is time consuming. Here, we use the MIBPB solver \cite{Chen:2011} to compute the solvation energies of proteins in Dataset~1, which are then used as labels for model training and testing.  A similar simulation using Dataset~1 is conducted, and the results are reported in Table~\ref{tb_se_d1}. From the table, we can see that the improvement obtained by using both features, compared to using only topological features or only electrostatic features, is significant. If we focus on the difference column, we see that adding topological features clearly improves the electrostatics-only model by reducing the MSE and MAPE while increasing the $R^2$ score. In addition, the results obtained using both features are noticeably better than those obtained using topological features alone, which yield MSE = 0.150, MAPE = 0.153, and $R^2$ = 0.826. These results validate {\bf purpose~3}, and demonstrate the advantage of combining topological and electrostatic features.

\begin{table}[h]
\caption{\small $E_\text{solv}$(kcal/mol) computed using e-features and both features on Dataset~1; the MSE, MAPE and $R^2$ for using Topological Features only is \textbf{0.150, 0.153} and \textbf{0.826} respectively.}
\label{tb_se_d1}
\begin{center}
{\small
\begin{tabular}{c|c|r||r|r|r||r|r|r||r|r|r}
\hline
$p$ & $L$ & $N_f(p,L)$ 
& \multicolumn{3}{c||}{e-features only} 
& \multicolumn{3}{c||}{both features} 
& \multicolumn{3}{c}{difference} \\
\hline
 & & 
& MSE & MAPE & $R^2$ 
& MSE & MAPE & $R^2$ 
& MSE & MAPE & $R^2$ \\
\hline
1 & 0 & 4     & 0.345 & 0.243 & 0.604 & 0.063 & 0.108 & 0.926 & -0.282 & -0.136 & 0.323 \\
2 & 0 & 10    & 0.271 & 0.256 & 0.690 & 0.090 & 0.094 & 0.895 & -0.180 & -0.162 & 0.206 \\
3 & 0 & 20    & 0.135 & 0.142 & 0.845 & \textbf{0.064} & \textbf{0.081} & \textbf{0.926} & -0.071 & -0.061 & 0.081 \\
4 & 0 & 35    & 0.121 & 0.138 & 0.861 & 0.065 & 0.083 & 0.925 & -0.056 & -0.055 & 0.064 \\
\hline
0 & 1 & 9     & 0.447 & 0.318 & 0.487 & 0.114 & 0.169 & 0.867 & -0.333 & -0.149 & 0.380 \\
1 & 1 & 36    & 0.148 & 0.161 & 0.830 & 0.077 & 0.104 & 0.911 & -0.072 & -0.057 & 0.081 \\
2 & 1 & 90    & \textbf{0.095} & \textbf{0.130} & \textbf{0.891} & 0.084 & 0.098 & 0.903 & -0.012 & -0.032 & 0.012 \\
3 & 1 & 180   & 0.111 & 0.138 & 0.873 & 0.105 & 0.103 & 0.878 & -0.006 & -0.036 & 0.005 \\
4 & 1 & 315   & 0.128 & 0.144 & 0.853 & 0.097 & 0.105 & 0.887 & -0.031 & -0.039 & 0.034 \\
\hline
0 & 2 & 73    & 0.214 & 0.205 & 0.754 & 0.092 & 0.116 & 0.892 & -0.122 & -0.089 & 0.138 \\
1 & 2 & 292   & 0.169 & 0.156 & 0.806 & 0.127 & 0.123 & 0.852 & -0.041 & -0.033 & 0.045 \\
2 & 2 & 730   & 0.164 & 0.156 & 0.812 & 0.128 & 0.116 & 0.851 & -0.036 & -0.039 & 0.039 \\
3 & 2 & 1460  & 0.212 & 0.168 & 0.757 & 0.138 & 0.124 & 0.839 & -0.074 & -0.045 & 0.082 \\
4 & 2 & 2555  & 0.218 & 0.168 & 0.750 & 0.165 & 0.128 & 0.808 & -0.053 & -0.040 & 0.058 \\
\hline
0 & 3 & 585   & 0.336 & 0.211 & 0.615 & 0.148 & 0.135 & 0.827 & -0.188 & -0.076 & 0.213 \\
1 & 3 & 2340  & 0.399 & 0.216 & 0.542 & 0.218 & 0.143 & 0.747 & -0.181 & -0.073 & 0.205 \\
2 & 3 & 5850  & 0.390 & 0.202 & 0.553 & 0.171 & 0.139 & 0.801 & -0.219 & -0.063 & 0.249 \\
3 & 3 & 11700 & 0.469 & 0.242 & 0.461 & 0.195 & 0.145 & 0.773 & -0.275 & -0.097 & 0.312 \\
4 & 3 & 20475 & 0.365 & 0.210 & 0.581 & 0.242 & 0.156 & 0.718 & -0.123 & -0.053 & 0.137 \\
\hline
\end{tabular}
}
\end{center}
\end{table}

\begin{figure}[htbp!]
\begin{center}
\includegraphics[width=2.1in]{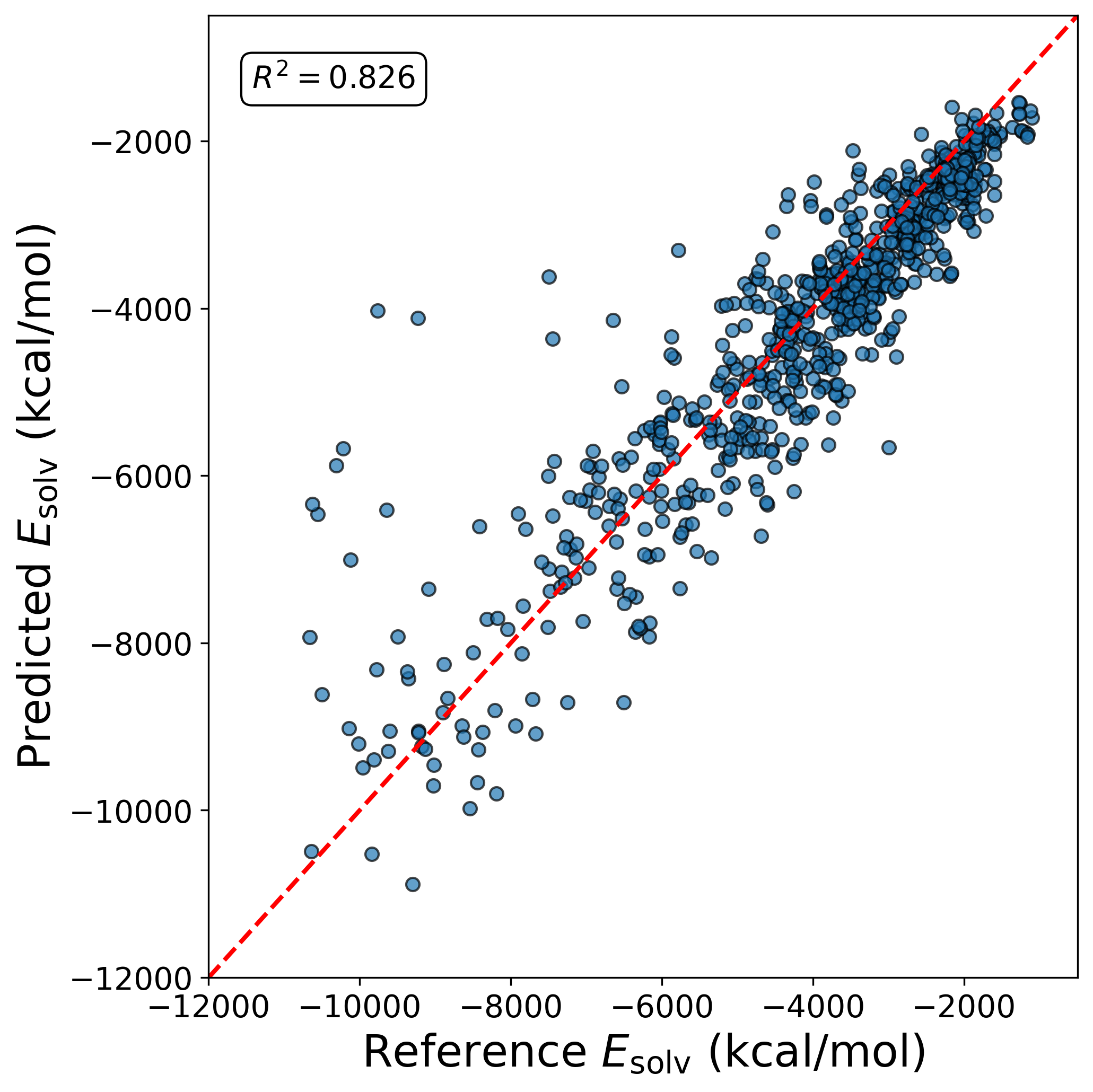}
\includegraphics[width=2.1in]{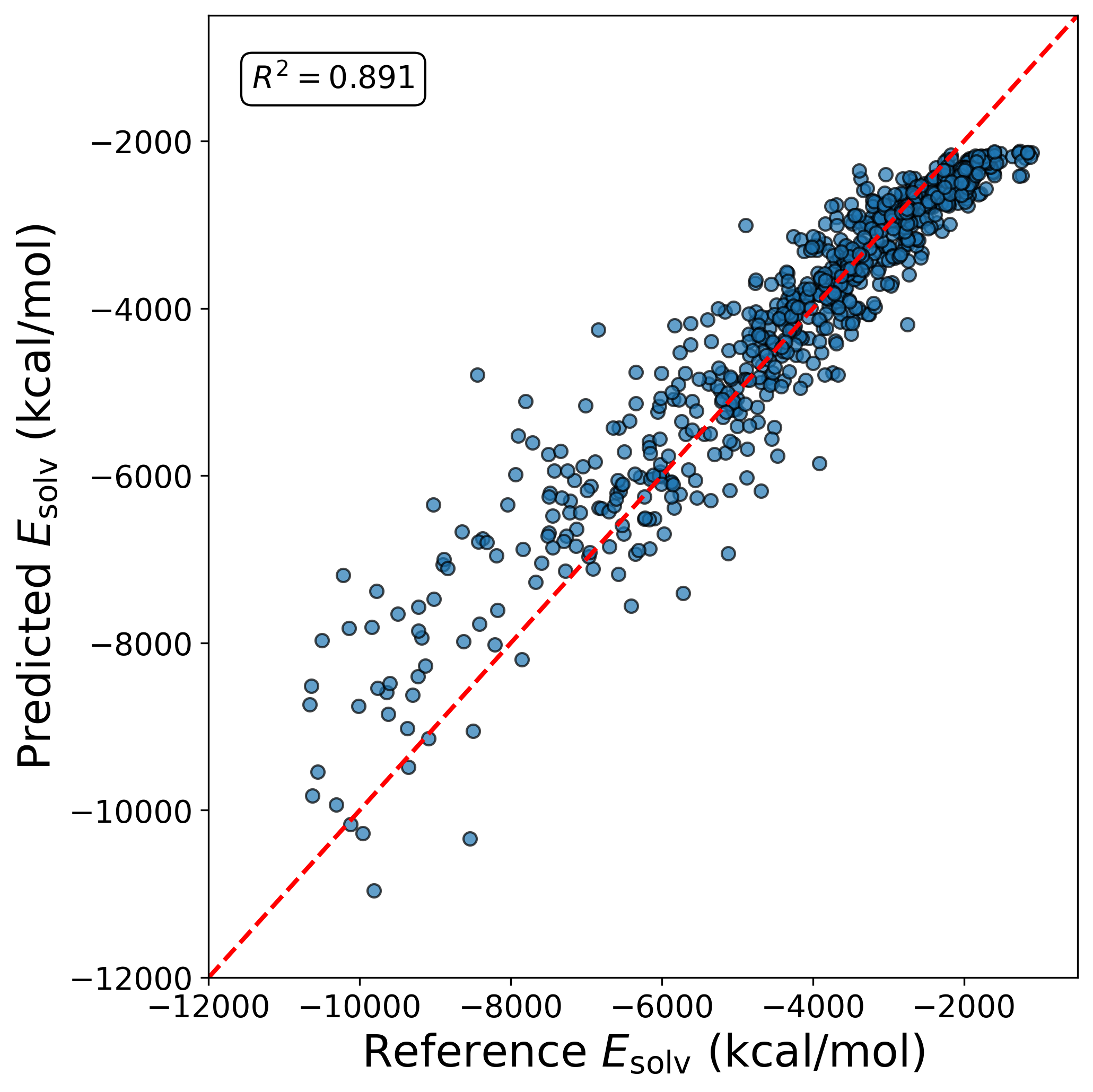}
\includegraphics[width=2.1in]{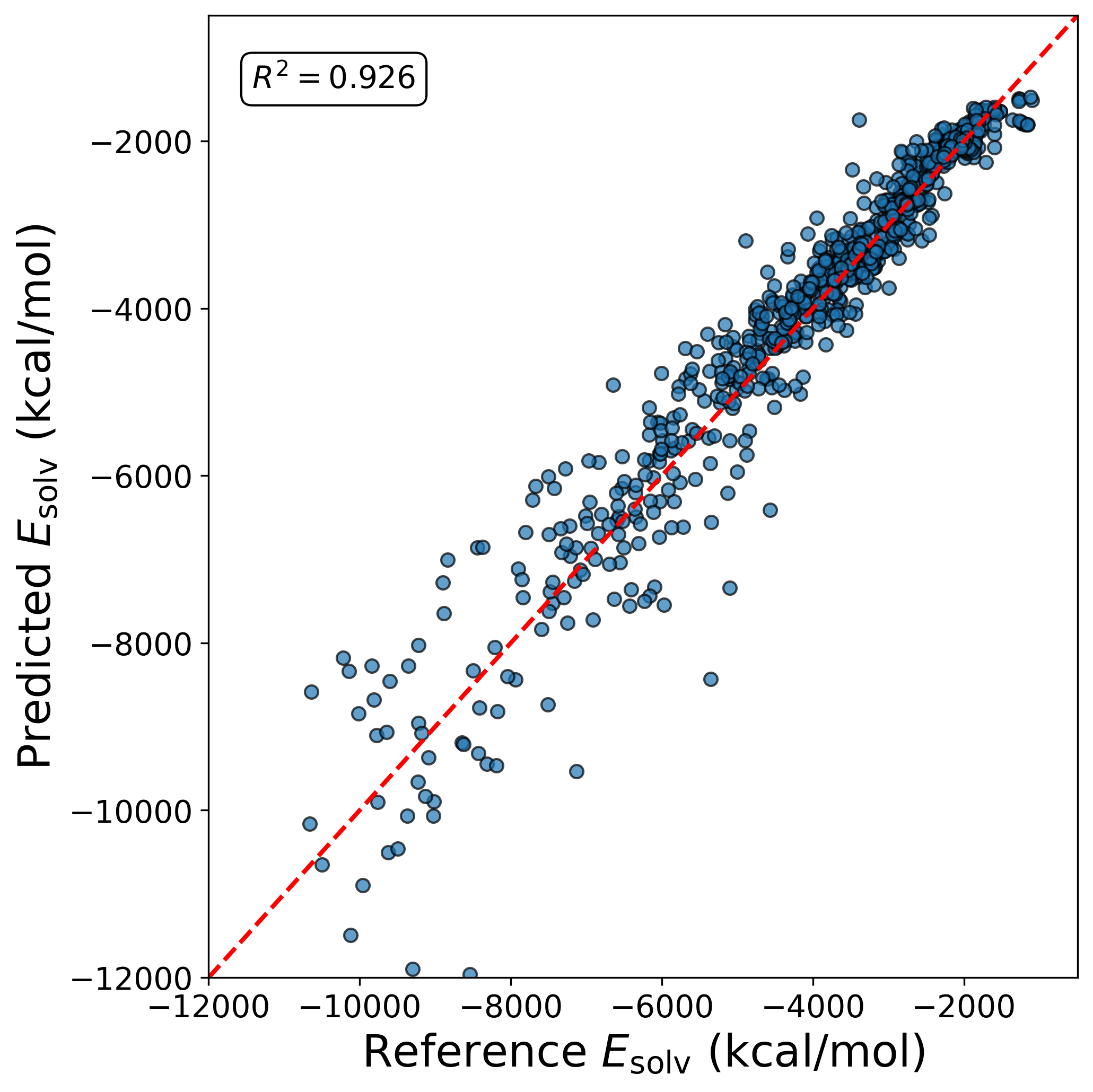}
(a) \hskip 2in (b) \hskip 2in (c)
\caption{\small  Scatter plots for the prediction of $E_\text{solv}$ on Dataset~1: (a) topological features only, (b) e-features only ($p=2$, $L=1$ with total 90 features), and (c) both features ($p=3$, $L=0$ with total 20 e-features). 
}
\label{fig_se_d1}
\end{center}
\end{figure}

The optimized case for the electrostatic-only model occurs at $p=2$ and $L=1$ with 90 features. Adding topological features further improves the performance, achieving an MSE of 0.064, a MAPE of 0.081, and an $R^2$ value of 0.926 using only 20 features at $p=3$ and $L=0$. Figure~\ref{fig_se_d1} presents scatter plots for the optimized cases, comparing reference and predicted values for models using topological features only, electrostatic features only, and the combined feature set. The benefits of combining both types of features, as illustrated in (c), can be clearly observed when compared with using only topological features in (a) and only electrostatic features in (b).

\subsection{Additional Comparison and Justification}
We here provide some comparison studies between the proposed models with 
baseline models, 
surrogate models, 
as well as 
different 
grouping and persist homologies strategies. 
In addition, some details about the cross-validation in the training process from a selected case is provided. The limitations of the proposed model and the features generation algorithms are also discussed. 
\subsubsection{$E_\text{coul}$ predicted from persistence image on Dataset~2}
\begin{figure}[htbp!]
\begin{center}
\includegraphics[width=2.1in]{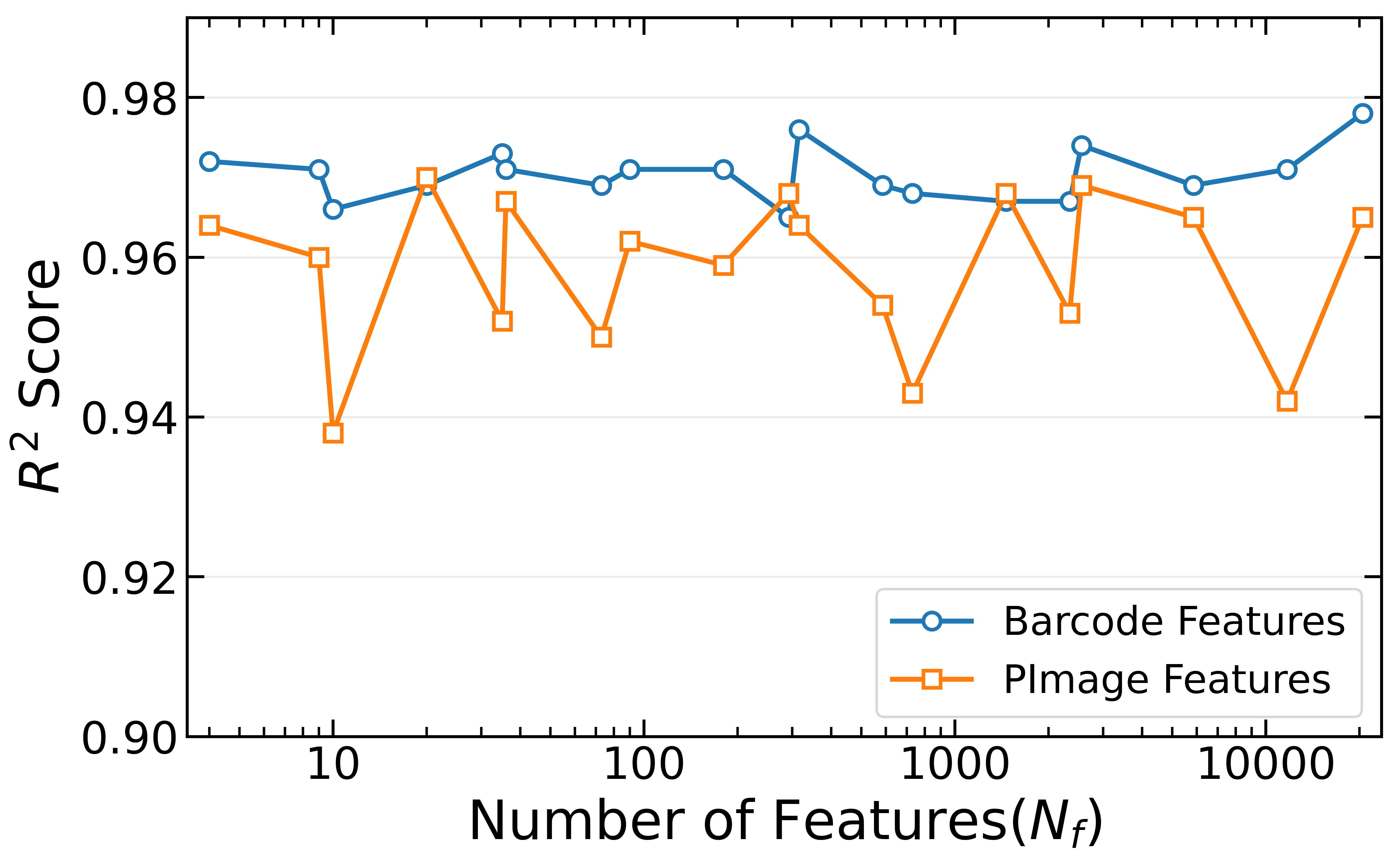}
\includegraphics[width=2.1in]{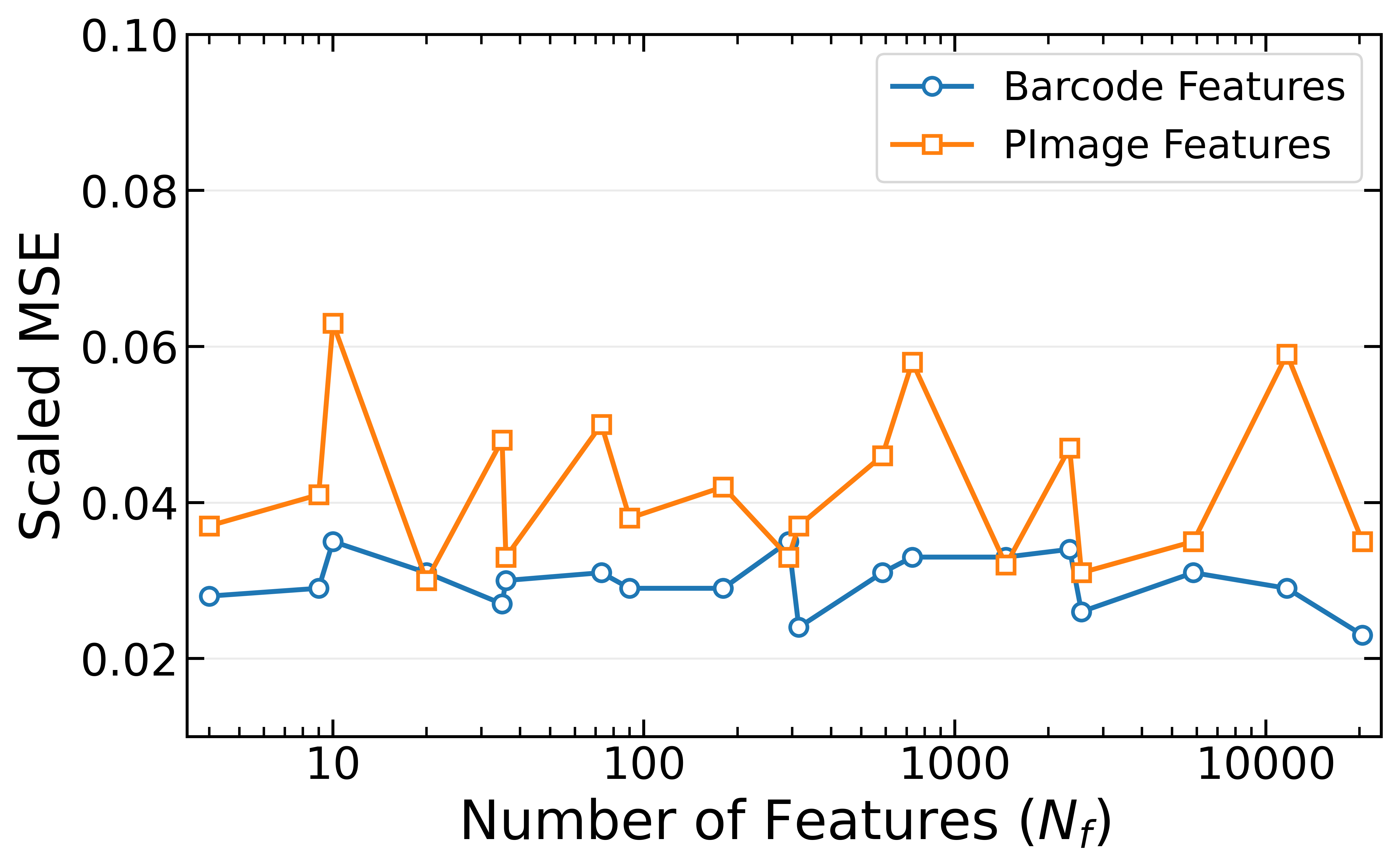}
\includegraphics[width=2.1in]{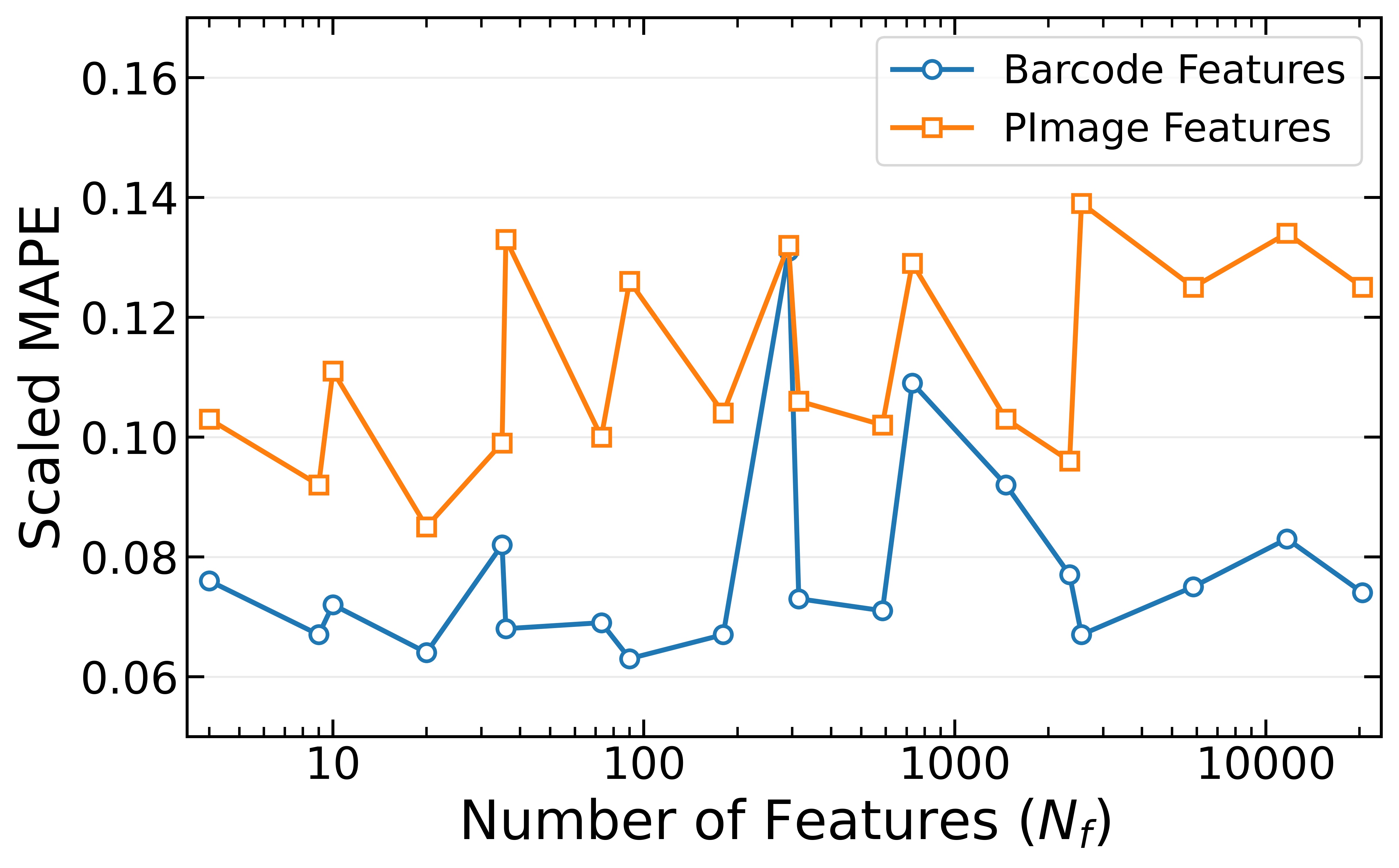}\\
(a) \hskip 2in (b) \hskip 2in (c)
\caption{\small
Prediction performance of $E_{\mathrm{coul}}$ on Dataset~2 using combined features:
(a) $R^2$ Score, (b) scaled MSE, and (c) scaled MAPE as functions of the number of features, comparing Barcode and PImage representations.
}
\label{fig_PI}
\end{center}
\end{figure}
As explained previously, the topological features can be extracted using either barcodes or persistent images. 
When only the persistence image-based topological features are used, the model achieves an MSE of 0.037, MAPE of 0.082 and an $R^2$ value of 0.964, which is comparable to the performance obtained using barcode-based topological features (MSE = 0.030, MAPE = 0.071 and $R^2$ = 0.970). Both approaches demonstrate excellent predictive accuracy even in the absence of electrostatic features. When these topological features are combined with electrostatic features, the barcode-based representation yields slightly better performance than the persistent image–based representation, as shown in Fig.~\ref{fig_PI}, in terms of $R^2$, MSE, and MAPE. Considering its relatively better performance, lower memory usage, and more efficient computation, we adopt the barcode-based approach throughout the paper. It worths noting, however, that the persistent image approach may offer advantages in terms of stability and vectorized structure, which will be investigated further in future work.
\subsubsection{Robustness of $E_\text{coul}$ prediction to homology-aware data splitting (Dataset~2)}
\begin{figure}[htbp!]
\begin{center}
\includegraphics[width=2.1in]{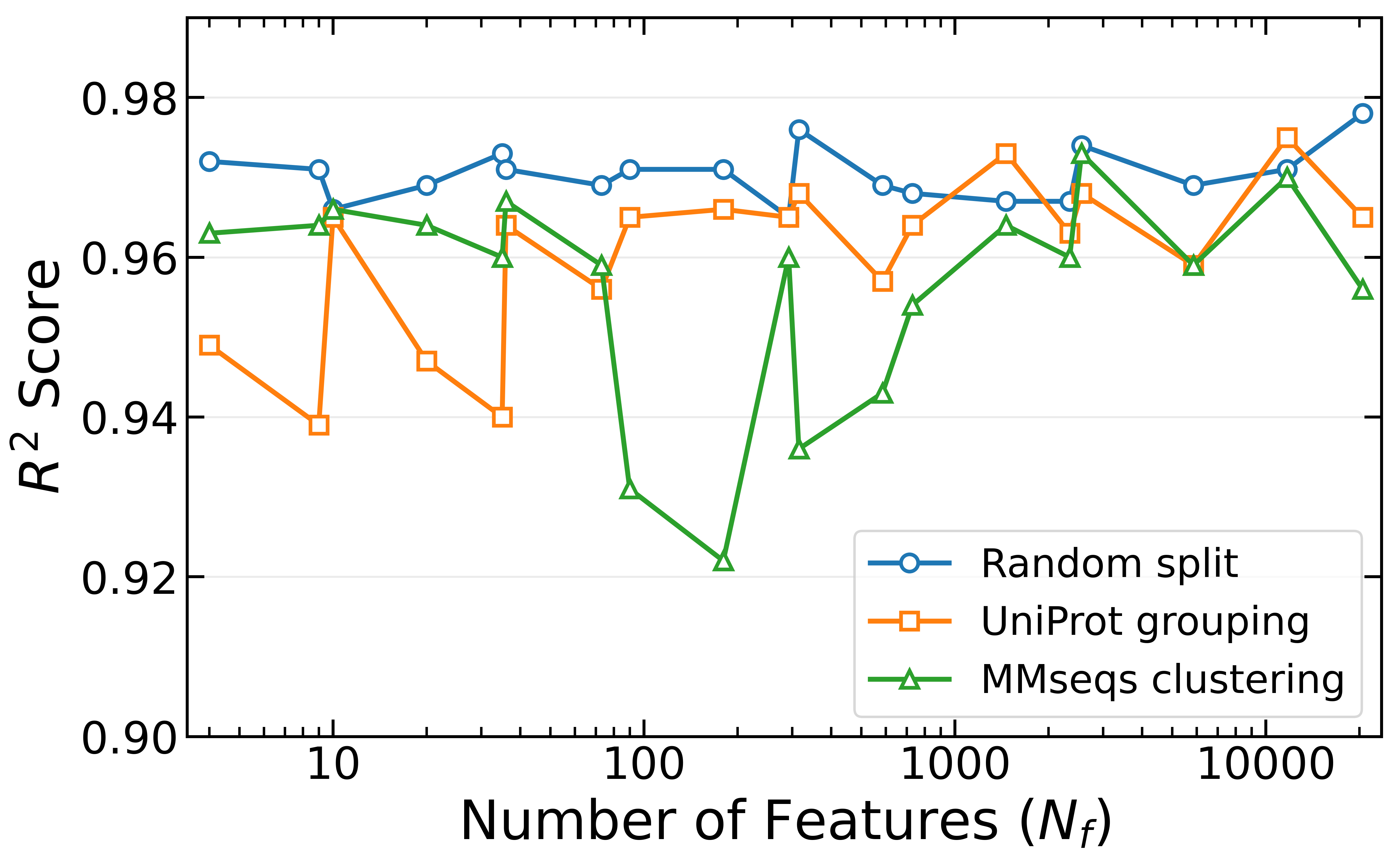}
\includegraphics[width=2.1in]{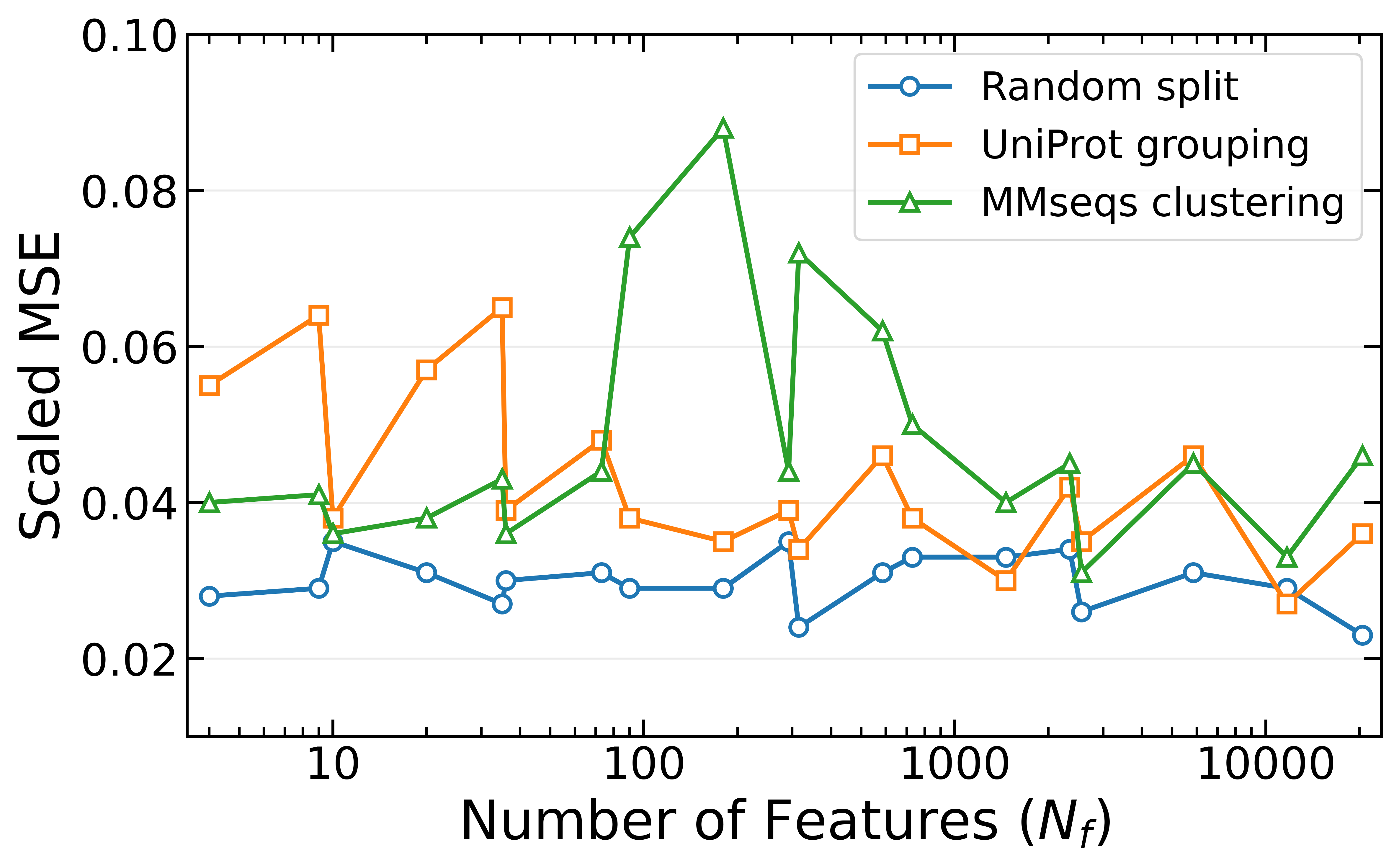}
\includegraphics[width=2.1in]{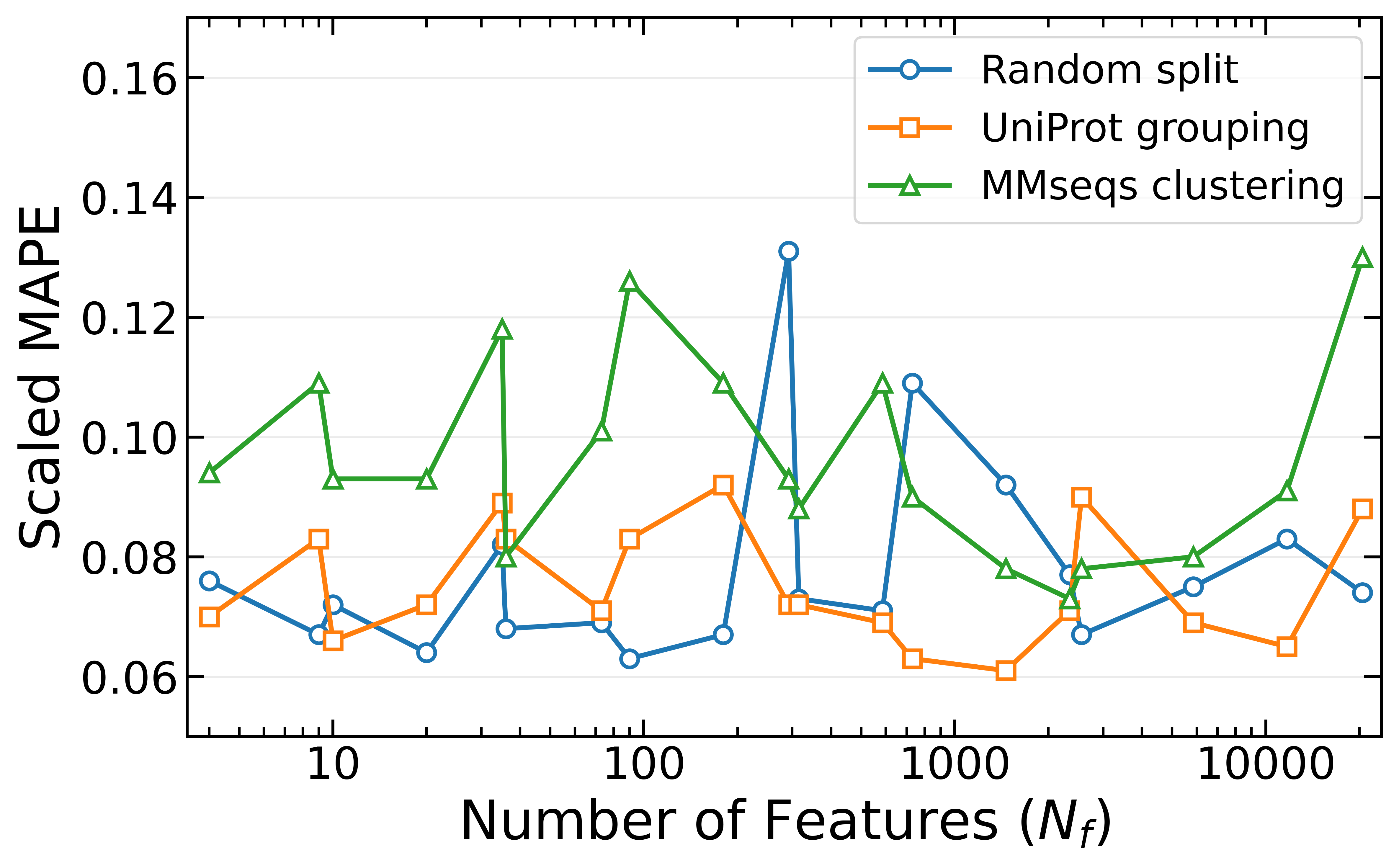}\\
\caption{\small
Performance comparison for $E_{\mathrm{coul}}$ prediction on Dataset~2 under random splitting, UniProt-based grouping, and MMseqs sequence-identity clustering (30\% identity, 80\% coverage), using combined features.
}
\label{fig_group}
\end{center}
\end{figure}
In our machine learning model, we first randomly divide the dataset into training sets and testing sets with proportions of 80\% and 20\%, respectively. The training set is further divided into five equal subsets for five-fold cross-validation, from which the model with the best validation performance is selected and subsequently evaluated on the held-out test set.
A potential concern is that the PDBbind dataset contains multiple structures corresponding to the same protein or closely related homologs (e.g., alternative conformations, mutants, or orthologous proteins). If such similar proteins are distributed in both training and testing sets, data leakage may artificially inflate performance. To mitigate this risk, we adopted two complementary strategies. First, we grouped PDB entries according to their UniProt (Universal Protein Resource) accession \cite{UniProt:2023}, ensuring that all structures of the same protein were assigned entirely to either the training or testing set. However, since proteins with different UniProt IDs may still share high sequence similarity, this alone does not eliminate homologous overlap.
We therefore applied MMseqs (Many-against-Many sequence searching) \cite{MMseqs:2017} clustering at 30\% sequence identity and 80\% coverage. Chain sequences were clustered, and PDB entries were grouped by connected components such that any entries sharing sequence similarity, including transitive relationships, were placed in the same group. Dataset splitting was then performed at the group level to prevent sequence-similarity leakage. As shown in Fig.~\ref{fig_group}, similar trends in $R^2$, scaled MSE, and scaled MAPE are observed across random, UniProt-based, and MMseqs-based splits. The slightly better performance under random splitting is expected due to less stringent generalization. 
Comparable results under these methods indicate that the model’s performance 
is not significantly driven by the memorization of homologous sequences.
%
\subsubsection{Cross-Validation Performance}
\begin{figure}[htbp!]
\begin{center}
\includegraphics[width=3.2in]{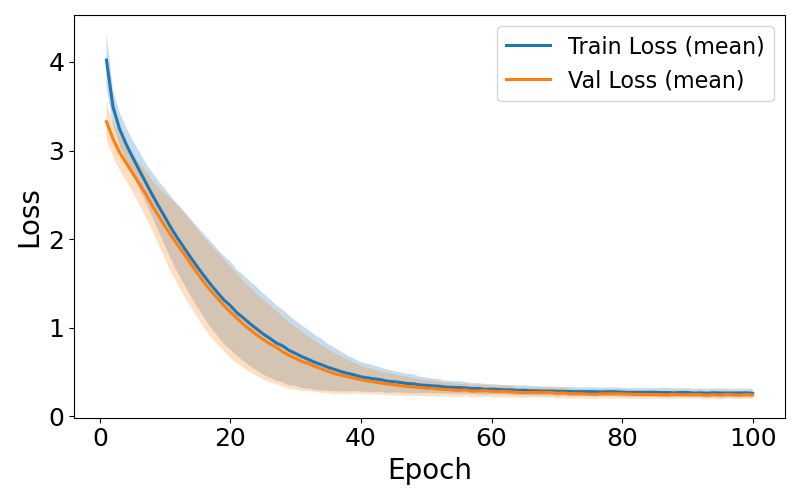}
\includegraphics[width=3.2in]{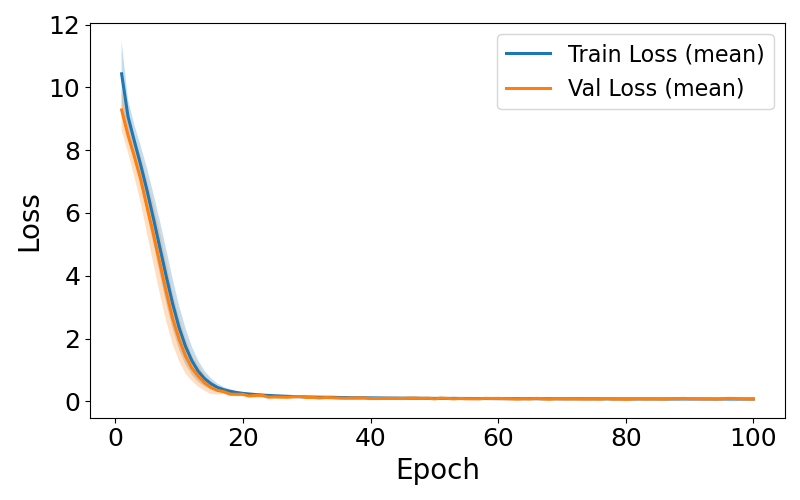}\\
(a) \hskip 3in (b)
\caption{\small
Cross-validation training histories for $E_{\mathrm{coul}}$ prediction ($p=4$ and $L=1$) on Dataset~2; (a) Electrostatic features only, (b) Combined features. The mean training and validation losses are shown as solid lines, with shaded bands representing standard deviation across folds. 
}
\label{fig_loss}
\end{center}
\end{figure}

Now we investigate the training process that uses five-fold cross-validation for over-fitting checks. 
In Fig.~\ref{fig_loss}, we report the cross-validation training histories or learning curves for $E_{\mathrm{coul}}$ prediction ($p=4$ and $L=1$) on Dataset~2. From the mean training and validation loss curves and the corresponding one-standard-deviation bands, we can see that the selected model is stable and free of overfitting. Meanwhile, the performance of using both types of features is significantly better than using electrostatic features only, as indicated by faster convergence and narrower bands. Similar trends are observed for models predicting other physical quantities across different datasets. 

\begin{figure}[htbp!]
\begin{center}
\includegraphics[width=3.2in]{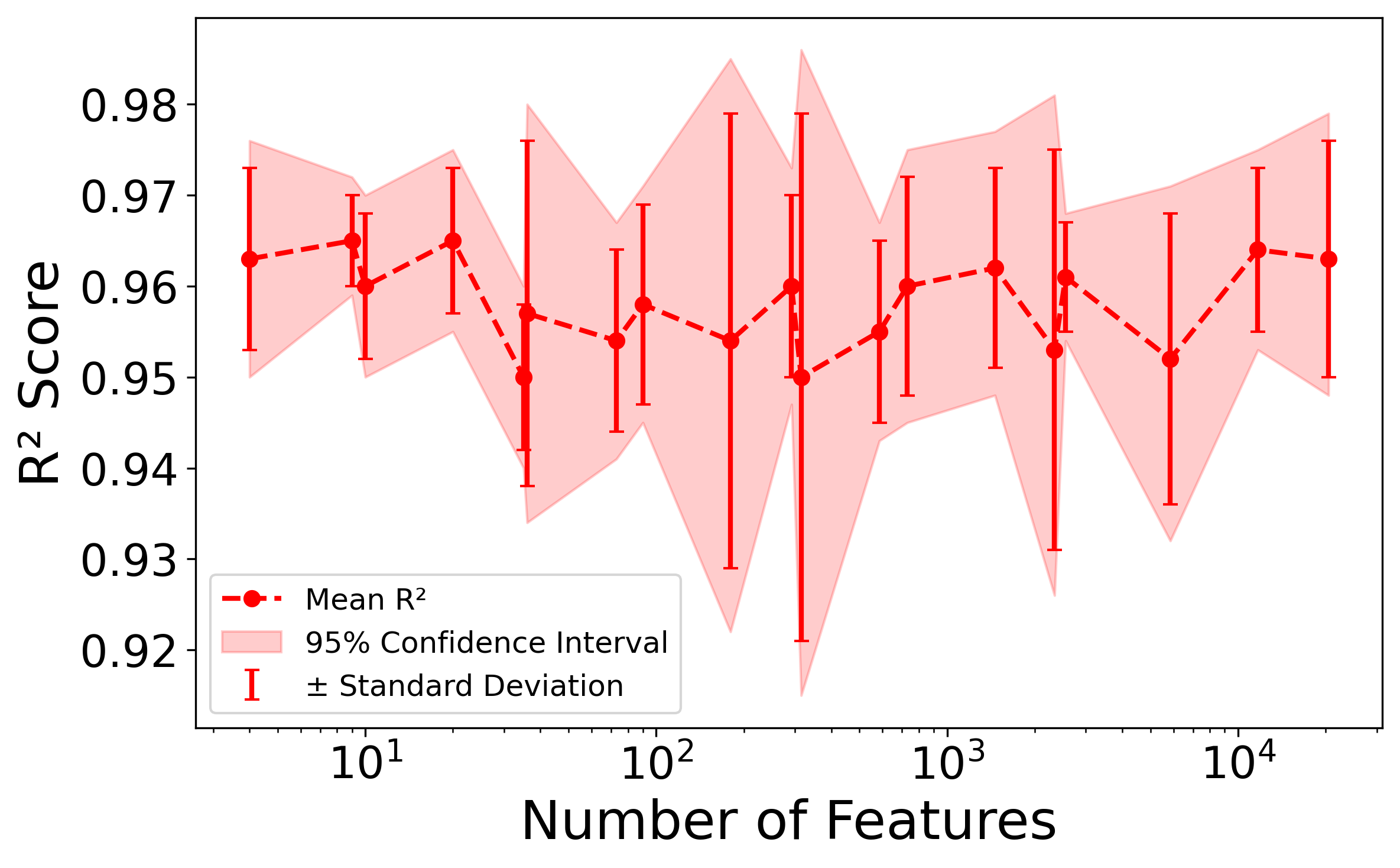}
\includegraphics[width=3.2in]{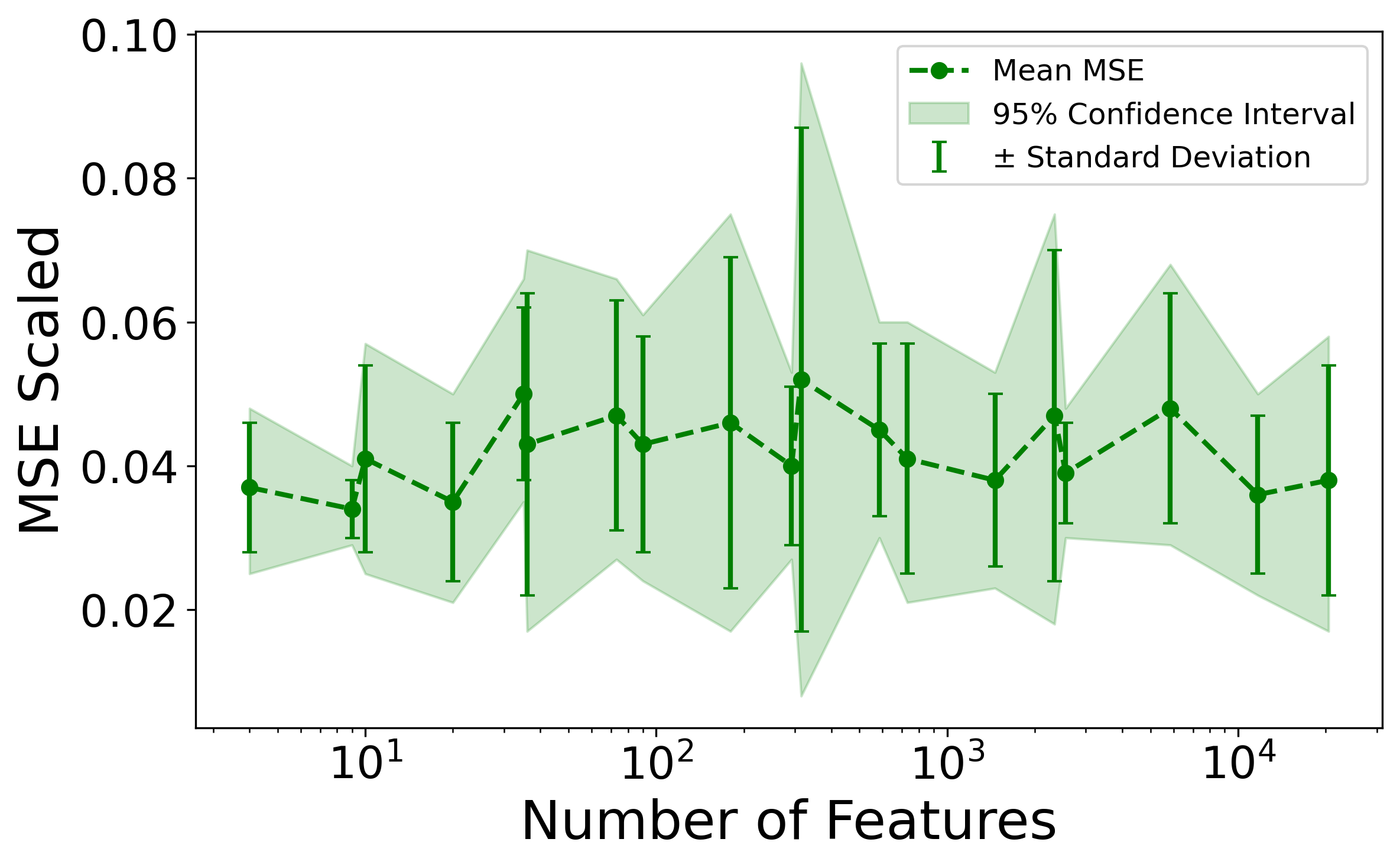}
\\
\caption{\small
Validation performance during training across all cross-validation folds using combined features. Solid lines show mean performance across folds, while shaded regions and error bars indicate 95\% confidence intervals and $\pm$ one standard deviation, respectively.
}
\label{fig_mean_std}
\end{center}
\end{figure}

We further show in Fig.~\ref{fig_mean_std} the mean, 95\% confidence interval (assuming $t$-distribution), and one standard deviation band of $R^2$ in (a) and MSE in (b) from the validation results of the five-fold cross-validation training procedure 
across different numbers of electrostatic features when combined features are used. The narrow band and interval widths demonstrate the stability of the model. 
Note that for $E_{\mathrm{coul}}$, the advantages of using more electrostatic features are not obvious, since the contribution of topological features dominates.
\subsubsection{Comparison with baseline models}
\begin{figure}[htbp!]
\begin{center}
\includegraphics[width=3.2in]{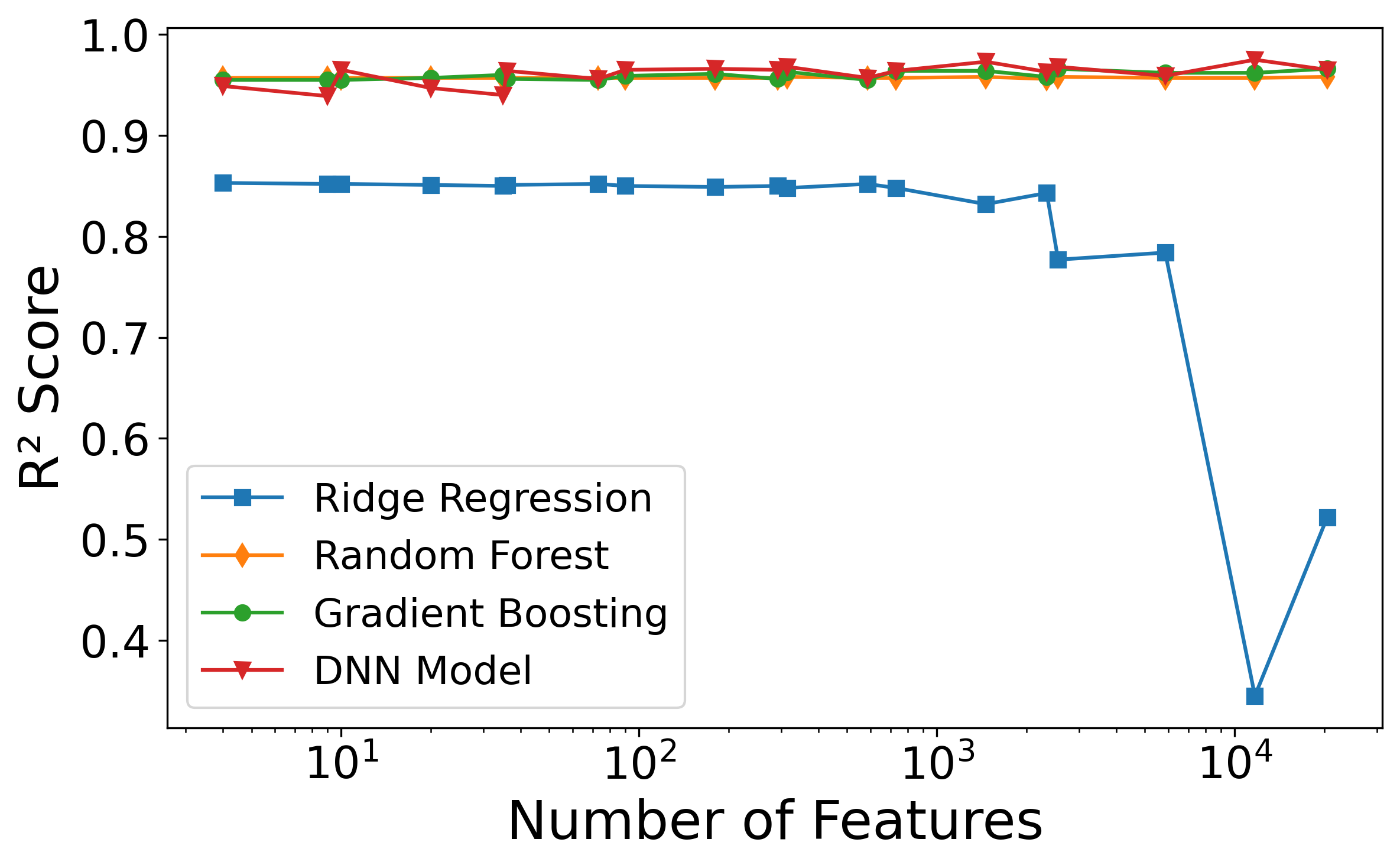}
\includegraphics[width=3.2in]{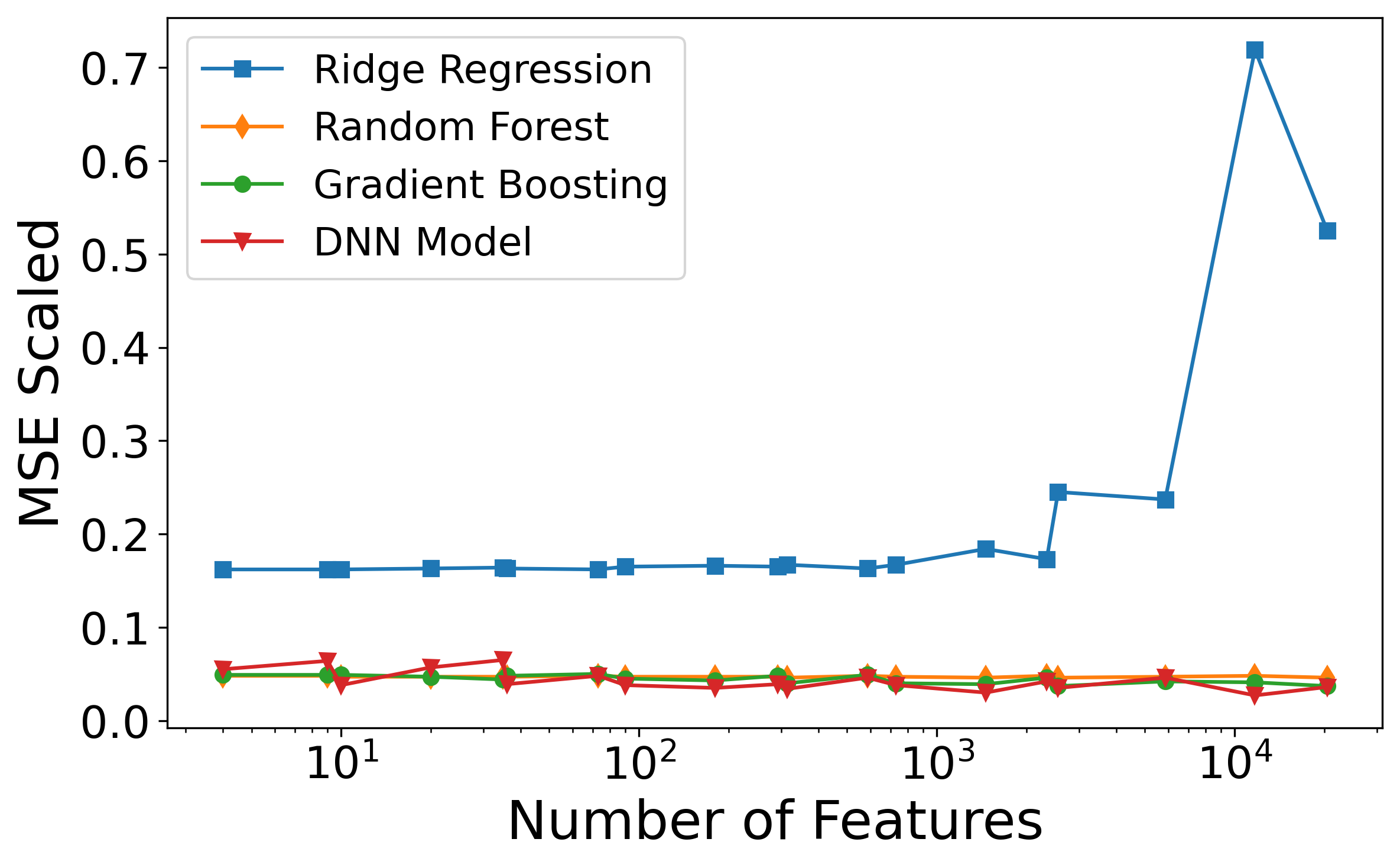}\\
\caption{\small
Prediction performance of the proposed DNN model compared with baseline methods—Ridge Regression, Random Forest, and Gradient Boosting—across varying numbers of input features. }
\label{fig_baseline}
\end{center}
\end{figure}
We next present the comparison results between the deep neural network model and other baseline models, including ridge regression, random forest, and gradient boosting. All methods using the same number of electrostatic and topological features, with the number of electrostatic features increasing as shown in Fig.~\ref{fig_baseline}. From the comparison, we observe that DNN, Random Forest, and Gradient Boosting, as more flexible methods, perform significantly better than Ridge Regression. The DNN shows slight improvements over Random Forest and Gradient Boosting in terms of $R^2$ and MSE. 

\subsubsection{CPU time comparison and Size-Dependent Performance}
\begin{figure}[htbp!]
\begin{center}
\includegraphics[width=3in]{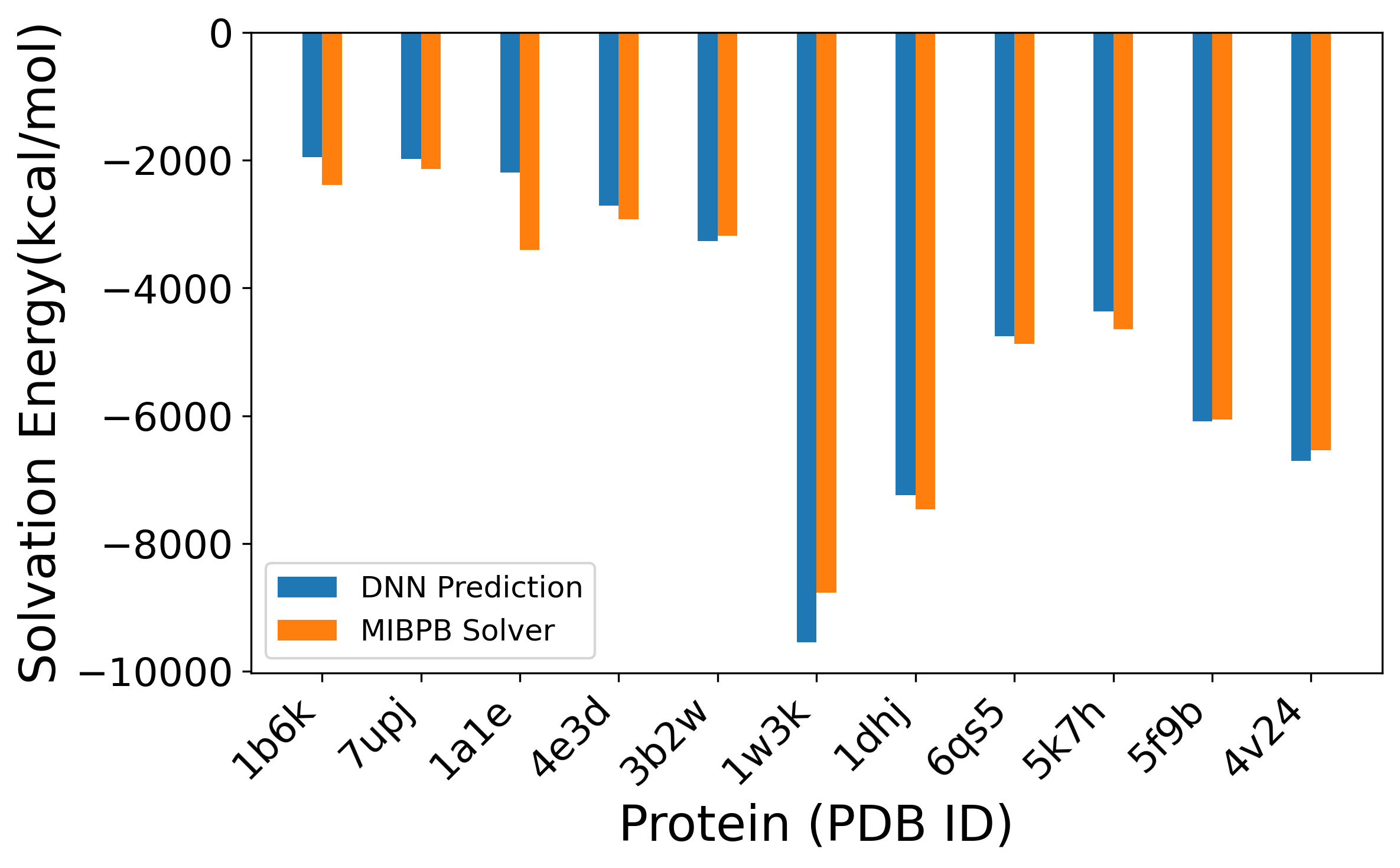}
\includegraphics[width=3in]{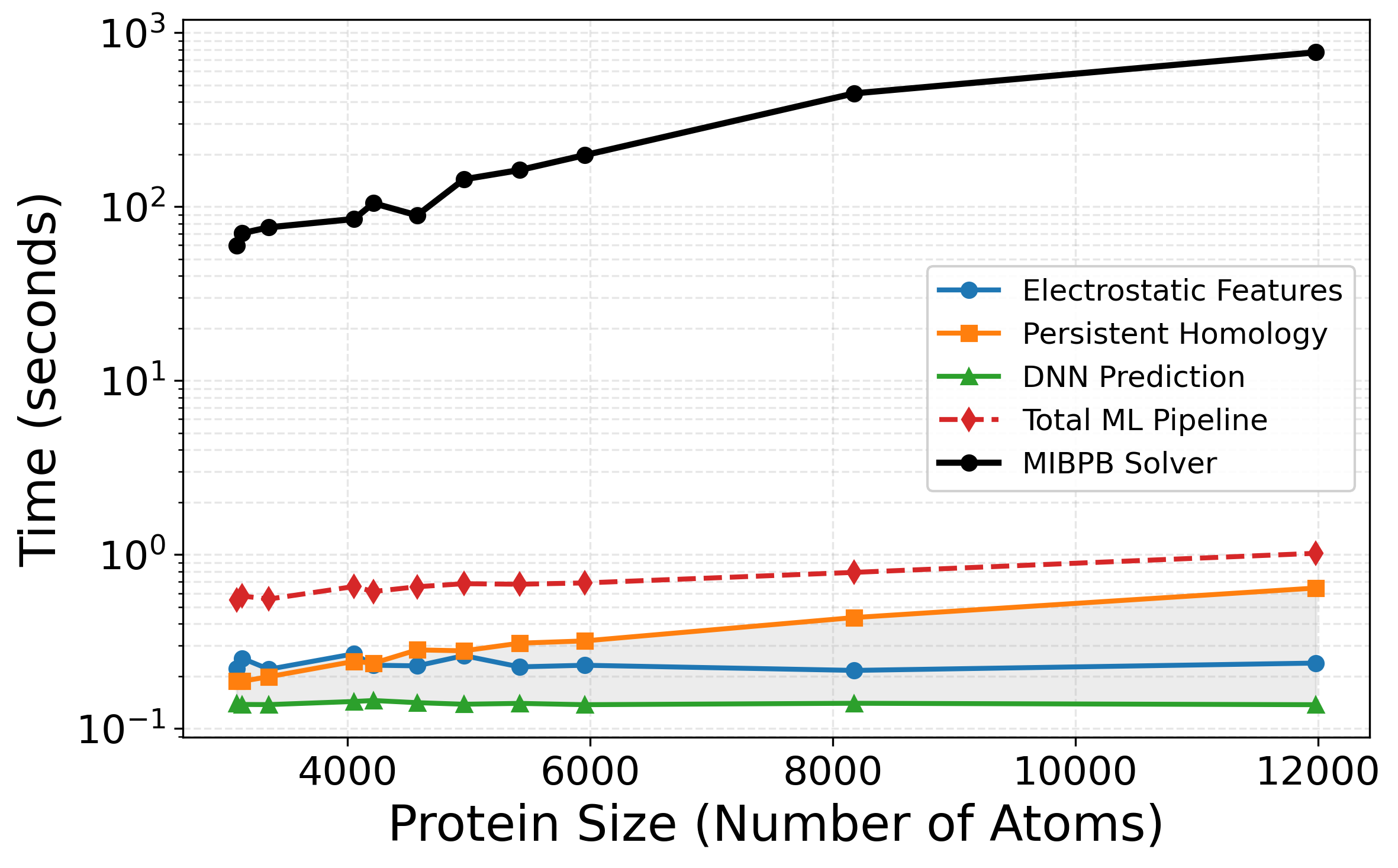}\\
(a) \hskip 3in (b) 
\caption{\small
Performance comparison between the proposed DNN model and the MIBPB solver. 
(a) $E_\text{solv}$ predicted by the learned model versus that is computed by MIBPB solver selected protein structures;
(b) Computational runtime with respect to protein size (number of atoms), demonstrating the significant efficiency of the DNN model. 
The MIBPB solver is evaluated with grid size $h=0.5$, dielectric constants $\epsilon_1=1$, $\epsilon_2=80$, and inverse Debye length $\kappa=0$. 
}
\label{fig_time}
\end{center}
\end{figure}

In order to show the efficiency of the learned model for predicting $E_\text{solv}$ compared with the MIBPB solver, we report the CPU time and solvation energies $E_\text{solv}$ obtained from both approaches in Fig.~\ref{fig_time}. As shown in (a), the predicted $E_\text{solv}$ values are closely aligned; whereas in (b), the time required by the learned model in red is significantly shorter than the runtime of the MIBPB solver in black, which increases rapidly with molecular size. 
Also as shown in Fig.~\ref{fig_time}(b), the total machine learning pipeline time in red is further decomposed into time to compute electrostatic features in blue, time to compute topological features in orange, the time to predict solvation energies using the learned model in green. 

\begin{figure}[htbp!]
\begin{center}
\includegraphics[width=3in]{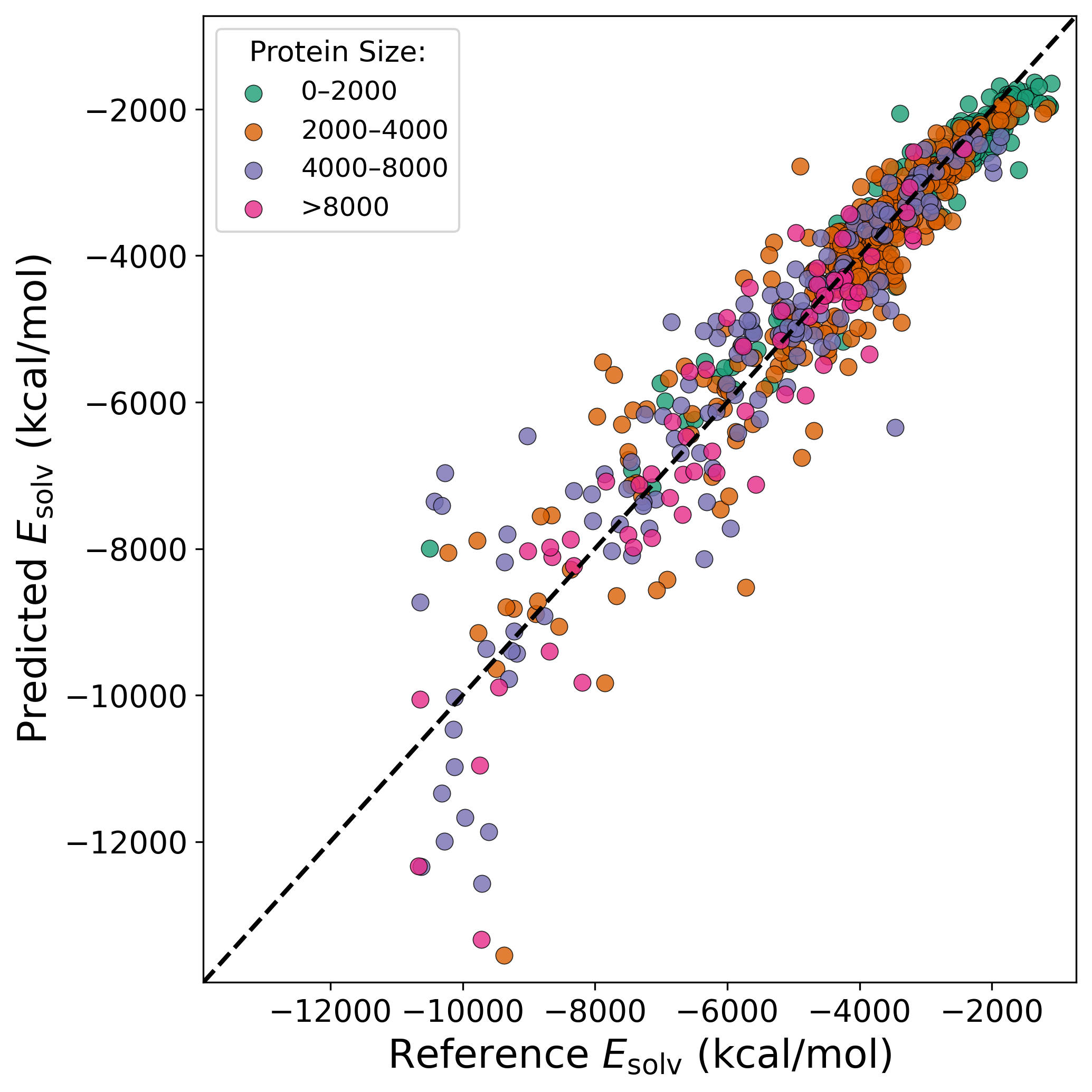}
\includegraphics[width=3in]{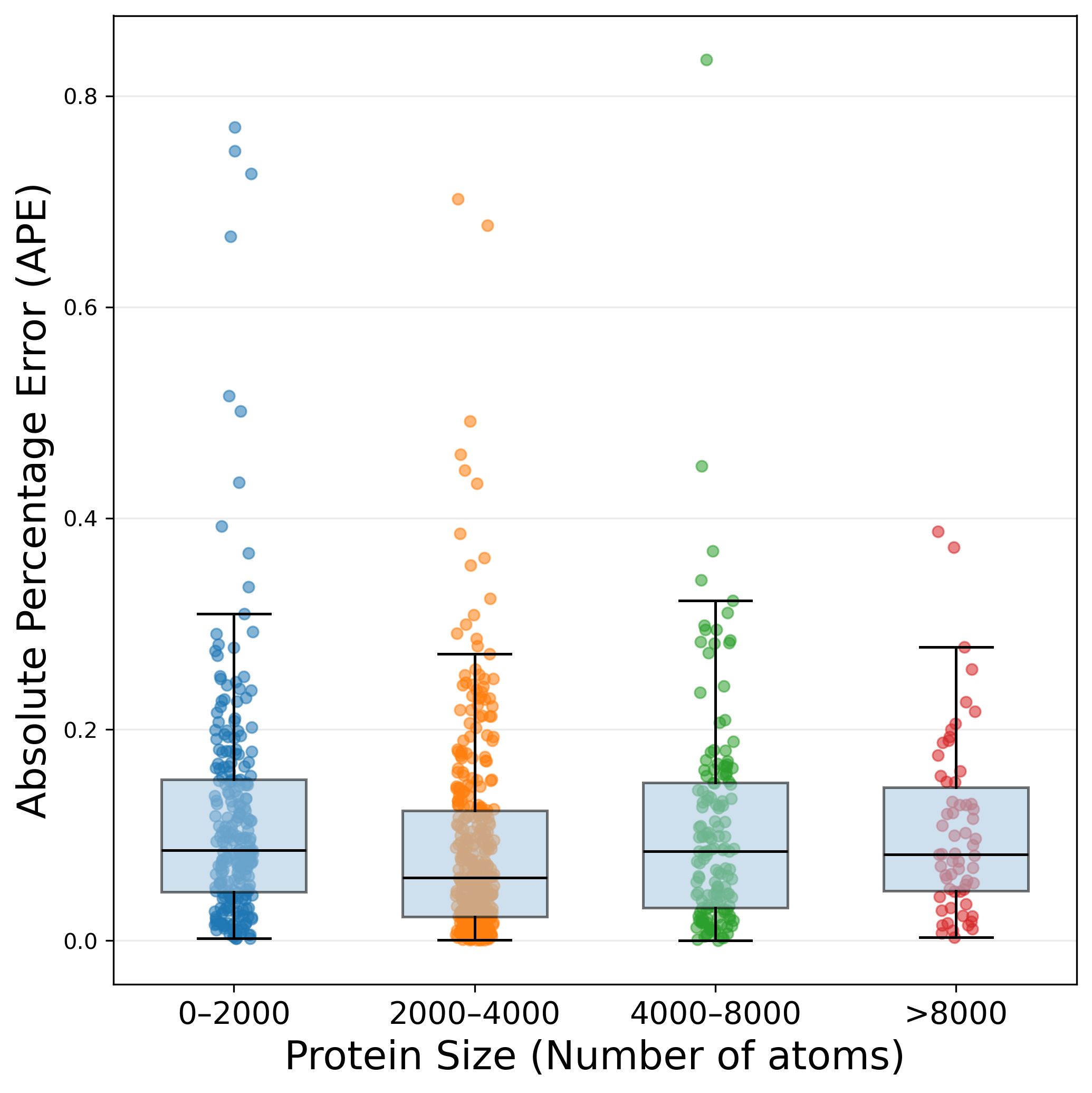}\\
(a) \hskip 3in (b) 

\caption{\small
Protein size–dependent performance of the DNN model. (a) Predicted versus reference $E_{\mathrm{solv}}$ colored by protein size. (b) Distribution of absolute percentage error across protein size ranges.}  

\label{fig_size}
\end{center}
\end{figure}

To further evaluate the robustness of the model with respect to protein size, 
Fig.~\ref{fig_size} presents a size-dependent performance analysis. 
The scatter plot (predicted value {\it vs} reference value) in (a) 
and Box-and-whisker plot (absolute percentage error) in (b) both indicate the model's uniform performance across protein sizes.

\subsubsection{Model comparison with testing hypothesis}
Theoretically, the three research purposes as mentioned earlier can be treated as statistical hypotheses and evaluated through testing hypothesis for model comparison. However, this would require pairwise comparisons between models with numerous simulations to generate statistically significant metrics, which is a very costly procedure. Here, we present one such case to illustrate the approach. We select a particular set of electrostatic features $(p=2, L=1)$ to compare three models $M_e$, $M_t$, and $M_{e,c}$, which use electrostatic features only, topological features only, and combined features, respectively. These three models are used to predict $E_\text{solv}$ on Dataset~2 using 10 independent random 80--20 train--test splits, and the corresponding $R^2$ and MSE values are recorded. The resulting $R^2$ and MSE metrics, summarized in Table~\ref{tb_wilcoxon}, are then used to test the following hypotheses:\\ 
$H_0$: $M_{e,t}$ performs as well as $M_e$ or $M_t$\\
$H_a$: $M_{e,t}$ performs better than $M_e$ or $M_t$\\
\begin{table}[t]
\centering
\caption{\small Metrics obtained by running models $M_e$, $M_t$, and $M_{e,t}$ for the prediction of $E_\text{solv}$ using 10 independent random 80--20 train--test splits on Dataset~2.}
\label{tb_wilcoxon}
{\small
\begin{tabular}{l|cccccccccc|c}
\hline
Metrics 
& $S_1$ & $S_2$ & $S_3$ & $S_4$ & $S_5$ & $S_6$ & $S_7$ & $S_8$ & $S_9$ & $S_{10}$ & $\bar{S_i}$ \\
\hline
$R^2(M_e)$ 
& 0.876 & 0.885 & 0.883 & 0.853 & 0.876 & 0.859 & 0.879 & 0.862 & 0.873 & 0.885 & 0.873\\
$R^2(M_t)$ 
& 0.838 & 0.802 & 0.816 & 0.847 & 0.857 & 0.803 & 0.844 & 0.812 & 0.852 & 0.810 & 0.828\\
$R^2(M_{e,t})$ 
& 0.935 & 0.936 & 0.907 & 0.937 & 0.926 & 0.887 & 0.931 & 0.914 & 0.921 & 0.899 & 0.919\\
\hline
MSE$(M_e)$ 
& 0.120 & 0.121 & 0.123 & 0.147 & 0.116 & 0.148 & 0.125 & 0.142 & 0.126 & 0.111 & 0.128\\
MSE$(M_t)$ 
& 0.157 & 0.214 & 0.199 & 0.155 & 0.133 & 0.207 & 0.161 & 0.200 & 0.147 & 0.186 & 0.176\\
MSE$(M_{e,t})$ 
& 0.063 & 0.067 & 0.097 & 0.063 & 0.068 & 0.119 & 0.073 & 0.092 & 0.078 & 0.098 & 0.082 \\
\hline
\end{tabular}
}
\end{table}\\
The four paired one-sided Wilcoxon signed-rank tests on $R^2$ and MSE between $M_{e,t}$ and $M_e$ or $M_t$ all return $p$-value < 0.001 thus the null hypotheses are rejected, indicating model $M_{e,t}$ is significantly better than $M_e$ or $M_t$.
\subsubsection{Limitation}
The manuscript focuses on providing algorithms for feature generation for proteins in general and is not restricted by protein size or type. However, we do notice some limitations for the proposed methods, as listed below. \\
(1) The electrostatic features primarily focus on the protein electrostatic profile therefore they have limited impact on capturing the interplay between the protein and its surrounding environment. 
This limitation is reflected in the simulation results, 
where Coulombic energy prediction performs better than solvation energy prediction. In future work, we plan to enhance the electrostatic features by incorporating reaction potentials generated from the GB model at atomic centers to better characterize protein-solvent interactions.\\    
(2) When the dataset size is small, for the same types of features, the advantages of DNN over Gradient Boosting or Random Forrest are difficult to justify. \\
(3) Identifying the optimal values of $p$ and $L$ for generating electrostatic features is challenging, as they depend on factors such as dataset size, response types, and the performance of topological features. 
This limitation requires multiple tests to determine the optimal $p$ and $L$ during model training. 
We are planning to implement and test the Barycentric treecode approach \cite{Wang:2020} as explained in the supplementary material for potential solutions. 

\section{Conclusion}

The project provides a novel approach for generating combined topological and electrostatic features to represent protein structures and charge distributions in a physical-informed multiscale and uniform manner. Using these features, we can predict important protein properties such as solvation energy and Coulomb energy. The numerical simulations validate our three main findings: larger dataset, higher-resolution electrostatic features, and  the combination of features all improve the performance of the ML model. Beyond predicting solvation and Coulomb energies, the resulting electrostatic and topological features have the potential for broad applicability, particularly in representing proteins as uniform and multiscale features for ML models.  
In the future work, we will further investigate the role of the reaction potential, which will be computed using the much faster GB model, as discussed in the manuscript. We will also explore the use of the barycentric Lagrange treecode for generating electrostatic features, in comparison with the Cartesian treecode.

\section{Data and Software Availability }
The code and data associated with this project are freely and publicly available in the GitHub repository: \url{https://github.com/atalha-cmd/Biophysics_DNN}, maintained by SMU graduate student Md Abu Talha. It contains the Fortran code for generating electrostatic features using Cartesian treecode, the Python code to generate topological features (a modified version as used in \cite{Cang:2017}), the Python code for the neural network originating from Elyssa Sliheet's thesis work \cite{Sliheet:2024}, and the data required to reproduce the reported results.

\begin{acknowledgement}

This work of ES, MAT, and WG was supported in part by National Science Foundation (NSF) grants DMS-2110922 and DMS-2110869.  ES was also partially supported by the NSF RTG-1840260 grant. We thank the SMU O'Donnell Data Science and Research Computing Institute (ODSRCI) for providing computing hardware and software. These resources together made this project possible.\\ 

\end{acknowledgement}

\begin{suppinfo}
%
%
The following files are available free of charge.
\begin{itemize}
\item supple-achemso.pdf: this file contains additional information of models, methods, and results which are less essential but could assist the readers in better understanding or reproducing the proposed work.  
\end{itemize}
\end{suppinfo}

\bibliography{pb_review_um}

\end{document}